\documentclass[a4paper, 10pt, conference]{ieeeconf} 
\IEEEoverridecommandlockouts  

\makeatletter
\let\NAT@parse\undefined
\makeatother
\usepackage{natbib}
\usepackage{ upgreek }
\usepackage{verbatim} 
\usepackage{tikz} 

\usepackage{amssymb}
\usepackage{graphicx}
\usepackage{url}
\usepackage{verbatim} 
\usepackage{color}
\usepackage{array}
\usepackage{relsize}
\usepackage{multirow}
\usepackage{amsmath}

\DeclareMathOperator*{\argmax}{argmax}

\long\def\symbolfootnote[#1]#2{\begingroup%
\def\thefootnote{\fnsymbol{footnote}}\footnotetext[#1]{#2}\endgroup}
\newlength\savedwidth
\newcommand\whline[1]{\noalign{\global\savedwidth\arrayrulewidth
                               \global\arrayrulewidth #1} %
                      \hline
                      \noalign{\global\arrayrulewidth\savedwidth}}

\newcommand{\todo}[1]{\textcolor{blue}{\textbf{#1}}}

\renewcommand{\Re}{\mathbb{R}}

\newcommand{\M}{{M}}             
\newcommand{\x}{{\mathbf x}}     
\newcommand{\xh}[1]{{\mathbf x^{h_{#1}}}}     
\newcommand{\y}{{\mathbf y}}     
\newcommand{\yh}[1]{{\mathbf y^{h_{#1}}}}     
\newcommand{\yhhat}[1]{{\mathbf {\hat{y}}^{h_{#1}}}}     
\newcommand{\ys}[1]{{\mathbf y_{#1}}}    
\newcommand{\ysc}[2]{{y_{#1}^{#2}}}    

\newcommand{\zsc}[2]{{z_{#1}^{#2}}}    

\newcommand{\fo}[1]{{\phi_o({#1})}}      
\newcommand{\fs}[1]{{\phi_a({#1})}}      
\newcommand{\fe}[3]{{\phi_{#1}(#2,#3)}}
\newcommand{\fte}[3]{{\phi^t_{#1}(#2,#3)}}
\newcommand{\w}{{\mathbf w}}           
\newcommand{\wh}[1]{{\mathbf w^{h_{#1}}}}           
\newcommand{\wo}[1]{{\mathbf{w}_\mathbf{o}^{#1}}}        
\newcommand{\ws}[1]{{\mathbf{w}_\mathbf{a}^{#1}}}        
\newcommand{\we}[3]{{\mathbf{w}_{\mathbf{#1}}^{#2#3}}}   
\newcommand{\wte}[3]{{\mathbf{w^t}_{\mathbf {#1}}^{#2#3}}}   

\newcommand{\df}[3]{{E_{#3}(#1,#2)}}   
\newcommand{\dg}[3]{{g_{#3}(#1,#2)}}   
\newcommand{\loss}[2]{{\Delta(#1,#2)}}   

\title{\LARGE \bf
Learning Human Activities and Object Affordances from RGB-D Videos}
\author{ Hema Swetha Koppula, Rudhir Gupta, Ashutosh Saxena\\
Department of Computer Science, Cornell University, USA.\\
\tt{\{hema,rg495,asaxena\}@cs.cornell.edu}}

\begin{document}
\maketitle

\begin{abstract} 
Understanding human activities and object affordances are two very important 
skills,
especially for personal robots which operate in human environments.
In this work, we consider the problem of extracting a descriptive labeling
of the sequence of sub-activities being performed by a human, and 
more importantly, of their interactions with the objects in the form of 
associated affordances.
Given a RGB-D video, we jointly model the human activities 
and object affordances
as a Markov random field where the nodes represent  objects and 
 sub-activities, and the edges represent the relationships between object 
affordances, their relations with sub-activities, and their evolution over
time. We formulate the learning problem using a structural support vector machine 
(SSVM) approach, where labelings over various alternate temporal 
segmentations are considered as latent variables. We tested our method
on a challenging dataset comprising 120 activity videos collected from 4 subjects,
and obtained an accuracy of 79.4\% for affordance, 63.4\% for  sub-activity and 
75.0\% for high-level activity labeling.
We then demonstrate the use of such descriptive labeling
in performing assistive tasks by a PR2 robot.\footnote{A first version of this work
was made available on arXiv \citep{Koppula12} for faster dissemination of scientific work.}
\end{abstract}



 \section{Introduction}

\begin{figure}[t!]
\centering
\includegraphics[width=.49\linewidth,height=1in]{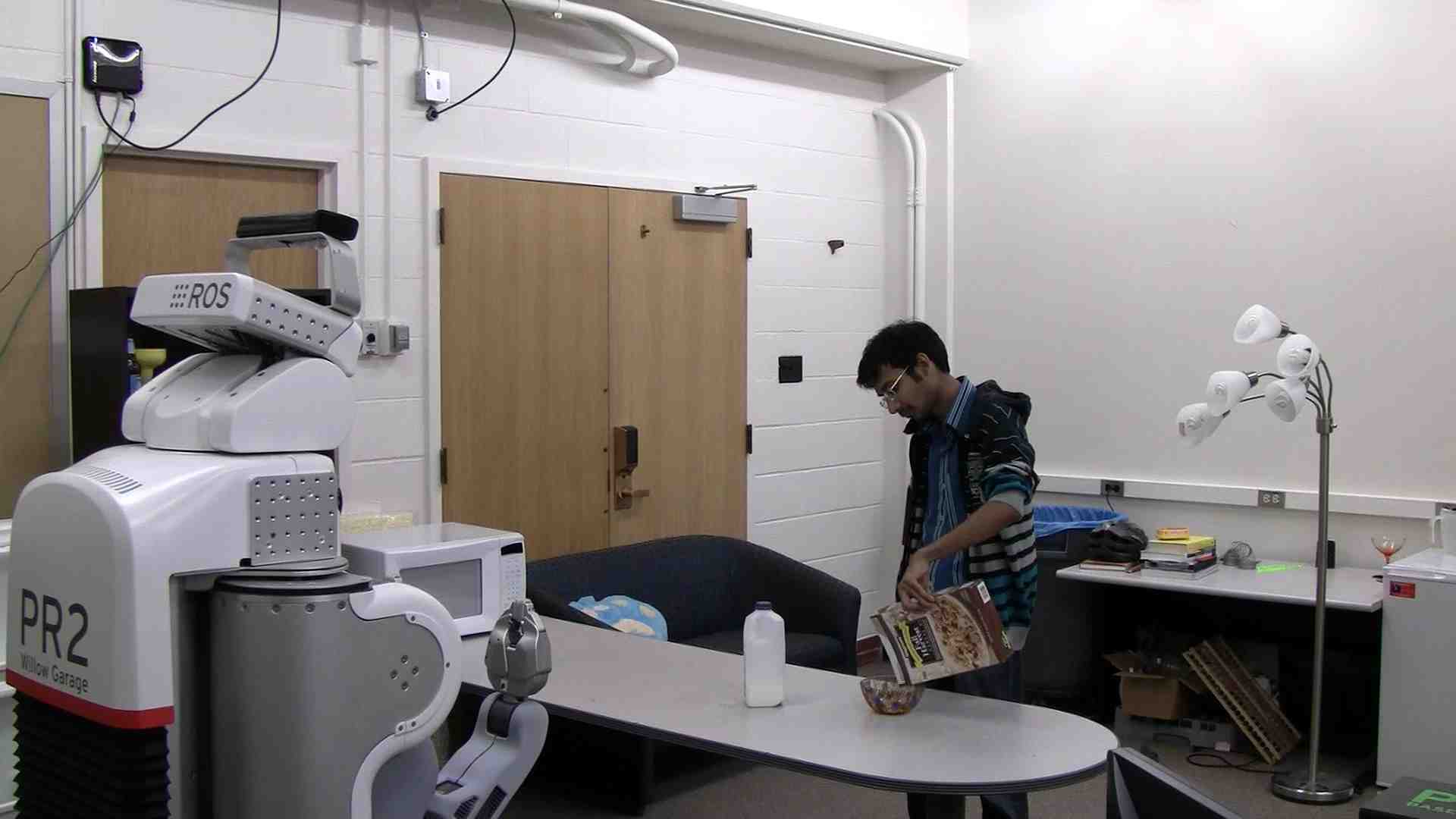}
\includegraphics[width=.49\linewidth,height=1in]{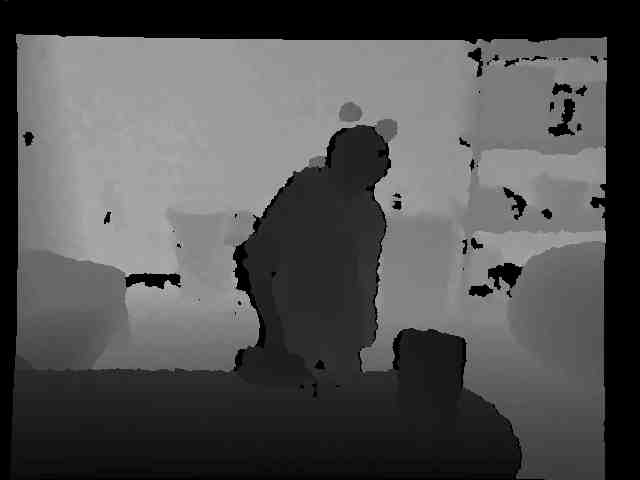}\\
\vskip 0.048in
\includegraphics[width=.49\linewidth,height=1in]{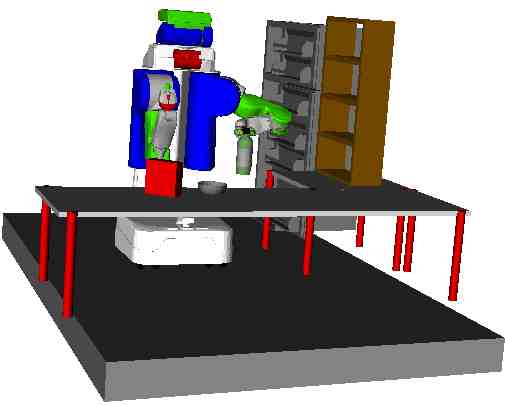}
\includegraphics[width=.49\linewidth,height=1in]{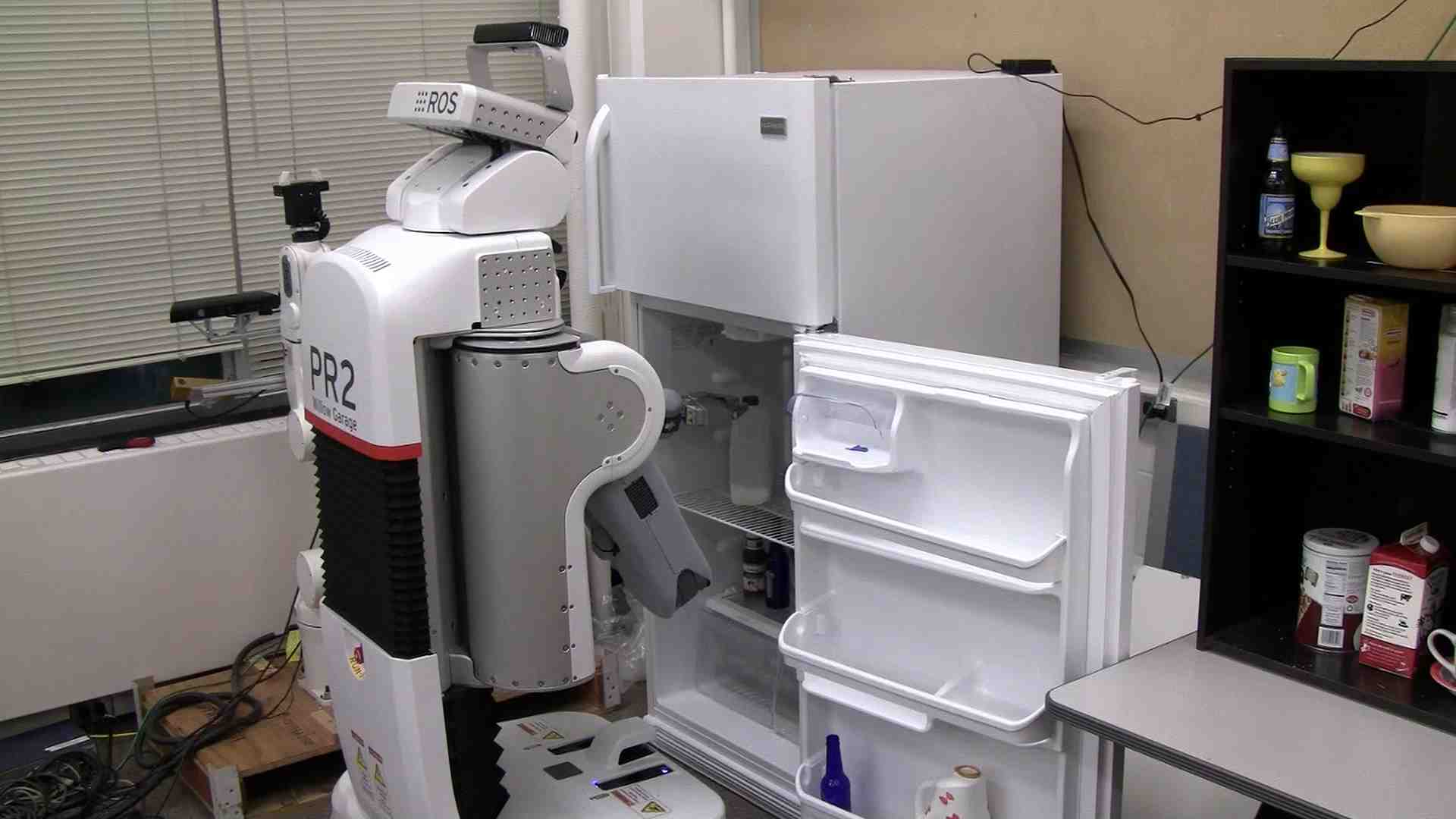}
\caption{ An assistive robot observes human activities (making cereal, top left).
Using RGB-D images (top right) as input,  our algorithm detects the activity being 
performed as well as the object affordances. 
This enables the robot to figure out how to interact with objects and plan actions (bottom left), 
and to respond appropriately to the activities being performed (cleaning up the table, bottom right).
}
 \label{fig:examplerobot}
\end{figure}

It is indispensable for a personal robot to perceive the environment in order to perform
assistive tasks.  Recent works in this area have addressed tasks such as estimating geometry \citep{HenryIJRR12},  tracking objects \citep{ChoiIJRR12}, recognizing objects \citep{ColletIJRR11}, 
placing objects \citep{jiang_ijrr2012} and labeling geometric classes \citep{koppula:Nips11,koppulaIJRR12}.  Beyond geometry and objects, humans are an important part of the indoor environments.
Building upon the recent advances in human pose detection from an RGB-D sensor \citep{ShottonCVPR2011}, in this paper we present learning algorithms to detect the human activities and object affordances. This information can then be used by assistive robots as shown in 
Fig.~\ref{fig:examplerobot}.

Most prior works in human activity detection have focussed on activity detection from still images
or from 2D videos. Estimating the human pose is the primary focus of these works, and they consider
 actions taking place over shorter time scales (see Section~\ref{sec:relatedwork}).
Having access to a 3D camera, which provides RGB-D videos, enables us to robustly 
 estimate human poses and use this information for learning complex human activities. 


 


\begin{figure*}[t!]
\centering
\includegraphics[width=.24\linewidth,height=0.9in]{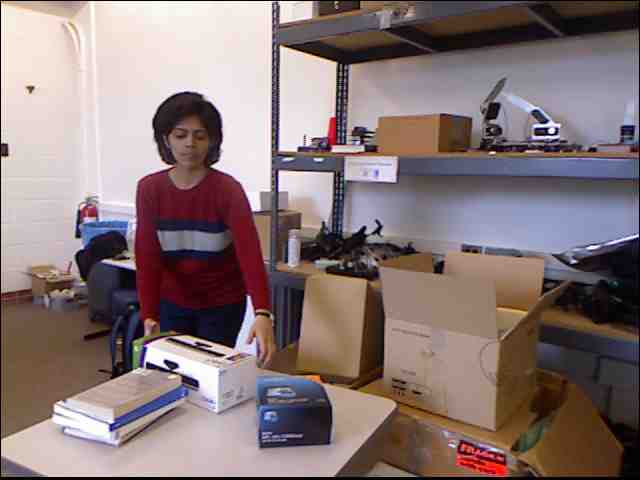}
\includegraphics[width=.24\linewidth,height=0.9in]{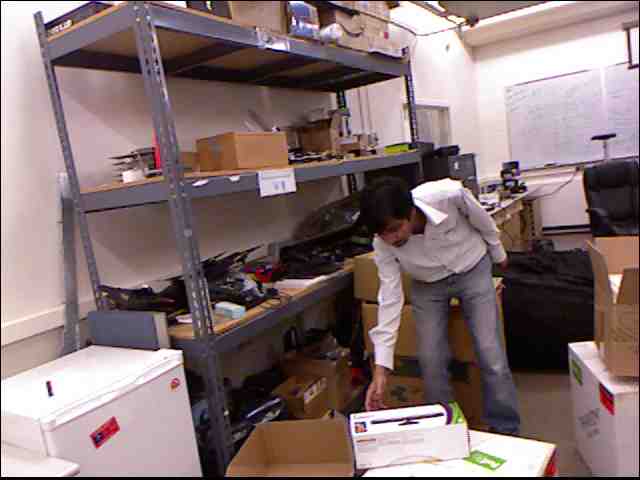}
\includegraphics[width=.24\linewidth,height=0.9in]{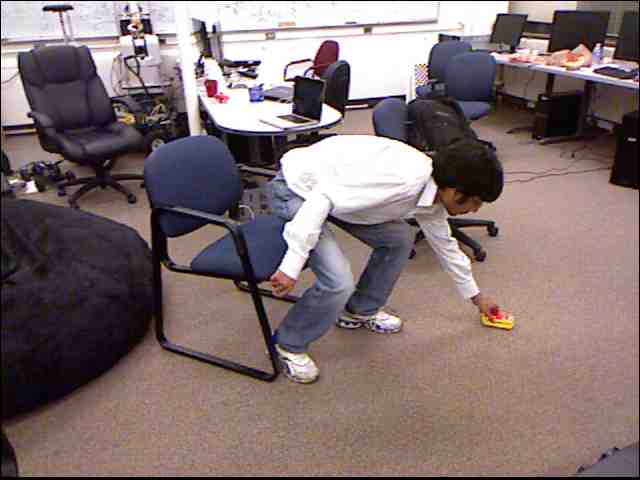}
\includegraphics[width=.24\linewidth,height=0.9in]{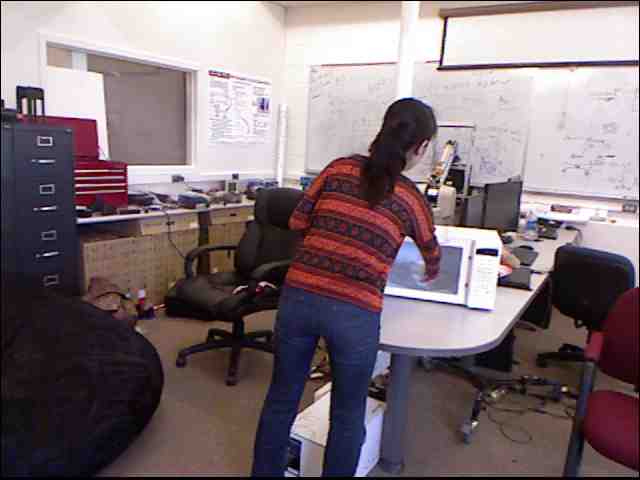}
\\ 
\vskip 0.048in
\includegraphics[width=.24\linewidth,height=0.9in]{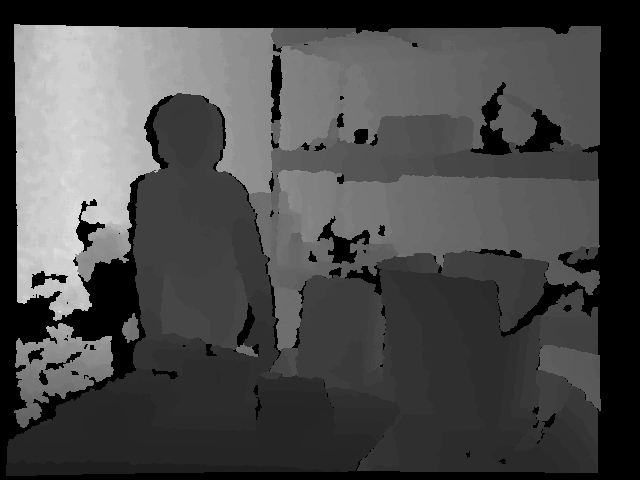}
\includegraphics[width=.24\linewidth,height=0.9in]{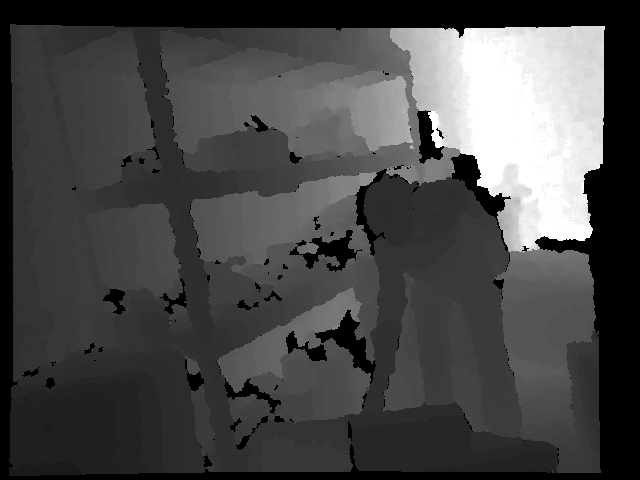}
\includegraphics[width=.24\linewidth,height=0.9in]{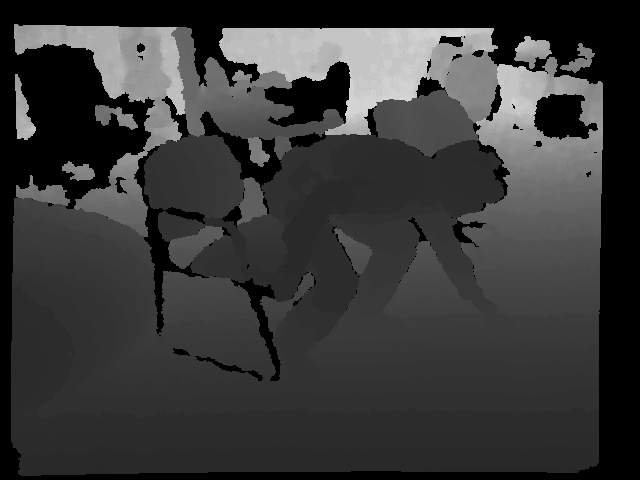}
\includegraphics[width=.24\linewidth,height=0.9in]{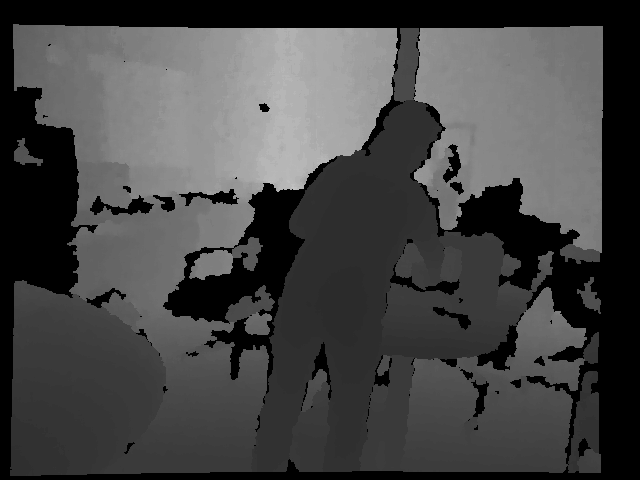}
 \\
\vskip 0.08in
\includegraphics[width=.24\linewidth,height=0.9in]{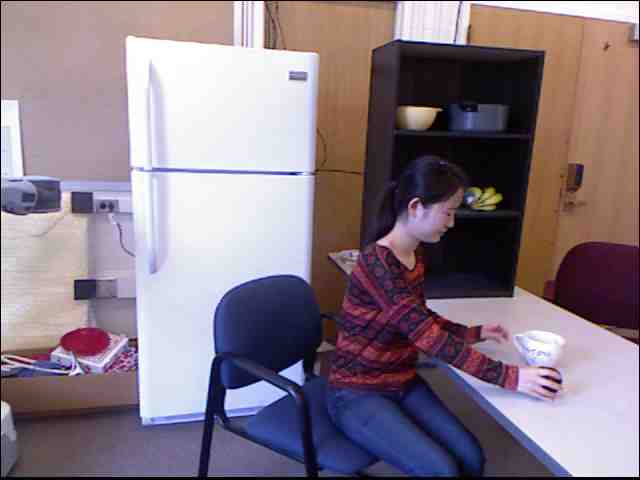}
\includegraphics[width=.24\linewidth,height=0.9in]{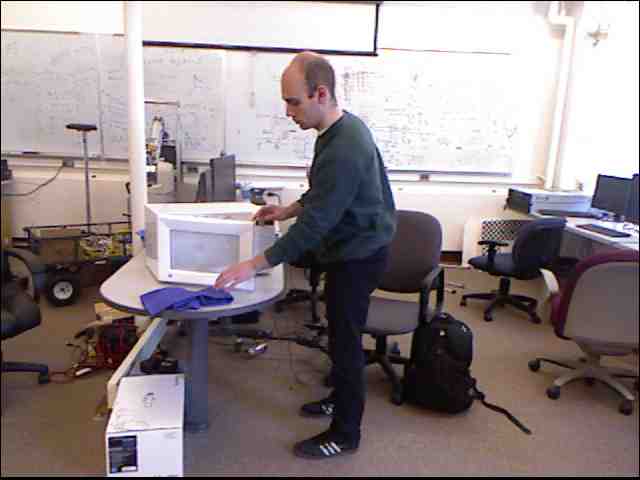}
\includegraphics[width=.24\linewidth,height=0.9in]{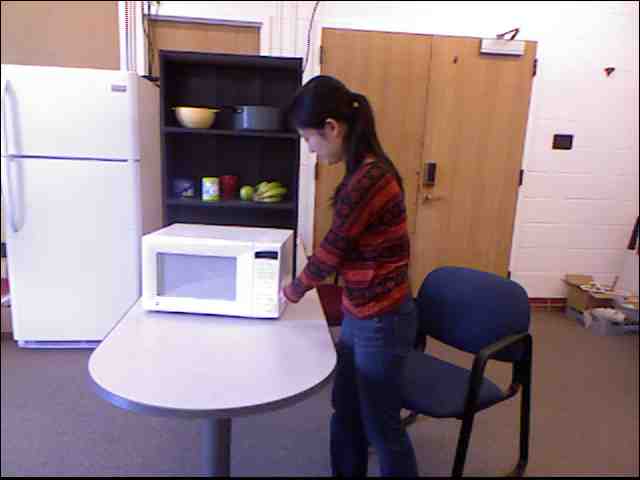}
\includegraphics[width=.24\linewidth,height=0.9in]{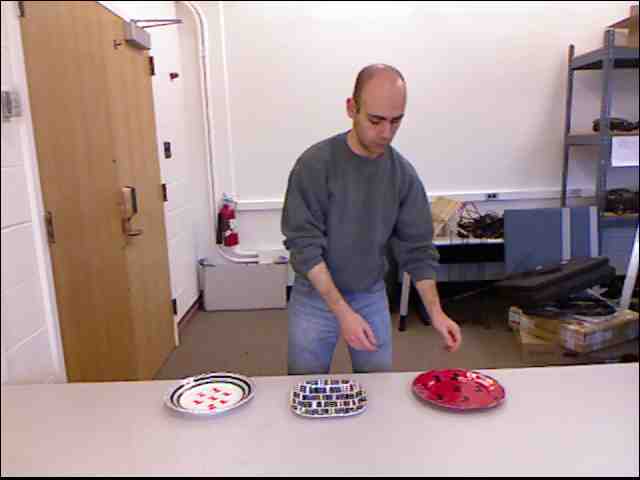}
\\
\vskip 0.048in
\includegraphics[width=.24\linewidth,height=0.9in]{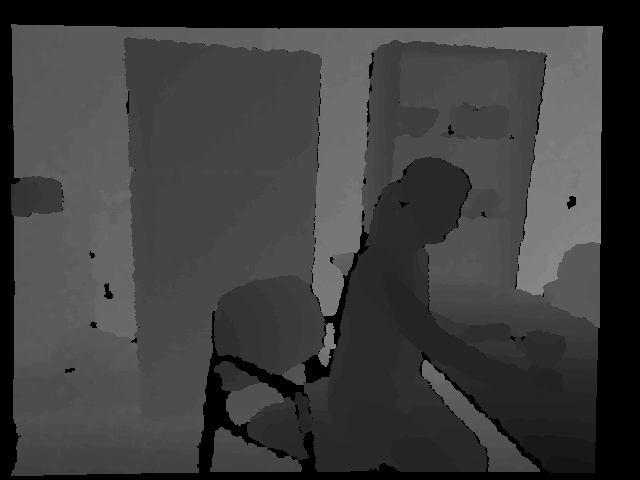}
\includegraphics[width=.24\linewidth,height=0.9in]{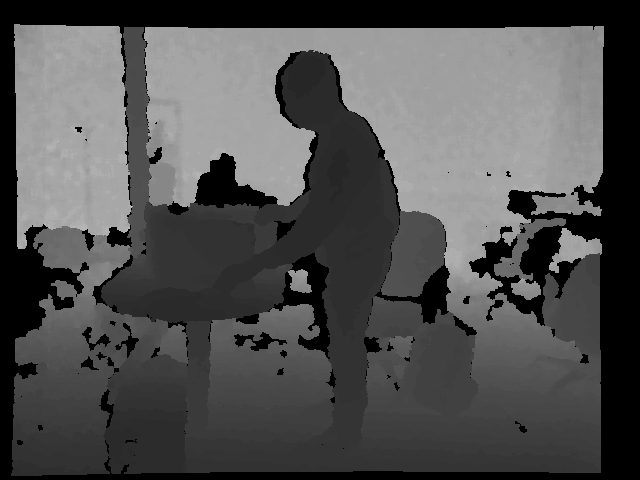}
\includegraphics[width=.24\linewidth,height=0.9in]{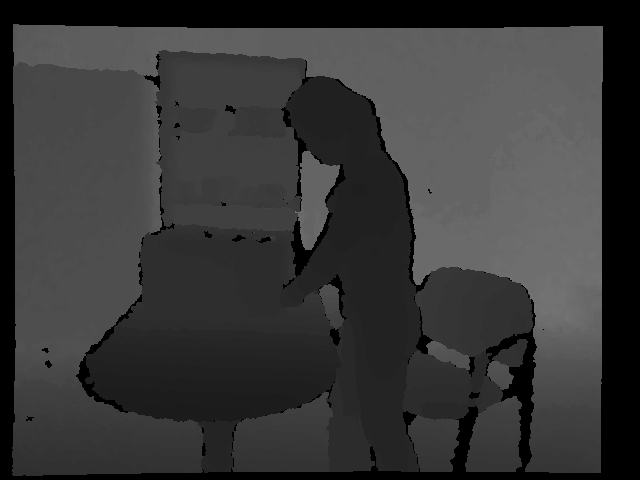}
\includegraphics[width=.24\linewidth,height=0.9in]{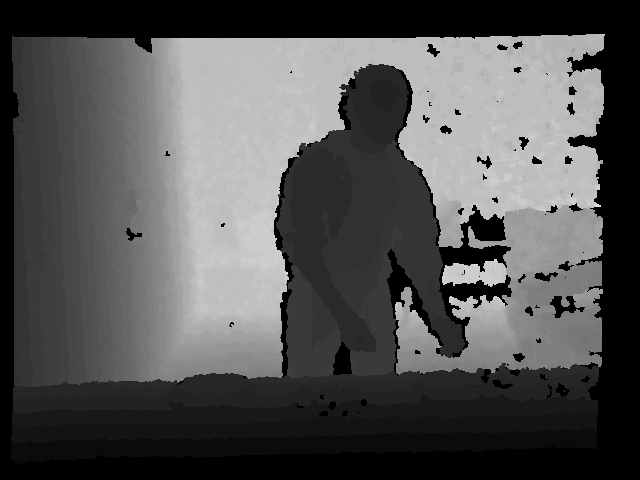}

\caption{{\bf Significant Variations, Clutter and Occlusions}:
 Example shots of \emph{reaching} sub-activity from our dataset. First and third rows show
the RGB images, 
and the second and bottom rows show the corresponding depth images from the RGB-D camera.
Note that there are significant variations in the way the subjects perform the sub-activity. In addition, 
there is significant background clutter and subjects are partially occluded (e.g., column 1) or not facing the camera (e.g., row 1 column 4) in many instances.
}
 \label{fig:reachingData}
\end{figure*}

Our focus in this work is to obtain a descriptive labeling of the complex human activities that take
 place over long time scales and consist of a \emph{long sequence of sub-activities}, 
 such as making cereal and arranging objects in a room (see Fig.~\ref{fig:labelingresult}).
For example, the making cereal activity consists of around 12 sub-activities on average, which includes
reaching the pitcher, moving the pitcher to the bowl, and then pouring the milk into the bowl. 
 This proves to be a very challenging task given the variability across individuals in performing each 
  sub-activity, and other environment induced conditions such as
  cluttered background and viewpoint changes. (See Fig.~\ref{fig:reachingData} for some examples.)

   In most previous works, object detection and activity recognition have been addressed as separate tasks. Only recently, some works have shown that modeling mutual context  is beneficial 
\citep{Gupta:TPAMI2009,Yao:CVPR10}. 
 The key idea in our work is to note that, in activity detection, it is sometimes more informative to 
 know \emph{how} an object is being used \citep[associated affordances,][]{Gibson:1979}
 rather than knowing \emph{what} the object is 
 (i.e. the object category). For example, both 
 chair and sofa might be categorized as `sittable,' and a cup might be categorized as both 
  `drinkable' and `pourable.'
Note that the affordances of an object change over time depending on its use, e.g.,
a pitcher may first be \emph{reachable}, then \emph{movable} and finally \emph{pourable}. 
In addition to helping activity recognition, recognizing object affordances is important by itself
because of their use in robotic applications \citep[e.g.,][]{pancake_robot,jiang2012humancontext,jiang2012placingobjects_context}.


We propose a method to learn human activities by modeling the sub-activities and affordances of the objects, 
how they change over time, and how they relate to each other.
More formally, we define a Markov random field (MRF)
 over two kinds of nodes: object and sub-activity nodes. The edges in the graph  model the pairwise relations among interacting nodes, namely the object--object interactions, object--sub-activity interactions, and the temporal interactions. This model is built with each
spatio-temporal segment being a node. The parameters of this model are learnt using a structural support vector machine (SSVM) formulation \citep{Finley/Joachims/08a}. 
Given a new sequence of frames, we label the high-level activity, all the sub-activities and the object affordances using our learned model.


 The activities take place over a long time scale, and different people execute sub-activities
 differently and for different periods of time. Furthermore, people also often merge two 
consecutive sub-activities together. 
Thus, segmentations in time are noisy and, in fact, there may not be one `correct' segmentation, especially
at the boundaries. 
 One approach could be to consider \textit{all} possible segmentations,
 and marginalize the segmentation; however, this is computationally infeasible. In this work, we 
perform sampling of several segmentations, and consider labelings over these temporal segments
as latent variables in our learning algorithm. 




We first demonstrate significant improvement over previous work on Cornell Activity Dataset (CAD-60).
We then contribute a new dataset comprising 120 videos collected from four subjects (CAD-120).
These datasets along with our code are available at  \url{http://pr.cs.cornell.edu/humanactivities/}.
In extensive experiments,
we show that our approach outperforms the baselines in both the tasks of activity as well as affordance detection. We achieved an accuracy of 91.8\% for affordance, 86.0\% for sub-activity labeling and 84.7\% for high-level activities respectively 
when given the ground truth segmentation, and an accuracy of 
79.4\%, 63.4\% and 75.0\%
on these respective tasks using our multiple segmentation algorithm.

In summary, our contributions in this paper are five-fold:
\begin{itemize}
\item
We provide a fully annotated RGB-D human activity dataset containing 120 long term activities such as
\emph{making cereal}, \emph{microwaving food}, etc. Each video is annotated with the human skeleton tracks, object tracks, object affordance labels, sub-activity labels, and high-level activities.   
\item
We propose a method for joint sub-activity and affordance labeling of RGB-D videos by modeling temporal and spatial interactions between humans and objects. 
\item
We address the problem of temporal segmentation by learning the optimal labeling from multiple temporal segmentation hypotheses. 
\item
We provide extensive analysis of our algorithms on two datasets and 
also demonstrate how our algorithm can be used 
by assistive robots. 
\item We release full source code along with ROS and PCL integration. 
\end{itemize}

The rest of the paper is organized as follows. 
We start with a review of the related work in Section \ref{sec:relatedwork}. We describe 
the overview of our methodology in Section \ref{sec:relations} and describe the model in Section
\ref{sec:model}. We then describe the object tracking and segmentation methods in Section 
\ref{sec:objtracking} and \ref{sec:temporalseg} respectively and describe the features used in our model in Section \ref{sec:features}.
 We present our learning, inference and temporal segmentation algorithms in Section \ref{sec:learninginference}. 
We present the experimental results along with robotic demonstrations in Section \ref{sec:experiments} 
and finally conclude the paper in Section \ref{sec:conclusion}.


 
 

 


\section{Related Work}\label{sec:relatedwork}

There is a lot of recent work in improving robotic perception in order 
to enable the robots to perform many useful tasks.
These works includes 3D modeling of indoor environments \citep{HenryIJRR12}, 
semantic labeling of environments by 
modeling objects and their relations to other objects in the scene \citep{koppula:Nips11,lai:icra11b,RosmanIJRR11,koppulaIJRR12},
developing frameworks for 
object recognition and pose estimation for manipulation \citep{ColletIJRR11}, object tracking for 3D object modeling \citep{KraininIJRR11}, etc.
Robots are now becoming more integrated in human environments and are being used in 
assistive tasks such as automatically interpreting and executing cooking recipes 
\citep{BolliniISER11}, robotic laundry folding \citep{MillerIJRR11} and
arranging a disorganized house \citep{jiang_ijrr2012,jiang2012placingobjects_context}.
Such applications makes it critical for the robots to understand both object affordances as well as human activities in order to work alongside with humans. We describe the recent advances in the various aspects of this problem here.   

\smallskip
\noindent
\textbf{Object affordances.}
An important accept of the human environment a robot needs to understand is the  object affordances. Most of the work within the robotics community related 
to affordances has focused on predicting opportunities for
interaction with an object either by using visual clues 
\citep{SunIJRR09,hermansSPME11,AldomaICRA12} or through observation of the effects of
exploratory behaviors \citep{MontesanoTR08,ridgeCVWW09,BogdanILP12}. 
For instance, \cite{SunIJRR09} proposed a probabilistic graphical
model that leverages visual object categorization for learning
affordances and \cite{hermansSPME11} proposed the use of physical and visual attributes as a 
mid-level representation for affordance prediction. \cite{AldomaICRA12} proposed a method 
to find affordances which depends solely on the objects of interest and their position and 
orientation in the scene. 
 These methods, do not consider the object affordances in human context, i.e. how the 
objects are usable by humans. 
We show that human-actor based affordances are essential for robots working 
in human spaces 
in order for them
to interact with objects in a human-desirable way.

There is some recent work in interpreting 
human actions and interaction with objects \citep{Lopes2005,saxena2008roboticgrasping,AksoyIJRR11,KonidarisIJRR12,li2011feccm} in
 context of learning to perform actions from demonstrations. 
 \cite{Lopes2005}  use  context from objects in terms of possible grasp affordances to focus the 
attention of their action recognition system and reduce ambiguities.
In contrast to these methods, we propose a model to learn human activities spanning over 
long durations and action-dependent affordances which make robots more capable in 
performing assistive tasks as we later describe in Section \ref{sec:robotapp}. 
\cite{saxena2008roboticgrasping} used supervised learning to detect grasp
affordances for grasping novel objects.
\cite{li2011feccm} used a cascaded classification model to model the interaction
between objects, geometry and depths. However, their work is limited to 
2D images.
In recent work, \cite{jiang2012humancontext} used a data-driven technique
for learning spatial affordance maps for objects.
 This work is different from ours in that we consider semantic affordances with spatio-temporal
grounding useful for activity detection.
Pandey and Alami \citeyearpar{PandeyIROS2010,Pandey2012} 
proposed mightability maps and taskability graphs that capture affordances such as
 reachability and visibility, while considering efforts required to be performed by the agents.
While this work manually defines object affordances in terms of kinematic and dynamic constraints,
 our approach learns them from observed data.




\smallskip
\noindent
\textbf{Human activity detection from 2D videos.}
There has been a lot of work on human activity detection from images \citep{YangWM10,Yao:ICCV11} 
and from videos \citep{action-Laptev-cvpr08,action-jingen-wild,HoaiLD11,ShiIJCV11,model-recommendation-activity,firstpersonactivity,fine-grained-activity,actionbank,complex-event-detection}. 
Here, we discuss works that are closely related to ours, and 
refer the reader to \cite{survey} for a survey of the field.
Most works \citep[e.g.][]{HoaiLD11,ShiIJCV11,model-recommendation-activity} consider detecting actions at a `sub-activity' level (e.g. \emph{walk}, \emph{bend}, and \emph{draw}) instead of considering high-level activities. Their methods
range from discriminative learning techniques for joint segmentation and recognition \citep{ShiIJCV11,HoaiLD11} to combining multiple models \citep{model-recommendation-activity}.
Some works, such as \cite{complex-event-detection}, consider high-level activities. 
\cite{complex-event-detection} proposed a latent model for high-level activity classification and have the advantage of requiring only high-level activity labels for learning. 
None of these methods explicitly consider the role of objects or object affordances that not only
help in identifying sub-activities and high-level activities, but are also 
important for several robotic applications \citep[e.g.][]{pancake_robot}. 
    
  Some recent works \citep{Gupta:TPAMI2009,Yao:CVPR10,conf/icra/AksoyAWD10,tcsvt:actioncontext,firstpersonactivity} show that modeling the interaction 
 between human poses and objects in 2D videos 
results in a better performance on the tasks of object detection and activity recognition. 
However, these works cannot capture the rich 3D relations between the activities and objects, and 
are also fundamentally limited by the quality of the human pose inferred from the 2D data.
More importantly, for activity recognition, the object \emph{affordance} matters more than
its category.

 \begin{figure*}[t]
 \centering
 \includegraphics[width=\linewidth, height=2.5in]{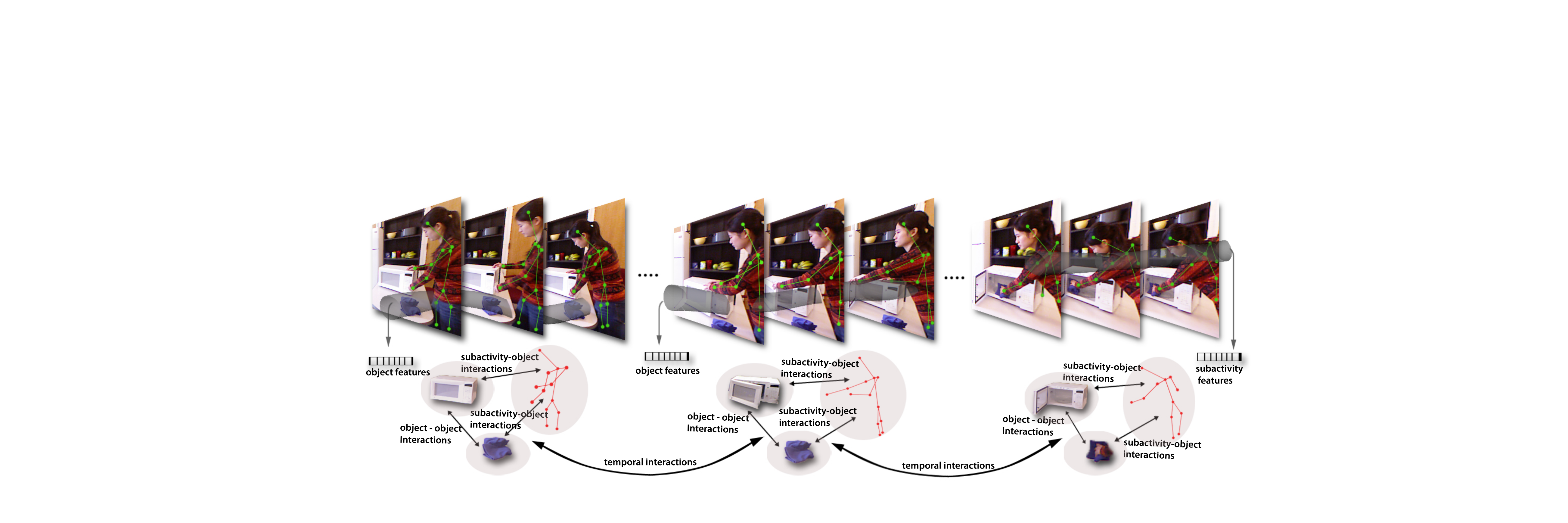}
 \caption{Pictorial representation of the different types of nodes and relationships modeled in part of the \emph{cleaning objects} activity comprising three sub-activities: \emph{reaching}, \emph{opening} and \emph{scrubbing}. (See Section \ref{sec:relations}.)}
 \label{fig:modelfigure}
 \end{figure*}


\cite{Kjellstrom:2011} 
used a factorial CRF to 
simultaneously segment and classify human hand actions, as well as classify the object affordances involved in the activity from 2D videos. However, this work is limited to classifying only hand actions and does not model interactions between the objects. We consider complex full-body activities and show that modeling object--object interactions is important as objects have affordances even if they are not directly interacted with human hands. 



\smallskip
\noindent
\textbf{Human activity detection from RGB-D videos.}
Recently, with the availability of inexpensive RGB-D sensors, some works  \citep{LiCVPR4HB10,NWM2011,SungAAAIPair2011,ZhangIROS11,SungICRA2012} consider
 detecting
 human activities from RGB-D videos.
 \cite{LiCVPR4HB10} proposed an expandable graphical model, to model the temporal dynamics of actions and use a bag of 3D points to model postures. They use their method to classify 20 different actions which are used in context of interacting with a game console, such as draw tick, draw circle, hand clap, etc. \cite{ZhangIROS11} designed 4D local spatio-temporal features and use an LDA classifier to identify six human actions such as lifting, removing, waving, etc., from a sequence of RGB-D images. However, both these works only address detecting actions which span short time periods. \cite{NWM2011} also designed feature representations such as spatio-temporal interest points and motion history images which incorporate depth information in order to achieve better recognition performance.  
 \cite{PanangadanIJSR10} used data from laser rangefinder to model observed movement patterns and interactions between persons. They segment tracks into activities based on difference in displacement distributions in each segment, and use a Markov model for capturing the transition probabilities.
  None of these works model interactions with objects which provide useful information for recognizing complex activities.



In recent previous work from our lab, \cite{SungAAAIPair2011,SungICRA2012} proposed a hierarchical maximum entropy Markov model to detect activities from RGB-D videos and treat the sub-activities as hidden nodes in their model. 
However, they use only human pose information for detecting activities and also constrain the number of sub-activities in each activity. In contrast, we model context from object interactions along with human pose,
and also 
 present a better learning algorithm. (See Section~\ref{sec:experiments} for further comparisons.)
\cite{GallCVPR11} also use depth data to perform sub-activity (referred to as action) classification and functional categorization of objects. 
Their method first detects the sub-activity being performed using the estimated human pose from depth data, and then 
performs object localization and clustering of the objects into functional categories based on the detected sub-activity. 
In contrast, our proposed method performs joint sub-activity and affordance labeling and uses these labels to 
perform high-level activity detection.

All of the above works lack a unified framework that combines all of the information available in human interaction activities and therefore we propose a model that  captures both the spatial and temporal relations between object affordances and human poses to perform joint object affordance and activity detection.

\section{Overview}
\label{sec:relations}

Over the course of a video, a human may interact with several objects and perform
several sub-activities over time. In this section we describe at a high level how we process the RGB-D
videos and model the various properties for affordance and activity labeling. 

Given the raw data containing the color and depth values for every pixel in the video, 
we first track the human skeleton using Openni's skeleton tracker\footnote{http://openni.org}
 for obtaining the locations of the various joints of the human skeleton. 
 However these values are not very accurate, as the Openni's skeleton tracker is only 
 designed to track human skeletons in clutter-free  environments and without any 
 occlusion of the body parts. 
In real-world human activity videos, some body parts are often occluded 
 and the interaction with the objects hinders accurate skeleton tracking. We show that 
  even with such noisy data, our method gets high accuracies by modeling
  the mutual context between the affordances and sub-activities.
  

We then segment 
the object being used in the activity and track them through out the 3D video,
 as explained in detail in Section \ref{sec:objtracking}. 
 We model the activities and affordances by defining a MRF
over the spatio-temporal sequence we get from an RGB-D video, as illustrated in Fig.~\ref{fig:modelfigure}. MRFs are a workhorse of machine learning, and have been successfully
applied to many applications \cite[e.g.][]{saxena-make3d-pami}. Please see \cite{DaphneBook} for a review.
 If we build our graph with nodes for objects and sub-activities for each time instant (at 30 fps), then we
will end up with quite a large graph. Furthermore, such a graph would not be able to model
meaningful transitions between the sub-activities because they take place over a long-time (e.g. a few seconds). 
Therefore, in our approach we first segment the video into small temporal segments, and our goal
is to label each segment with appropriate labels. 
We try to over-segment, so that we end up
with more segments and avoid merging two sub-activities into one segment.
Each of these segments occupies a small length of time and therefore, considering nodes per segment 
gives us a meaningful and concise representation for the graph $\mathcal{G}$.  
With such a representation, we can model meaningful transitions of a sub-activity following another, e.g. \emph{pouring}  followed by \emph{moving}. 
 Our temporal segmentation algorithms are described in Section \ref{sec:temporalseg}. 
 The outputs from the skeleton and object tracking 
 along with the segmentation information and RGB-D videos are then used to generate the 
 features described in Section \ref{sec:features}.

 Given the tracks and segmentation, the graph structure ($\mathcal{G}$) is constructed 
 with a node for each object and a node for the sub-activity of a temporal segment,
 as shown in Fig.~\ref{fig:modelfigure}.
 The nodes are connected to each other 
 within a temporal segment and each node is connected to its temporal neighbors by edges as further described in 
 Section \ref{sec:model}. The learning and inference algorithms for our model are presented in Section \ref{sec:learninginference}.
 We capture the following properties in our model:


\begin{itemize}

\item
{\bf Affordance--sub-activity relations.} At any given time, the affordance of the 
object depends on the sub-activity it is involved in. For example, a cup has the affordance 
of \emph{`pour-to'} in a \emph{pouring} sub-action and has the affordance of \emph{`drinkable'} 
in a \emph{drinking} sub-action. 
We compute relative geometric features between the object and 
the human's skeletal joints to capture this. These features are incorporated in the energy 
function as described in Eq.~(\ref{eq:objectactivity}).

 \item 
{\bf Affordance--affordance relations.}  
 Objects have affordances even if they are not interacted directly with by the human,
 and their affordances depend on the affordances of other objects around them.
For example, in the case of \emph{pouring} from a pitcher to a cup, the cup is not 
interacted with by the human directly but has the affordance \emph{`pour-to'}. 
We therefore use relative geometric features such as `on top of', `nearby', `in front of', etc., 
to model the affordance--affordance relations. These features are incorporated in the energy 
function as described in Eq.~(\ref{eq:objectobject}).

\item 
{\bf Sub-activity change over time.} Each activity consists of a sequence of sub-activities that
change over the course of performing the activity. We model this by incorporating 
temporal edges in $\mathcal{G}$. Features capturing the change in human pose across 
the temporal segments are used to model the sub-activity change over time and the corresponding
 energy term is given in Eq.~(\ref{eq:activitytemporal}).

\item
{\bf Affordance change over time.} The object affordances depend on the sub-activity being 
performed and hence change along with the sub-activity over time. We model the temporal 
change in affordances of each object using features such as change in appearance 
and location of the object over time. These features are incorporated in the energy 
function as described in Eq.~(\ref{eq:objecttemporal}).

\end{itemize}

\begin{figure}
\centering
\includegraphics[width=\linewidth]{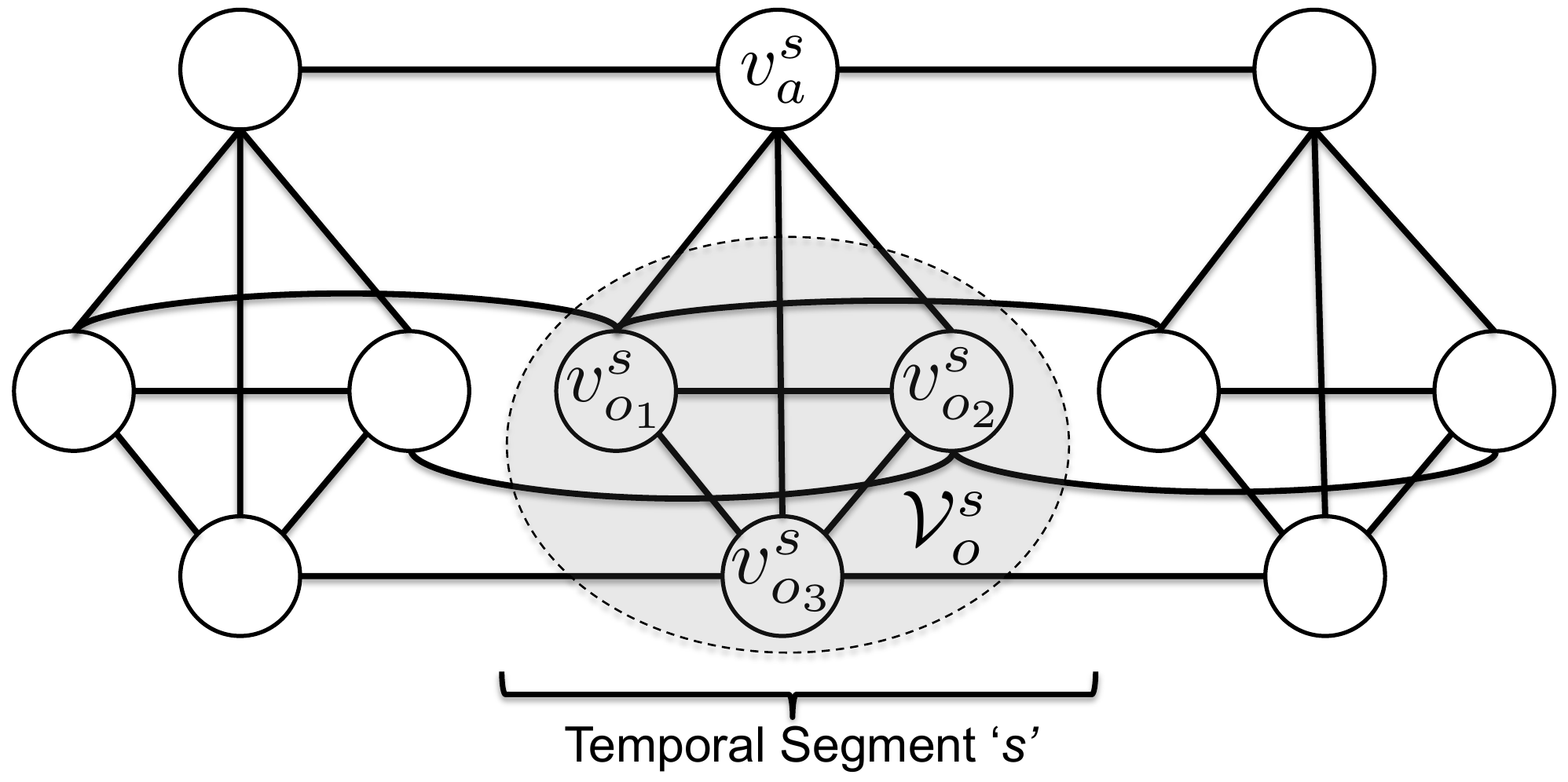}
\caption{An illustrative example of our Markov random field (MRF) for three temporal segments, with one activity node, $v^s_a$,  and three object nodes, \{$v^s_{o_1}$,$v^s_{o_2}$,$v^s_{o_3}$\},  per temporal segment.}
\label{fig:mrf}
\end{figure}

\section{Model}
\label{sec:model}
Our goal is to perform joint activity and object affordance labeling of RGB-D videos.
We model the spatio-temporal structure of an activity using a model isomorphic to a MRF with log-linear node and pairwise edge potentials (see Fig.~\ref{fig:modelfigure} 
and~\ref{fig:mrf} for an illustration). 
The MRF is represented as a graph $\mathcal{G} = (\mathcal{V},\mathcal{E})$. 
Given a temporally segmented 
3D video, with temporal segments $s  \in \{1,...,N\}$, 
our goal is to predict a labeling 
$\y=(\ys{1},...,\ys{N})$ for the video, where $\ys{s}$ is the set of sub-activity and object affordance 
labels for the temporal segment $s$.
Our input is a set of features $\x$ extracted from the segmented 3D video as described 
in Section \ref{sec:features}.
The prediction $\hat{\y}$ is computed as the argmax of a energy function $\df{\x}{\y}{\w}$ that is parameterized by weights $\w$.
\begin{align} 
\hat{\y} & = \argmax_\y \df{\x}{\y}{\w} \label{eq:argmax} \\
 \df{\x}{\y}{\w} & = E_o + E_a + E_{oo} + E_{oa} + E^t_{oo} + E^t_{aa}  
\end{align}   
The energy function consists of six types of potentials that define the energy of 	a particular 
assignment of sub-activity and object affordance labels to the sequence of segments in the 
given video. The various potentials capture the dependencies between the sub-activity and 
object affordance labels as defined by an undirected graph $\mathcal{G}  = (\mathcal{V},\mathcal{E})$.

We now describe the structure of this graph along with the corresponding potentials. 
There are two types of nodes in $\mathcal{G}$: object nodes denoted by $\mathcal{V}_o$ and sub-activity nodes denoted by $\mathcal{V}_a$. 
Let $K_a$ denote the set of sub-activity labels, and $K_o$ denote the set of object affordance labels.  
Let $\ysc{i}{k}$ be a binary variable representing the node $i$ having label $k$, where $k \in K_o$ for object nodes and $k \in K_a$ for sub-activity nodes. All $k$ binary variables together represent the label of a node.
Let $\mathcal{V}_o^s$ denote set of object nodes of segment $s$, and $v_a^s$ denote the sub-activity node of segment $s$.  Figure \ref{fig:mrf} shows the graph structure for three temporal segments for an activity with three objects.

The energy term associated with labeling the object nodes is denoted by $E_o$ and is defined as the sum of object node potentials $\psi_o(i)$ as: 

{\small
\begin{equation}
E_o = \sum_{i \in \mathcal{V}_o} \psi_o(i) = \sum_{i \in \mathcal{V}_o} \sum_{k \in K_o}  \ysc{i}{k} \left[\wo{k} \cdot \fo{i} \right] ,
\end{equation}
}

\noindent
where $\fo{i}$ denotes the feature map describing the object affordance of the object node $i$ in its corresponding temporal segment through a vector of features, and there is one weight vector for each affordance class in $K_o$. 
Similarly, we have an energy term, $E_a$, for labeling the sub-activity nodes which is defined as the sum of 
sub-activity node potentials as 

{\small
\begin{equation}
E_a = \sum_{i \in \mathcal{V}_a} \psi_a(i) = \sum_{i \in \mathcal{V}_a}  \sum_{k \in K_a} \ysc{i}{k} \left[\ws{k} \cdot \fs{i} \right] ,
\end{equation}
}

\noindent
where $\fs{i}$ denotes the feature map describing the sub-activity node $i$ through a vector of features, and there is one weight vector for each sub-activity class in $K_a$. 

 For all segments $s$,  there is an edge connecting all the nodes in $\mathcal{V}_o^s$ to each other, denoted by $\mathcal{E}_{oo}$, and to the sub-activity node  $v_a^s$, denoted by $\mathcal{E}_{oa}$. These edges signify the relationships within the objects, and between the objects and the human pose within a segment and are referred to as \emph{`object--object interactions'} and 
\emph{`sub-activity--object interactions'}  in the Fig.~\ref{fig:modelfigure}, respectively. 

The sub-activity node of segment $s$ is connected to the sub-activity nodes in segments $(s-1)$ and 
$(s+1)$. These temporal edges are denoted by $\mathcal{E}^t_{aa}$. Similarly every object node of segment $s$ is connected to the corresponding object nodes in segments $(s-1)$ and 
$(s+1)$, denoted by $\mathcal{E}^t_{oo}$. These edges model the \emph{temporal interactions} between the human poses and the objects, respectively, and are represented by dotted edges in the Fig.~\ref{fig:modelfigure}. 

We have one energy term for each of the four interaction types and are defined as:

{\small
\begin{align}
E_{oo} & =  \sum_{(i,j)\in \mathcal{E}_{oo}}   \sum_{(l,k) \in K_o \times K_o}  \ysc{i}{l} \ysc{j}{k}  \left[\we{oo}{l}{k} \cdot \fe{oo}{i}{j}\right] ,\label{eq:objectobject}\\
E_{oa} & =  \sum_{(i,j)\in \mathcal{E}_{oa}}   \sum_{(l,k) \in K_o \times K_a}  \ysc{i}{l} \ysc{j}{k}  \left[\we{oa}{l}{k} \cdot \fe{oa}{i}{j}\right] , \label{eq:objectactivity}\\
E^t_{oo} & =  \sum_{(i,j)\in \mathcal{E}^t_{oo}}   \sum_{(l,k) \in K_o \times K_o}  \ysc{i}{l} \ysc{j}{k}  \left[\wte{oo}{l}{k} \cdot \fte{oo}{i}{j}\right] , \label{eq:objecttemporal}\\
E^t_{aa} & =  \sum_{(i,j)\in \mathcal{E}^t_{aa}}   \sum_{(l,k) \in K_a \times K_a}  \ysc{i}{l} \ysc{j}{k}  \left[\wte{aa}{l}{k} \cdot \fte{aa}{i}{j}\right] .\label{eq:activitytemporal}
\end{align}
}

The feature maps $\fe{oo}{i}{j}$ and $\fe{oa}{i}{j}$ describe the interactions between pair of objects and interactions between an object and the human skeleton within a temporal segment, respectively, and the feature maps $\fte{oo}{i}{j}$ and $\fte{aa}{i}{j}$ describe the temporal interactions between objects and sub-activities, respectively. Also, note that there is one weight vector for every pair of labels in each energy term. 

 Given $\mathcal{G}$, we can rewrite the energy function based on individual node potentials and edge potentials compactly as follows:

{\small
\begin{align} 
\df{\x}{\y}{\w}  & =   \sum_{i \in \mathcal{V}_a}  \sum_{k \in K_a} \ysc{i}{k} \left[\ws{k} \cdot \fs{i} \right] + 
 \sum_{i \in \mathcal{V}_o} \sum_{k \in K_o}  \ysc{i}{k} \left[\wo{k} \cdot \fo{i} \right] \nonumber \\
& +  \sum_{ t \in {\cal T}} \sum_{(i,j)\in \mathcal{E}_t}   \sum_{(l,k)\in T_t} \ysc{i}{l} \ysc{j}{k}  \left[\we{t}{l}{k} \cdot \fe{t}{i}{j}\right]  
\label{eq:model}
\end{align}
}

\noindent
where $\cal T$ is the set of the four edge types described above. 
Writing the energy function in this form allows us to 
apply efficient inference and learning algorithms as described later in Section \ref{sec:learninginference}.

\section{Object Detection and Tracking}
\label{sec:objtracking}



For producing our graph $\mathcal{G}$, we need as input the segments corresponding to 
the objects (but not their labels) and their tracks over time.
In order to do so,
we run pre-trained object 
detectors on a set of frames sampled from the video and then use particle filter 
tracker to obtain tracks of the detected objects. We then find consistent tracks that 
connect the various detected objects in order to find reliable object tracks. We present the
details below.

  
\smallskip
\noindent  
{\bf Object Detection:}  We first train a set of 2D object detectors for the common objects
present in our dataset (e.g. mugs, bowls). We use features that capture the inherent 
local and global properties of the object encompassing the appearance and the 
shape/geometry. Local features includes color histogram and the histogram of oriented 
gradients (HoG, \citeauthor{dalal2005histograms}, \citeyear{dalal2005histograms}) which provide the intrinsic properties of the target object while viewpoint 
features histogram (VFH, \citeauthor{RusuVFH}, \citeyear{RusuVFH}) provides the global orientation of the normals from the object's 
surface.
For training we used the RGB-D object dataset by \cite{lai:icra11a} and built a one-vs-all SVM classifier using the local features for each object class in order to obtain the probability 
estimates of each object class. We also build a k-nearest neighbor (kNN) classifier over VFH features. 
The kNN classifier provides the detection score as inverse of the distance between training and the test instance. We obtain a final classification score by adding the scores from these two classifiers. 

At test time, for a given point cloud, we 
first reduce the set of 3D bounding boxes by only considering those that belong
to a volume around the hands of the skeleton.
This reduces the number of 
false detections as well as the detection time. We then run our SVM-based object detectors 
on the RGB image. This gives us the exact $x$ and $y$ coordinates of the possible 
detections. The predictions with score above a certain threshold are further examined by 
calculating the kNN score based on VFH features. 
To find the exact depth of the object, we do pyramidal window search inside 
the current 3D bounding box and get the highest scoring box. In order to remove the empty 
space and any outlier points within a bounding box, we shrink it towards the highest-density 
region that captures 90\% of the object points. These bounding box detections are ordered according to 
their final classification score.

\smallskip
\noindent
{\bf Object Tracking:}
We used the particle filter tracker implementation\footnote{http://www.willowgarage.com/blog/2012/01/17/tracking-3d-objects-point-cloud-library} provided under the PCL library for tracking our target object. The tracker uses the color values and the normals to find the next probable state of the object.

\smallskip
\noindent
{\bf Combining Object Detections with Tracking:}
We take the top detections, track them, and assign a score to each of them in order to 
compute the potential nodes in the graph $\mathcal{G}$. 

We start with building a graph with the initial bounding boxes as the nodes. In our current implementation, this method needs an initial guess on the 2D bounding boxes of the objects to keep the algorithm tractable. We can obtain this by considering only the tabletop objects by using a tabletop object segmenter \cite[e.g.][]{Rusu2009tabletop}.
 We initialize the graph with these guesses.
 We then perform tracking through the video and grow the graph by adding a node for every object detection and connect two nodes with an edge 
if a track exists between their corresponding bounding boxes. 
Our object detection algorithm is run after every fixed number of frames, 
and the frames on which it is run are referred to as the detection frames. 
Each edge is assigned a weight corresponding to its track score as defined in Eq.~(\ref{eq:trackscore}). After
the whole video is processed, the best track for every initial node in the graph is found by taking the highest weighted path starting at that node.

\begin{figure}[t!]
\centering
\includegraphics[width=\linewidth,height=1.5in]{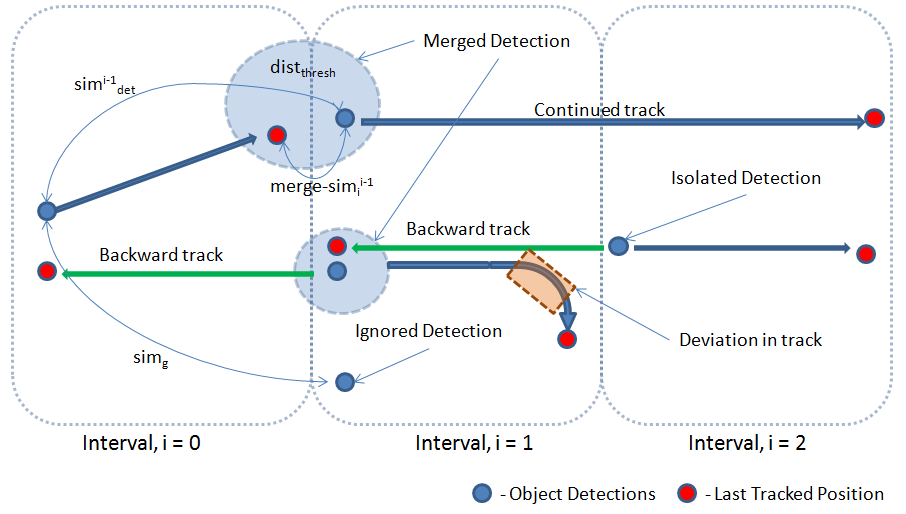}
\caption{Pictorial representation of our algorithm for combining object detections with tracking.}
 \label{fig:tracking}
\end{figure}

The object detections at the current frame 
are categorized into one of the following categories: 
\{\emph{merged detection}, \emph{isolated detection}, \emph{ignored detection}\}
based on their vicinity and similarity to the currently tracked objects as shown in Figure \ref{fig:tracking}.
If a detection occurs close to a currently tracked object and has high similarity with it, the 
detection can be merged with the current object track. Such detections are called \emph{merged detections}. 
The detections which have high detection score but do not occur close to the current tracks are 
labeled as \emph{isolated detections} and are tracked in both directions. This helps in correcting the tracks which have gone wrong due to partial occlusions or missing due to full occlusions of the objects.
The rest of the detections are labeled as \emph{ignored detections} and are not tracked.





More formally, let $d^i_j$ represent the bounding box of the $j^{th}$ detection in the $i^{th}$ detection frame and let $D^i_j$ represent its tracking score. 
 Let $\hat{d}^i_j$ represent the tracked bounding box at the current frame with the track starting from $d^i_j$. 
 We define a similarity score, $S(a,b)$, between two image bounding boxes, $a$ and $b$, as the
 correlation score of the color histograms of the two bounding boxes.  
The track score of an edge $e$ connecting the detections $d^{i-1}_k$ and $d^i_j$ is given by
\begin{equation}
 ts(e) = S(d^{i-1}_k,d^i_j)  + S(\hat{d}^{i-1}_k,d^i_j)  + \lambda  D^i_j
 \label{eq:trackscore}
\end{equation}
\noindent

Finally, the best track of a given object bounding box $b$ is the path having the highest cumulative track score from all paths originating at node corresponding to $b$ in the graph, represented by the set $P_b$
\begin{equation}
 \hat{t}(b) = \argmax_{p \in P_b} \sum\limits_{e \in p } ts(e).
\end{equation}

\section{Temporal Segmentation} 
\label{sec:temporalseg}
We perform temporal segmentation of the frames in an activity in order to obtain groups of frames representing atomic 
movements of the human skeleton and objects in the activity. This will group similar frames into one segment, thus reducing the 
total number of nodes to be considered by the learning algorithm significantly.

There are several problems with naively performing one temporal segmentation---if we make a mistake
here, 
then our followup activity detection would perform poorly. In certain cases, 
when the 
features are additive,
 methods based on dynamic programming  
\citep{HoaiLD11,ShiIJCV11,HoaiAISTATS12} could be used to search for an
optimal segmentation.

However, in our case, we have the following three challenges. First, the feature maps we consider
are non-additive in nature, and the feature computation cost is exponential in the number of frames
 if we want to consider all the possible segmentations.
Therefore, we cannot apply dynamic programming techniques to find the optimal segmentation.
Second, the complex human-object interactions 
are poorly approximated with
a linear dynamical system, therefore techniques such as \cite{FoxTSP2011} cannot be directly applied.
Third, the boundary between two sub-activities is often not very clear, as people often
start performing the next sub-activity before finishing the current sub-activity.
The amount of overlap might also depend on which sub-activities are being performed. 
Therefore, there may not be one optimal segmentation.
In our work, we consider 
several temporal segmentations and propose a method to combine 
them in Section \ref{sec:segmentation}.

We consider three basic methods for temporal segmentation of the video frames and 
generate a number of temporal segmentations by varying the parameters of these 
methods. The first method is uniform segmentation, in which we consider a set of 
continuous frames of fixed size as the temporal segment. There are two parameters 
for this method: the segment size and the offset (the size of the first segment). 
The other two segmentation methods use the graph-based segmentation proposed by 
\cite{Felzenszwalb:2004} adapted to temporally segment the videos. 
The second method uses the sum of the 
Euclidean
distances between the 
skeleton joints as the edge weights, whereas the third method uses the 
rate of change of the Euclidean distance as the edge weights. 
These methods consider smooth movements of the skeleton joints to belong to one segment and 
identify sudden changes in skeletal motion 
as the sub-activity boundaries.

In detail, we have one node per frame representing the skeleton in the graph based methods. Each node is connected to its temporal neighbor, therefore giving a chain graph.
The algorithm begins with having each node as a separate segmentation,
and iteratively merges the components if the edge weight is less than a certain
threshold (computed based on the current segment size and a constant parameter). We obtain different 
segmentations by varying the parameter.%
\footnote{Details: In order to handle occlusions, we only use the upper 
body skeleton joints for computing the edge weights
that are  estimated more reliably by the skeleton tracker. 
When changing the parameters for the three segmentation methods for obtaining multiple segmentations,
we select the parameters such that we always err on the side of over-segmentation instead
of under-segmentation.
This is because our learning model can handle over-segmentation by assigning the same label
to the consecutive segments for the same sub-activity, but under-segmentation is
is bad as the  model can only assign one label to that segment.}

 \section{Features}
 \label{sec:features}

\begin{table}[t!]
\caption{Summary of the Features used in the Energy Function.}

\begin{tabular}{|p{6.0cm}|c|} 
\hline
 Description  & Count\\ \hline
{\bf Object Features} & {\bf 18} \\
\hline
N1. Centroid location & 3 \\
N2. 2D bounding box & 4 \\
N3. Transformation matrix of SIFT matches between adjacent frames & \multirow{2}{*}{6} \\
N4. Distance moved by the centroid & 1 \\
N5. Displacement of centroid & 1 \\
\hline
{\bf Sub-activity Features} & {\bf 103} \\
\hline
N6. Location of each joint (8 joints) & 24 \\
N7. Distance moved by each joint  (8 joints) & 8 \\
N8. Displacement of each joint  (8 joints) & 8 \\
N9. Body pose features & 47 \\
N10. Hand position features & 16 \\
\hline
{\bf Object-object Features} (computed at start frame, middle frame, end frame, max and min)  & {\bf 20} \\
\hline
E1.  Difference in centroid locations $(\Delta x,\Delta y,\Delta z)$  & 3 \\
E2. Distance between centroids & 1 \\
\hline
{\bf Object--sub-activity Features} (computed at start frame, middle frame, end frame, max and min)  &  \multirow{2}{*}{ \bf 40 }\\
\hline
E3. Distance between each joint location and  object centroid & \multirow{2}{*}{8} \\
\hline
{\bf Object Temporal Features} & {\bf 4} \\
\hline
E4. Total and normalized vertical displacement & 2 \\
E5. Total and normalized distance between centroids & 2 \\
\hline
{\bf Sub-activity Temporal Features} & {\bf 16} \\
\hline
E6.  Total and normalized distance between each corresponding joint locations (8 joints) &  \multirow{2}{*}{16} \\
\hline
\end{tabular}
\label{tab:Features}
\end{table}

For a given object node $i$, the node feature map $\fo{i}$ is a vector of features representing the object's location in the scene and how it changes within the temporal segment. These features include the $(x,y,z)$ coordinates of the object's centroid and the coordinates of the object's bounding box at the middle frame of the temporal segment. We also run a SIFT feature based object tracker \citep{PeleECCV2008} to find the corresponding 
points between the adjacent frames and then compute the transformation matrix based on the matched image points. We add the transformation matrix corresponding to the object in the middle frame with respect to its previous frame to the features in order to capture the object's motion information.
 In addition to the above features, we also compute the total displacement and the total distance moved by the object's centroid in the set of frames belonging to the temporal segment. We then perform cumulative binning of the feature values into 10 bins.
In our experiments, we have $\fo{i} \in \Re^{180}$.

Similarly, for a given sub-activity node $i$, the node feature map $\fs{i}$ gives a vector of features computed using the human skeleton information obtained from running Openni's skeleton 
tracker\footnote{http://openni.org} 
on the RGB-D video.  We compute the features described above 
for each of the upper-skeleton joint (neck, torso, left shoulder, left elbow, left palm, right shoulder, right elbow and right palm) locations relative to the subject's head location. In addition to these, we also consider the body pose and hand position features as described by \cite{SungICRA2012}, thus giving us $\fs{i} \in \Re^{1030}$.

 The edge feature maps $\fe{t}{i}{j}$ describe the relationship between node $i$ and $j$. 
 They are used for modeling  four types of interactions: object--object within a temporal
 segment, object--sub-activity within a temporal segment, object--object between
 two temporal segments, and sub-activity--sub-activity between two temporal segments.
For capturing the \emph{object-object} relations within a temporal segment,
we compute relative geometric features such as the difference in $(x,y,z)$ coordinates of the object centroids and the distance between them. These features are computed at the first, middle and last frames of the temporal segment along with minimum and maximim of their values across all frames in the temporal segment to capture the relative motion information. 
This gives us $\fe{1}{i}{j} \in \Re^{200}$. Similarly for \emph{object--sub-activity} relation features $\fe{2}{i}{j} \in \Re^{400}$, we use the same features as for the \emph{object--object} relation features, but we compute them between the upper-skeleton joint locations and each object's centroid. The temporal relational features capture the change across temporal segments and we use the vertical change in position and the distance between the corresponding object and the joint locations. 
This
gives us $\fe{3}{i}{j} \in \Re^{40}$ and $\fe{4}{i}{j} \in \Re^{160}$ respectively.

\section{Inference and Learning Algorithm}
\label{sec:learninginference}
\subsection{Inference.}
  Given the model parameters $w$, the inference problem is to find the best labeling $\hat{\y}$ for a new video $\x$, i.e. solving the argmax in Eq.~(\ref{eq:argmax}) for the discriminant function in Eq.~(\ref{eq:model}). This is a NP hard problem. However, its equivalent formulation as the following mixed-integer program has a linear relaxation
which can be solved efficiently as a quadratic pseudo-Boolean optimization problem using a graph-cut method
\citep{Kolmogorov/Rother/07}.

{\footnotesize
\begin{align}
& \hat{\y}\!=\!\argmax_{\y}\max_{\mathbf z} \sum_{i \in \mathcal{V}_a}  \sum_{k \in K_a} \ysc{i}{k} \left[\ws{k} \cdot \fs{i} \right] + \!\!\! \sum_{i \in \mathcal{V}_o} \sum_{k \in K_o}  \ysc{i}{k} \left[\wo{k} \cdot \fo{i} \right] \nonumber \\
& \qquad \qquad \qquad  \quad  +  \sum_{t \in {\cal T}} \sum_{(i,j)\in \mathcal{E}_t}   \sum_{(l,k)\in T_t} \zsc{ij}{lk} \left[\we{t}{l}{k} \cdot \fe{t}{i}{j}\right]  
 \label{eq:relaxobj}\\
& \forall i,j,l,k\!:  \:\: \zsc{ij}{lk}\le \ysc{i}{l}, \:\:
\zsc{ij}{lk}\le \ysc{j}{k}, \:\:
\ysc{i}{l} + \ysc{j}{k} \le \zsc{ij}{lk}+1, \:\:
\zsc{ij}{lk},\ysc{i}{l} \in \{ 0,1 \} 
\label{eq:relaxconst}
\end{align}
}

Note that the products $\ysc{i}{l} \ysc{j}{k}$ have been replaced by auxiliary variables $z^{lk}_{ij}$. 
Relaxing the variables $\zsc{ij}{lk}$ and $\ysc{i}{l}$ to the interval $[0,1]$ results in a linear program that can be 
 shown to always have half-integral solutions (i.e. $\ysc{i}{l}$ only take values $\{0,0.5,1\}$ at the solution) \citep{hammer1984roof}. 
Since every node in our experiments has exactly one class label, we also consider the linear relaxation from above with the additional constraints $\forall i \in \mathcal{V}_a: \sum_{l \in K_a} \ysc{i}{l} = 1$ and $\forall i \in \mathcal{V}_o: \sum_{l \in K_o} \ysc{i}{l} = 1$. This problem can no longer be solved via graph cuts.
We compute the exact mixed integer solution including these additional constraint using a general-purpose MIP solver\footnote{http://www.tfinley.net/software/pyglpk/readme.html} during inference. 

In our experiments, we obtain a processing rate of \emph{74.9 frames/second} for inference and 
\emph{16.0 frames/second} end-to-end (including feature computation cost) on
 a 2.93 GHz Intel processor with 16 GB of RAM on Linux.
 In detail, the MIP solver takes 6.94 seconds for a typical video with 520 frames and 
 the corresponding graph has 12 sub-activity nodes and 36 object nodes, i.e. 15908 variables. 
 This is the time 
 corresponding to solving the argmax in Eq.~(\ref{eq:relaxobj}-\ref{eq:relaxconst}) and does not involve the feature 
 computation time. The time taken for end-to-end classification including feature generation is 32.5 seconds.

\subsection{Learning.} 
\label{sec:learning}
We take a large-margin approach to learning the parameter vector $\w$ of Eq.~(\ref{eq:model}) from labeled training examples $(\x_1,\y_1),...,(\x_\M,\y_\M)$ 
\citep{Taskar/AMN,Tsochantaridis/04}.
Our method optimizes a regularized upper bound on the training error

{\small
\begin{align}\textstyle
R(h) = \frac{1}{\M} \sum_{m=1}^{\M} \loss{\y_m}{\hat{\y}_m}, \nonumber
 \label{eq:emprisk}
\end{align}
}

\noindent
where $\hat{\y}_m$ is the optimal solution of Eq.~(\ref{eq:argmax}) and $\loss{\y}{\hat{\y}}$ is the loss function defined as

{\small
\begin{align}\textstyle
\loss{\y}{\hat{\y}} =\sum_{i \in \mathcal{V}_o} \sum_{k \in K_o} |\ysc{i}{k} - \hat{y}_i^k | + \sum_{i \in \mathcal{V}_a} \sum_{k \in K_a} |\ysc{i}{k} - \hat{y}_i^k |. \nonumber
\end{align}
} 

To simplify notation, note that Eq.~(\ref{eq:relaxobj}) can be equivalently written as $\w^T \Psi(\x,\y)$ by appropriately stacking the $\ws{k}$ , $\wo{k}$ and $\we{t}{l}{k}$ into $\w$ and the $\ysc{i}{k}\fs{i}$, $\ysc{i}{k}\fo{i}$ and $\zsc{ij}{lk}\fe{t}{i}{j}$ into $\Psi(\x,\y)$, where each $\zsc{ij}{lk}$ is consistent with Eq.~(\ref{eq:relaxconst}) given~$\y$. Training can then be formulated as the following convex quadratic program \citep{joachims2009cutting}:

{\small
\begin{eqnarray} \label{eq:trainqp}
\min_{w,\xi} & &  \frac{1}{2} \w^T\w + C\xi\\
s.t. & &   \forall \bar{\y}_1,...,\bar{\y}_\M \in \{0,0.5,1\}^{N \cdot K} : \cr
& & \frac{1}{M} \w^T \sum_{m=1}^{M} [\Psi( \x_m, \y_m) - \Psi(\x_m,\bar{\y}_m)] \ge \Delta(\y_m,\bar{\y}_m) -\xi \nonumber
\end{eqnarray}
}

While the number of constraints in this QP is exponential in $\M$, $N$ and $K$, it can nevertheless be solved efficiently using the cutting-plane algorithm 
\citep{joachims2009cutting}. 
The algorithm needs access to an efficient method for computing

{\small
\begin{align}
\bar{\y}_m & = & \!\!\!\!\!\argmax_{\y \in \{0,0.5,1\}^{N \cdot K}} \left[ \w^T \Psi(\x_m,\y) + \loss{\y_m}{\y} \right].
\end{align} }

Due to the structure of $\loss{.}{.}$, this problem is identical to the relaxed prediction problem in Eqs.~(\ref{eq:relaxobj})-(\ref{eq:relaxconst}) and can be solved efficiently using graph cuts.



\begin{figure*}[t]
\centering
\includegraphics[width=.16\linewidth,height=0.6in]{images/reaching/frame1065.jpg}
\includegraphics[width=.16\linewidth,height=0.6in]{images/reaching/frame1075.jpg}
\includegraphics[width=.16\linewidth,height=0.6in]{images/reaching/frame1115.jpg}
\includegraphics[width=.16\linewidth,height=0.6in]{images/reaching/frame1200.jpg}
\includegraphics[width=.16\linewidth,height=0.6in]{images/reaching/frame1140.jpg}
\includegraphics[width=.16\linewidth,height=0.6in]{images/reaching/frame1130.jpg}
\vskip 0.096in
\includegraphics[width=.16\linewidth,height=0.6in]{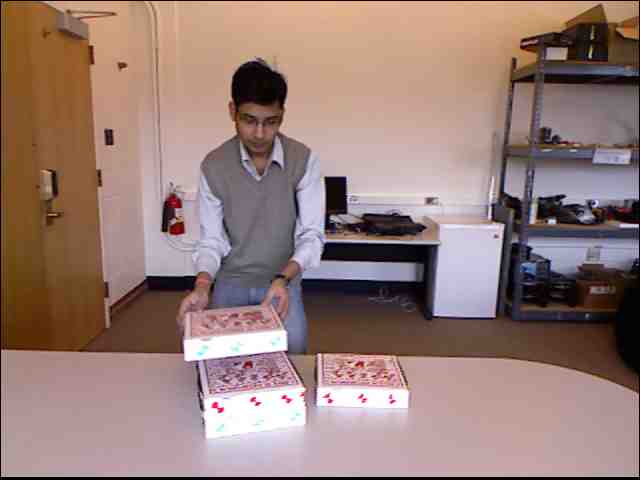}
\includegraphics[width=.16\linewidth,height=0.6in]{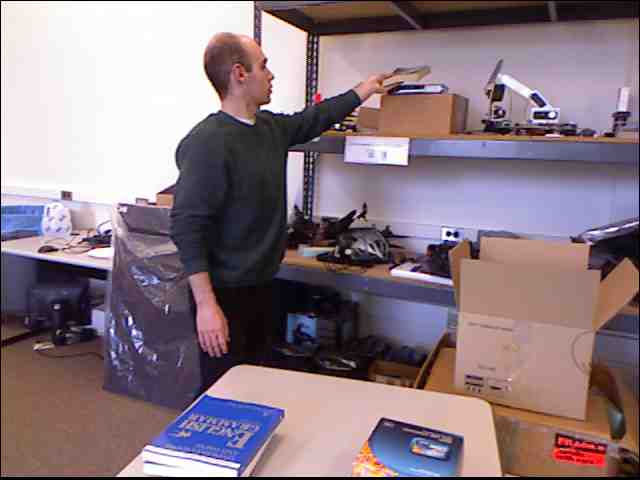}
\includegraphics[width=.16\linewidth,height=0.6in]{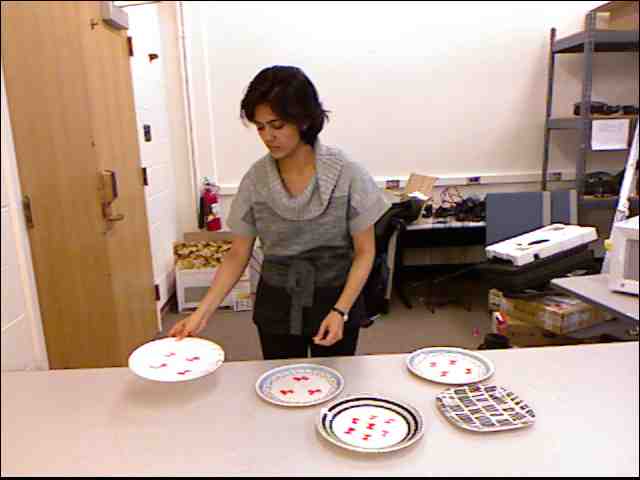}
\includegraphics[width=.16\linewidth,height=0.6in]{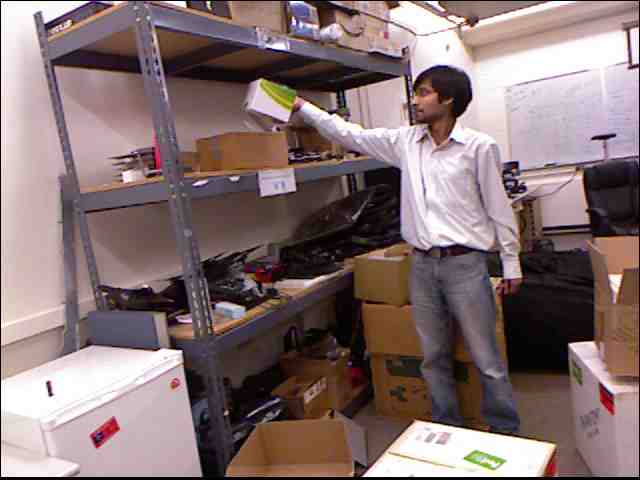}
\includegraphics[width=.16\linewidth,height=0.6in]{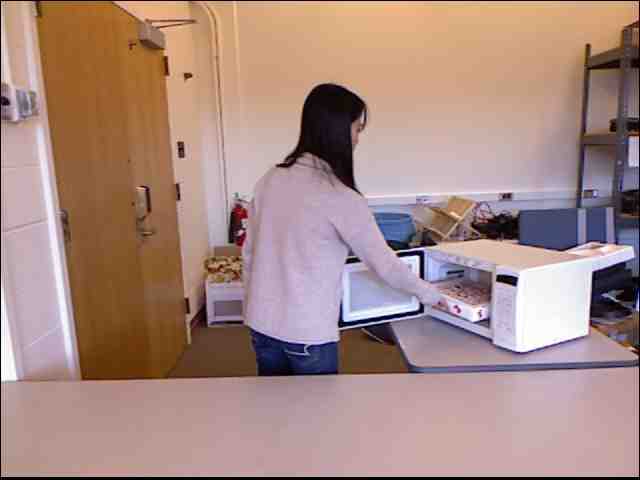}
\includegraphics[width=.16\linewidth,height=0.6in]{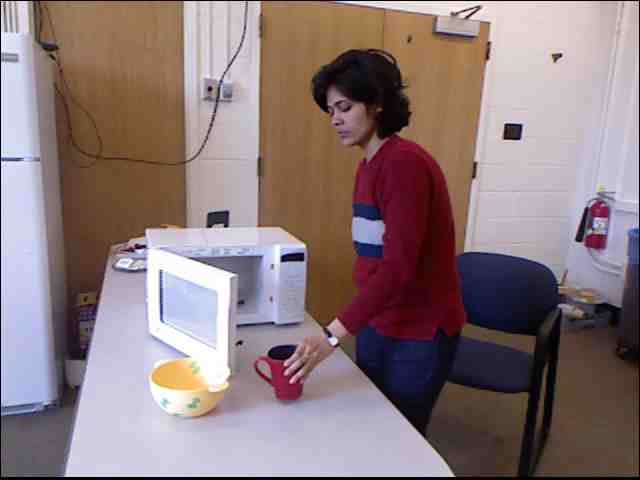}
\vskip 0.096in
\includegraphics[width=.16\linewidth,height=0.6in]{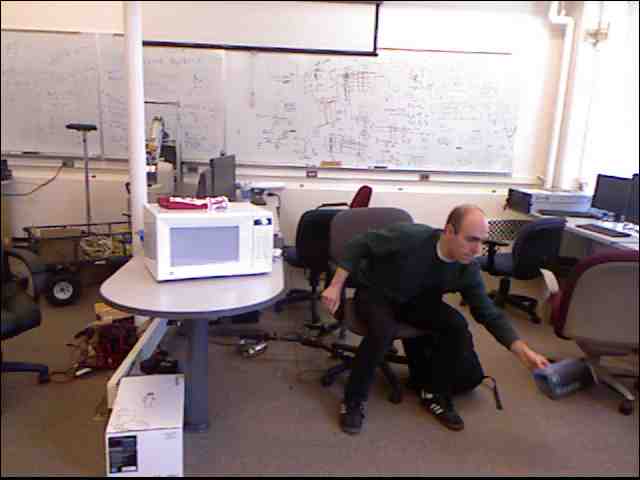}
\includegraphics[width=.16\linewidth,height=0.6in]{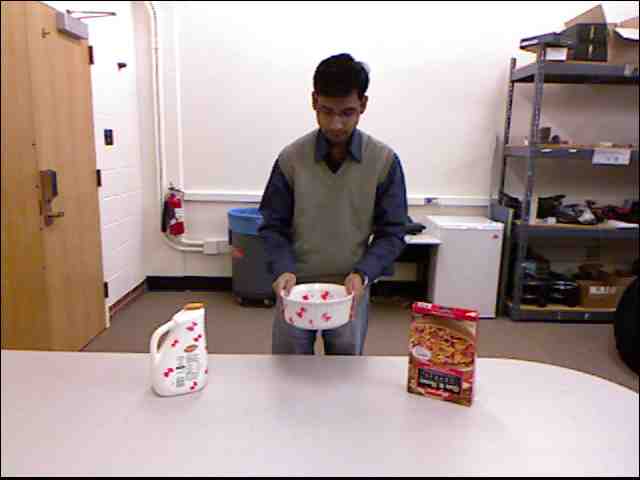}
\includegraphics[width=.16\linewidth,height=0.6in]{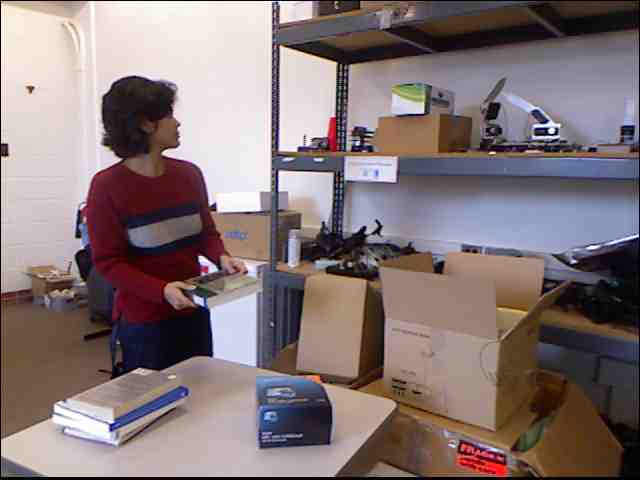}
\includegraphics[width=.16\linewidth,height=0.6in]{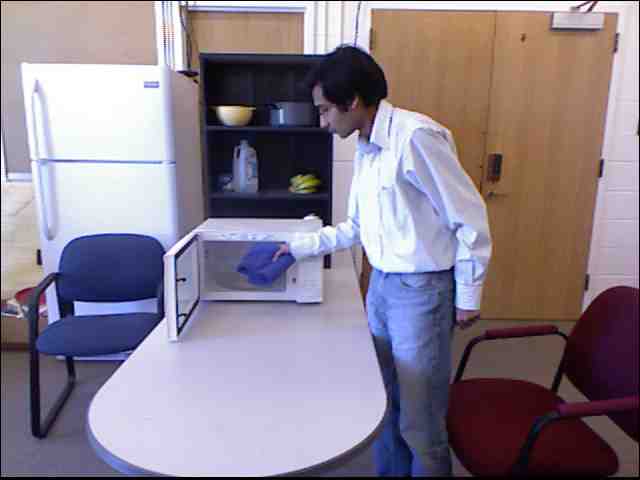}
\includegraphics[width=.16\linewidth,height=0.6in]{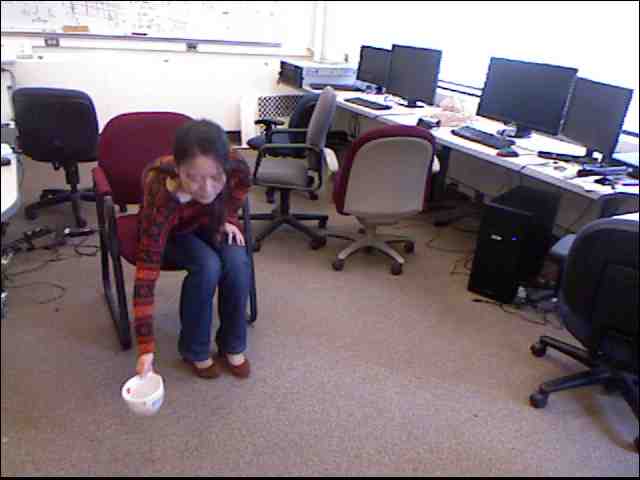}
\includegraphics[width=.16\linewidth,height=0.6in]{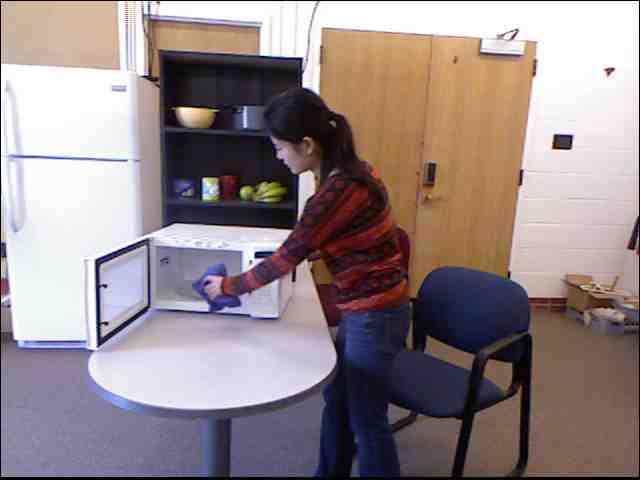}
\vskip 0.096in
\includegraphics[width=.16\linewidth,height=0.6in]{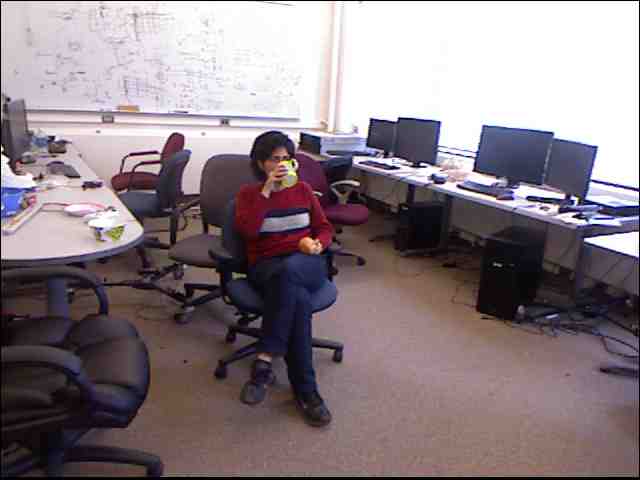}
\includegraphics[width=.16\linewidth,height=0.6in]{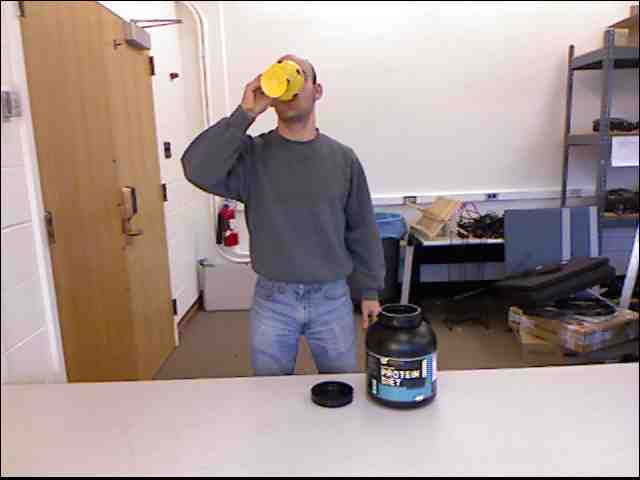}
\includegraphics[width=.16\linewidth,height=0.6in]{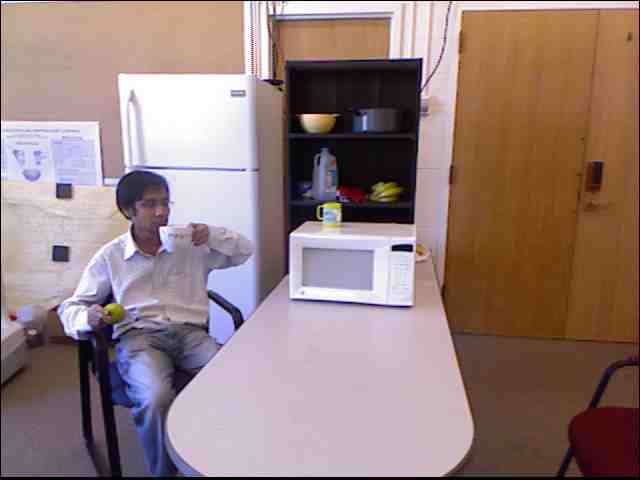}
\includegraphics[width=.16\linewidth,height=0.6in]{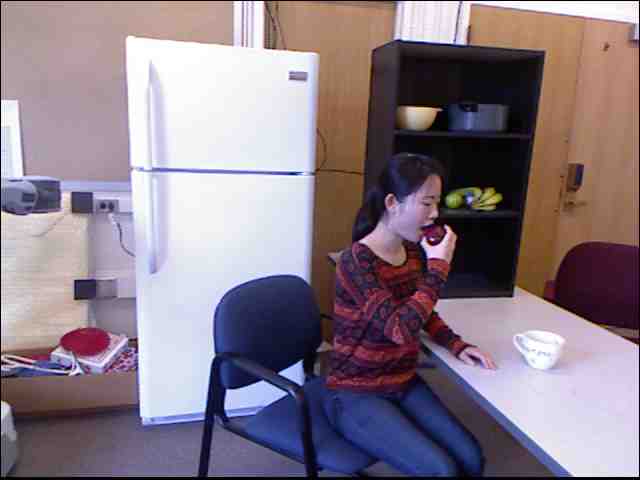}
\includegraphics[width=.16\linewidth,height=0.6in]{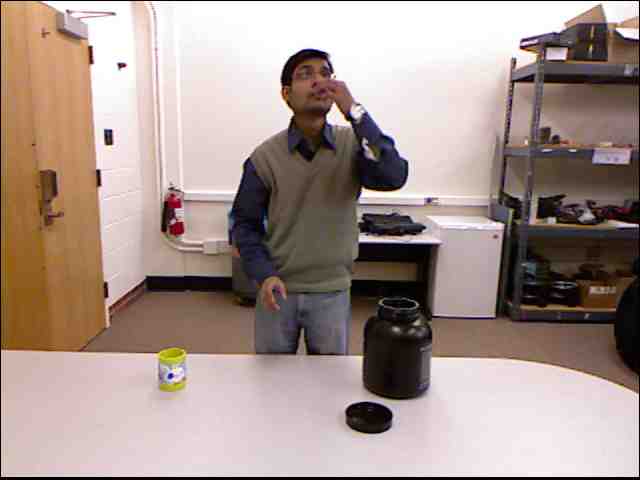}
\includegraphics[width=.16\linewidth,height=0.6in]{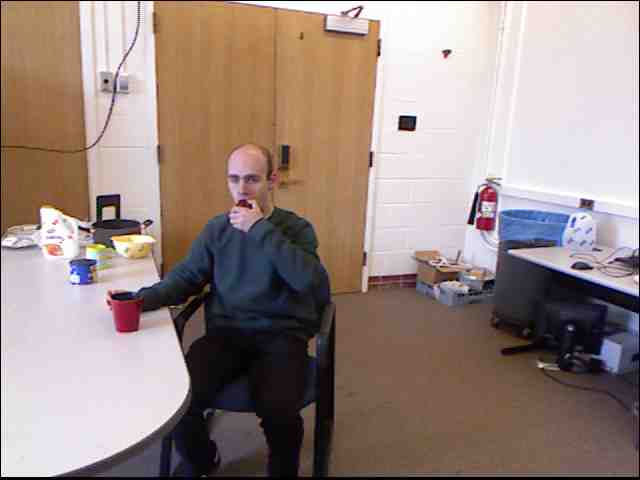}

\caption{Example shots of \emph{reaching} (first row), \emph{placing} (second row), \emph{moving} (third row),  \emph{drinking} (fourth row) and \emph{eating}  (fourth row)
 sub-activities from our dataset.
There are significant variations in the way the subjects perform the sub-activity.
}

 \label{fig:dataEx}
\end{figure*}

\begin{table*}[t]
\caption{{\small {\bf Description of high-level activities in terms of sub-activities.} Note that some activities consist of same sub-activities but are executed in different order. The high-level activities (rows) are learnt using the algorithm in Section \ref{sec:learning2} and the sub-activities (columns) are learnt using the algorithm in Section \ref{sec:learning}.}}
 \label{tbl:activities}
{\small
\newcolumntype{P}[2]{>{\small#1\hspace{0pt}\arraybackslash}p{#2}}
\setlength{\tabcolsep}{2pt}
\centering
\resizebox{\hsize}{!}
 {
\begin{tabular}
{p{0.20\linewidth} |P{\centering}{12mm}|P{\centering}{12mm}|P{\centering}{12mm}|P{\centering}{12mm}|P{\centering}{12mm}|P{\centering}{12mm}|P{\centering}{12mm}|P{\centering}{12mm}|P{\centering}{14mm}|P{\centering}{12mm}}
\whline{1.1pt}
 & reaching & moving & placing & opening & closing & eating & drinking & pouring & scrubbing & null \\ 
\whline{1.1pt}
Making Cereal & \checkmark  & \checkmark & \checkmark  &  &  &  &   &\checkmark  &  & \checkmark  \\ 
 \whline{0.8pt} 
Taking  Medicine & \checkmark  & \checkmark & \checkmark  & \checkmark &  & \checkmark &  \checkmark  &  &  & \checkmark  \\
 \whline{0.8pt} 
 Stacking Objects & \checkmark  & \checkmark & \checkmark  &  &  &  &  &  &  & \checkmark  \\
 \whline{0.8pt} 
 Unstacking Objects & \checkmark  & \checkmark & \checkmark  &  &  &  &  &   &  & \checkmark  \\
 \whline{0.8pt} 
Microwaving Food & \checkmark  & \checkmark & \checkmark  & \checkmark & \checkmark &  &   & &  &\checkmark   \\
 \whline{0.8pt} 
Picking Objects & \checkmark  & \checkmark &   &  &  &  &  &   &  &  \checkmark \\
 \whline{0.8pt} 
Cleaning Objects & \checkmark  & \checkmark &  & \checkmark & \checkmark &  &  &   &  \checkmark&  \checkmark \\
 \whline{0.8pt} 
 Taking Food & \checkmark  & & \checkmark  & \checkmark & \checkmark &  &  &  &  & \checkmark  \\
 \whline{0.8pt} 
 Arranging Objects & \checkmark  & \checkmark & \checkmark  &  &  &  &  &   &  & \checkmark  \\
 \whline{0.8pt} 
  Having a Meal & \checkmark  & \checkmark &  &  &  & \checkmark & \checkmark &  &  & \checkmark  \\  
 \whline{1.1pt}
\end{tabular}
}
}

\end{table*}

\subsection{Multiple Segmentations}
\label{sec:segmentation}

Segmenting an RGB-D video in time can be noisy, and multiple segmentations may be valid. 
Therefore, we perform multiple segmentations by using different methods and criterion
of segmentation (see Section~\ref{sec:temporalseg} for details). Thus, we get a set 
$\mathcal{H}$ of multiple segmentations, and
 let $h_n$ be the $n^{th}$ segmentation.
A discriminant function $\df{\xh{n}}{\yh{n}}{\wh{n}}$ can now be defined for each 
$h_n$ as in Eq.~(\ref{eq:model}).  We now define a score function 
$\dg{\yh{n}}{\y}{\theta}$ which gives a score for assigning the labels of the segments from $\yh{n}$ to $\y$, 

{\small
\begin{eqnarray}
\dg{\yh{n}}{\y}{\theta_n} = \sum_{k \in K} \sum_{i \in \mathcal{V}} \theta_{n}^k \mathbf y^{h_nk}_i  \y_i^k 
\end{eqnarray} }

\noindent
where $K=K_o \cup K_a$. Here, $ \mathbf \theta_{n}^k$ can be interpreted as the confidence of labeling the segments of label 
$k$ correctly in the $n^{th}$ segmentation hypothesis. We want to find the labeling that maximizes the assignment score across all the segmentations. Therefore we can write inference in terms of a joint objective function as follows

{\small
\begin{eqnarray}
\label{eq:combinedObj}
\hat{\y} = \argmax_\y \max_{\yh{n} \forall h_n \in \mathcal{H}} \sum_{h_n \in \mathcal{H}} [ \df{\xh{n}}{\yh{n}}{\wh{n}} +  \dg{\yh{n}}{\y}{\theta_n} ]
\end{eqnarray}
}

This formulation is equivalent to considering the labelings $y^{h_n}$ over the segmentations as unobserved variables. It is possible to use 
the latent structural SVM \citep{Yu:2009} to solve this, but it becomes intractable if the size of the segmentation hypothesis space is large. Therefore we propose an approximate two-step learning procedure to address this.
For a given set of segmentations $\mathcal{H}$, we first learn the parameters $\wh{n}$ independently 
as described in Section \ref{sec:model}.  We then train the parameters $\mathbf{\theta}$  on a separate held-out training dataset. This can now be formulated as a QP

{\small
\begin{align}
\label{eq:model2}
\min_{\theta} & &  \frac{1}{2} \theta^T\theta - \sum_{h_n \in \mathbf{H}} \dg{\yh{n}}{\y}{\theta_n} \cr
s.t. & & \forall k \in K : \sum_{n =1}^{|\mathbf{H}|} \theta_n^k = 1  
\end{align}  
}

Using the fact that  the objective function defined in Eq.~(\ref{eq:combinedObj}) is convex,
we design an iterative two-step procedure where we solve for $\yh{n}, \forall h_n \in \mathbf{H}$  in parallel 
and then solve for $\y$. This method is guaranteed to converge,
and when the number of variables scales linearly with the number of segmentation hypothesis considered, the original
problem in Eq.~(\ref{eq:combinedObj})  will become considerably slow, but our method will still scale. More formally,
we iterate between the following two problems:

{\small

\begin{eqnarray}
\yhhat{n} &=& \argmax_\yh{n} \df{\xh{n}}{\yh{n}}{\wh{n}} + \dg{\yh{n}}{\hat{\y}}{\theta_n} \qquad\label{eq:inference1} \\
\hat{\y} &=& \argmax_\y \dg{\yhhat{n}}{\y}{\theta_n} \label{eq:inference2}
\end{eqnarray}

}

\subsection{High-level Activity Classification.}  
\label{sec:learning2}
For classifying the high-level activity, 
we compute the histograms of sub-activity and affordance labels and use 
them as features. 
 However, some high-level activities, such as \emph{stacking objects} and \emph{unstacking objects}, 
 have the same sub-activity and affordance sequences. 
 Occlusion of objects plays a major role in being able to differentiate such activities.
 Therefore, we compute additional occlusion features
 by dividing the video into $n$ uniform length segments and 
finding the fraction of objects that are occluded fully or partially in the temporal segments.
We then train a multi-class SVM classifier on training data using these features.

\section{Experiments}\label{sec:experiments}

\subsection{Data} 
We test our model on two 3D activity datasets: Cornell Activity Dataset - 60 \citep[CAD-60,][]{SungICRA2012}
and one that we collected.
The CAD-60 dataset has 60 {RGB-D} videos of four different subjects performing 12 high-level activity classes. However, some of these activity classes contain only one sub-activity (e.g. \emph{working on a computer}, \emph{cooking (stirring)}, etc.) and do not contain object interactions (e.g. \emph{talking on couch}, \emph{relaxing on couch}). 

We collected the CAD-120 dataset (available at: http://pr.cs.cornell.edu/humanactivities, along with 
the code), which contains 120 
activity sequences of ten different high-level activities performed 
by four different subjects, where each high-level activity was performed three times. 
We thus have a total of {\bf 61,585} RGB-D video frames in our dataset. 
The high-level activities are \{\emph{making cereal}, \emph{taking medicine}, \emph{stacking objects}, \emph{unstacking objects}, \emph{microwaving food}, \emph{picking objects}, \emph{cleaning objects}, \emph{taking food}, \emph{arranging objects}, \emph{having a meal}\}. The subjects were only given a high-level 
description of the task,\footnote{For example, the instructions for \emph{making cereal} were: 
1) Place bowl on table, 2) Pour cereal, 3) Pour milk. For \emph{microwaving food}, they were: 1) Open microwave door, 2) Place food inside, 3) Close microwave door.}  
and were asked to perform the activities multiple times with \emph{different} objects.
For example, the stacking and unstacking activities were performed with 
pizza boxes, plates and bowls.  
They performed the activities through a long sequence of sub-activities, which varied from subject to subject significantly in terms of length of the sub-activities, order of the sub-activities as well as in the way they executed
the task. Table \ref{tbl:activities} specifies the set of sub-activities involved in each high-level activity. 
The camera was mounted so that 
the subject was in view (although the subject may not be facing the camera), but often there were significant occlusions of the body parts.
See Fig.~\ref{fig:reachingData} and Fig.~\ref{fig:dataEx} for some 
examples.

We labeled our CAD-120 dataset with the sub-activity and the object affordance labels. 
Specifically, our sub-activity labels are \{\emph{reaching}, \emph{moving}, \emph{pouring}, \emph{eating}, \emph{drinking}, \emph{opening}, \emph{placing}, \emph{closing}, \emph{scrubbing}, \emph{null}\}  and our affordance labels are \{\emph{reachable}, \emph{movable}, \emph{pourable}, \emph{pourto}, \emph{containable}, \emph{drinkable}, \emph{openable}, \emph{placeable}, \emph{closable}, \emph{scrubbable}, \emph{scrubber}, \emph{stationary}\}.

\subsection{Object Tracking Results}
\begin{figure*}[t]
\centering
\includegraphics[width=.16\linewidth,height=0.8in]{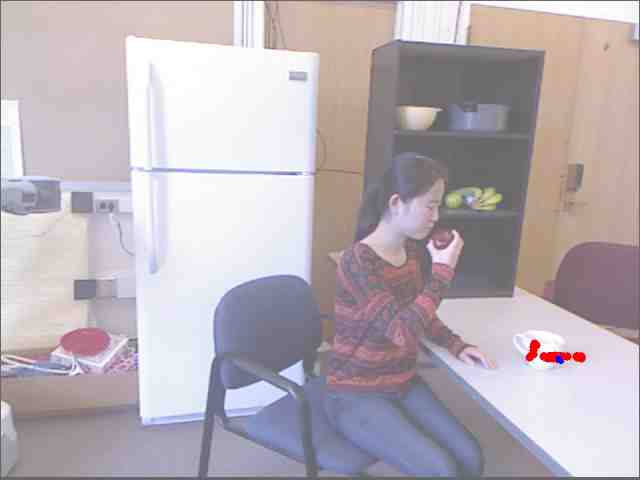}
\includegraphics[width=.16\linewidth,height=0.8in]{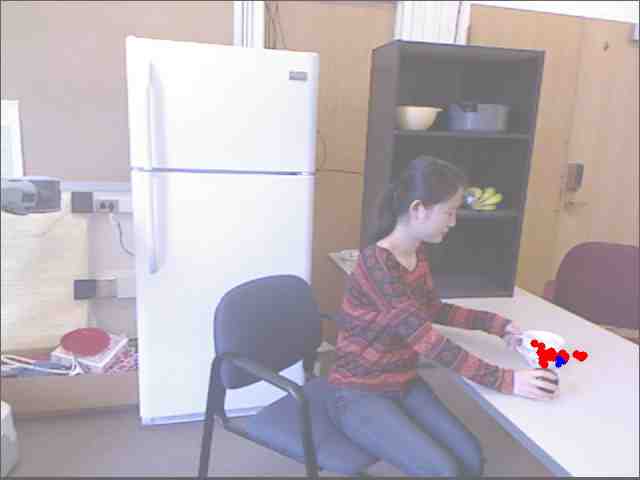}
\includegraphics[width=.16\linewidth,height=0.8in]{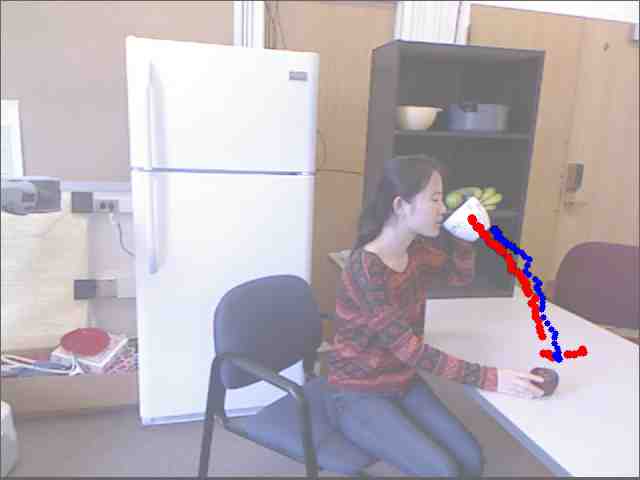}
\includegraphics[width=.16\linewidth,height=0.8in]{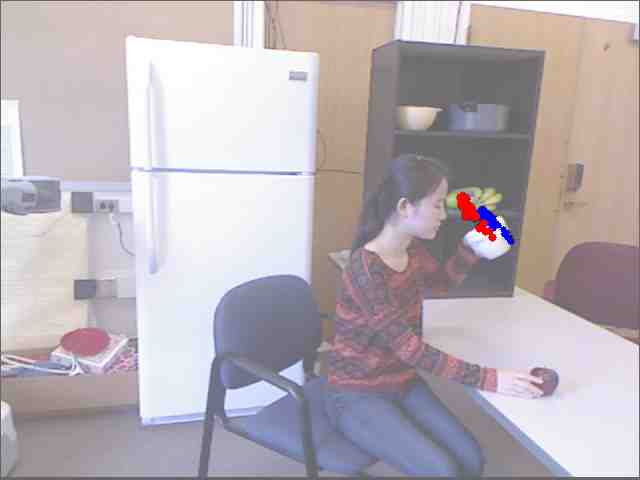}
\includegraphics[width=.16\linewidth,height=0.8in]{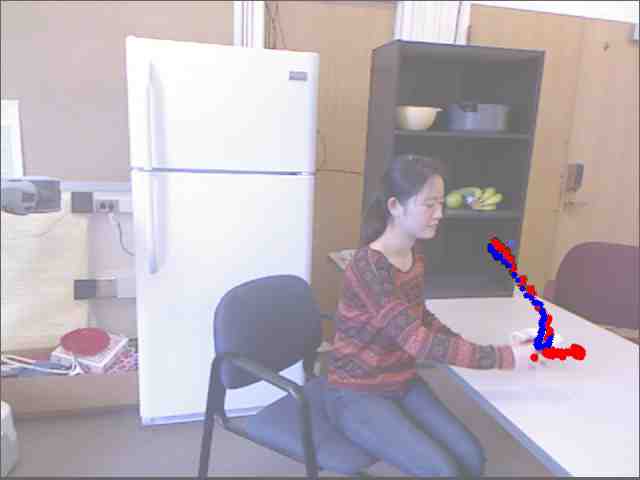}
\includegraphics[width=.16\linewidth,height=0.8in]{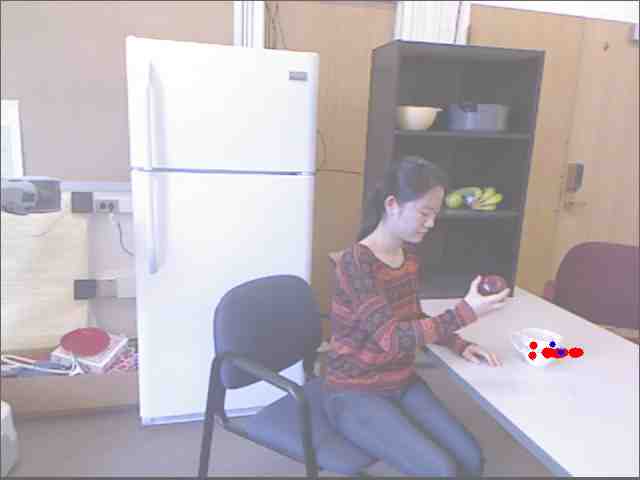}
\\ 
\vskip 0.048in
\includegraphics[width=.16\linewidth,height=0.8in]{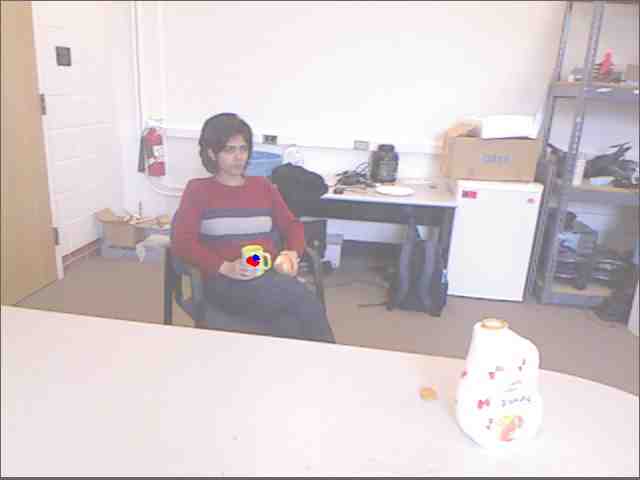}
\includegraphics[width=.16\linewidth,height=0.8in]{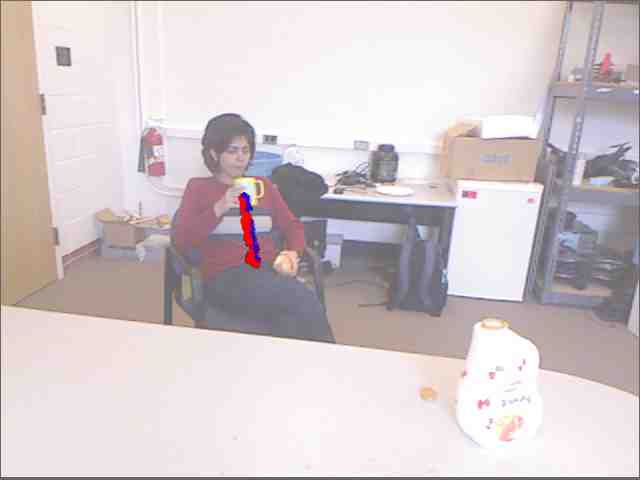}
\includegraphics[width=.16\linewidth,height=0.8in]{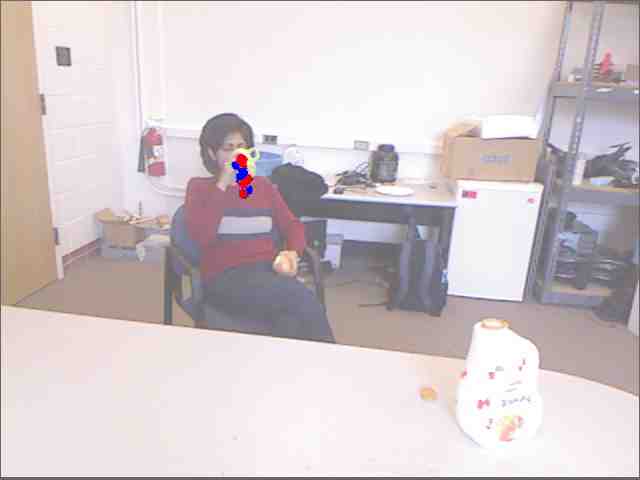}
\includegraphics[width=.16\linewidth,height=0.8in]{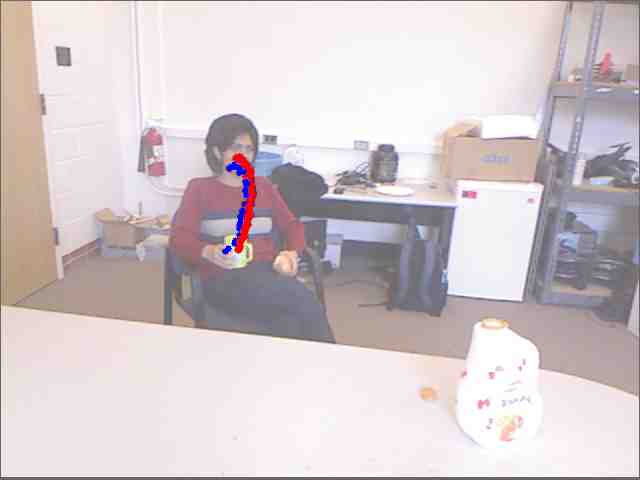}
\includegraphics[width=.16\linewidth,height=0.8in]{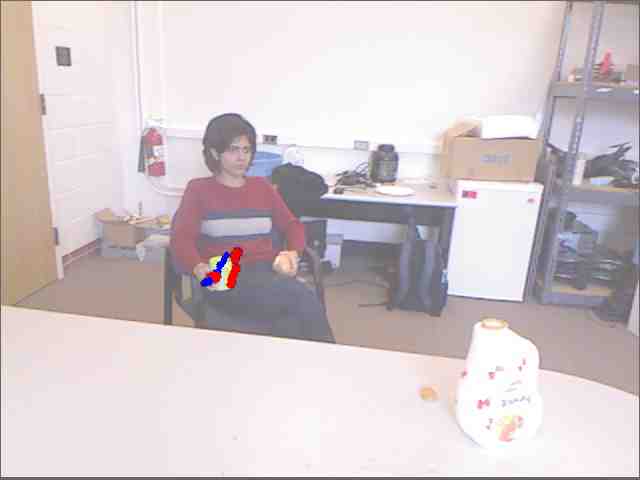}
\includegraphics[width=.16\linewidth,height=0.8in]{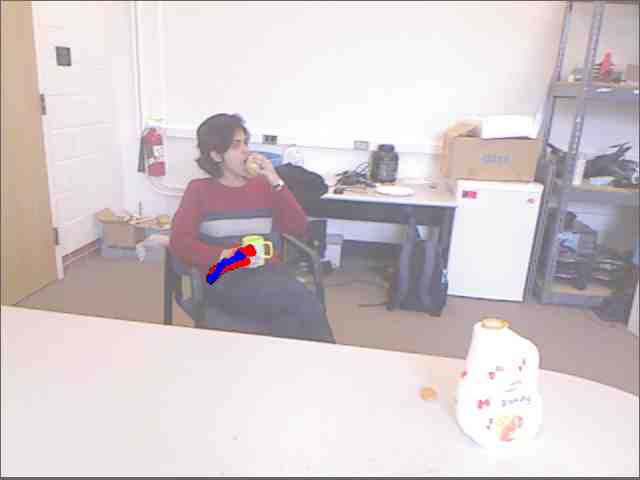}
\\
\vskip 0.048in
\includegraphics[width=.16\linewidth,height=0.8in]{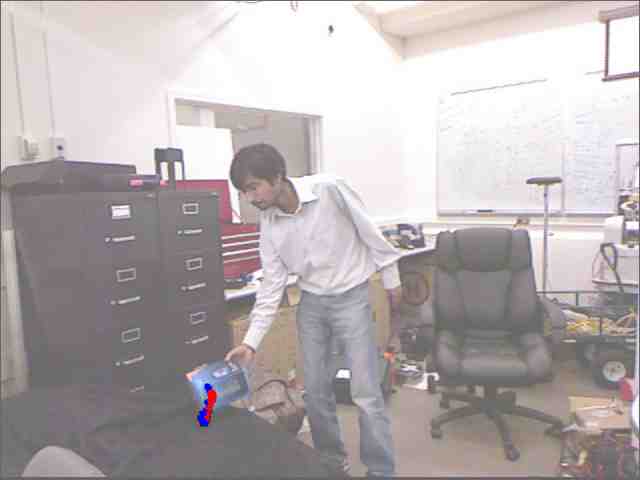}
\includegraphics[width=.16\linewidth,height=0.8in]{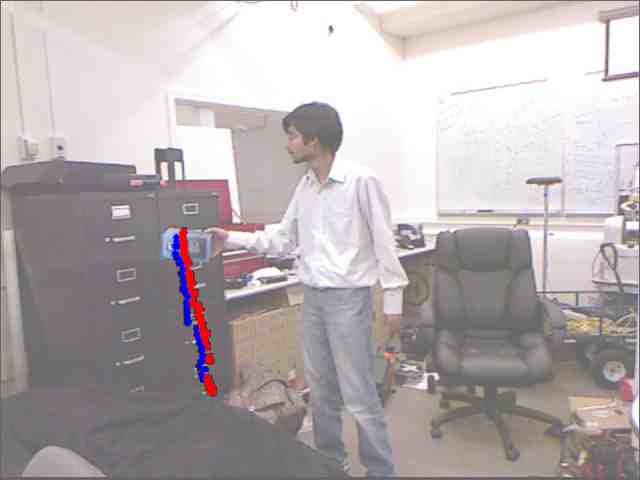}
\includegraphics[width=.16\linewidth,height=0.8in]{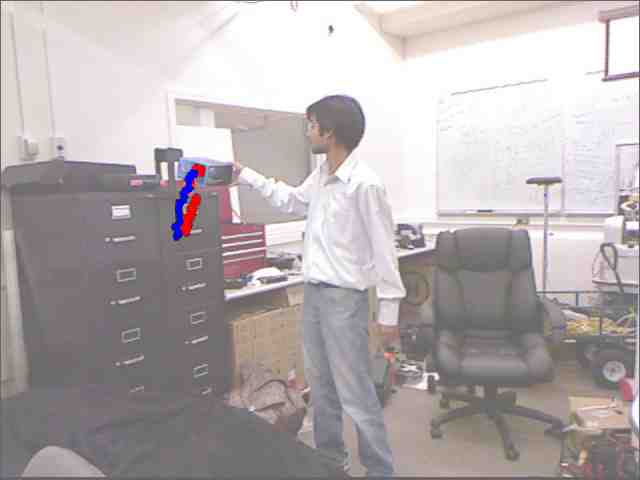}
\includegraphics[width=.16\linewidth,height=0.8in]{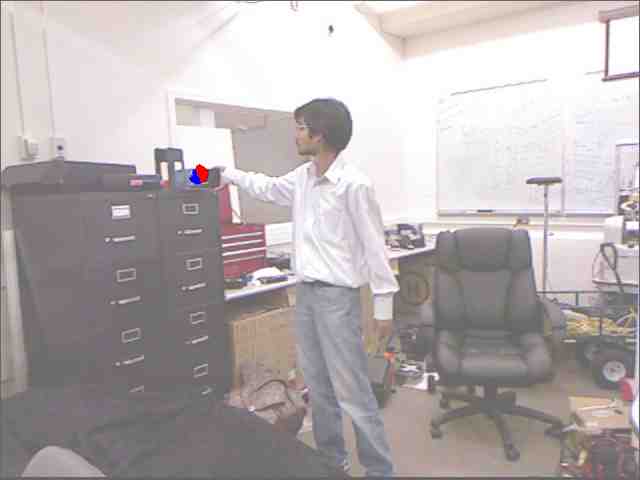}
\includegraphics[width=.16\linewidth,height=0.8in]{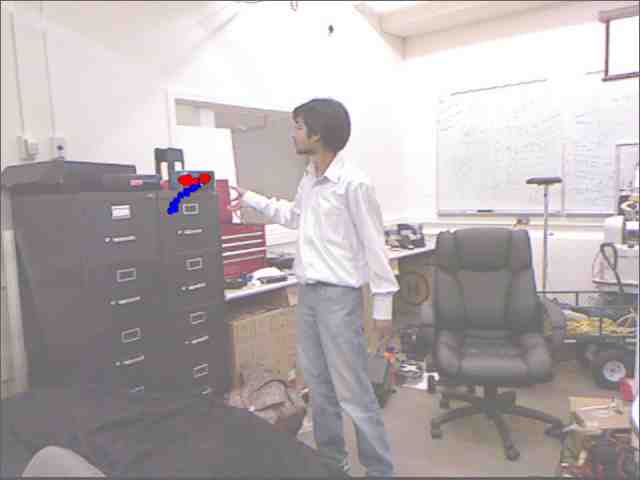}
\includegraphics[width=.16\linewidth,height=0.8in]{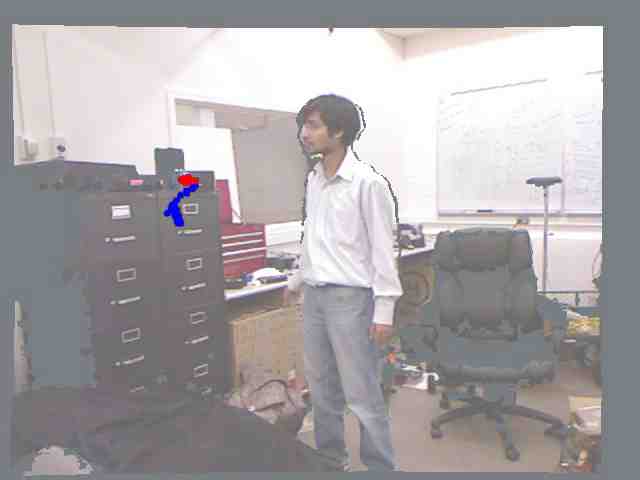}
\\
\vskip 0.048in
\includegraphics[width=.16\linewidth,height=0.8in]{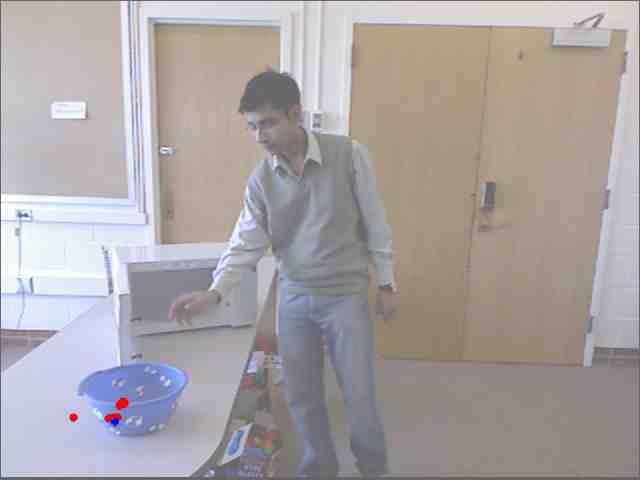}
\includegraphics[width=.16\linewidth,height=0.8in]{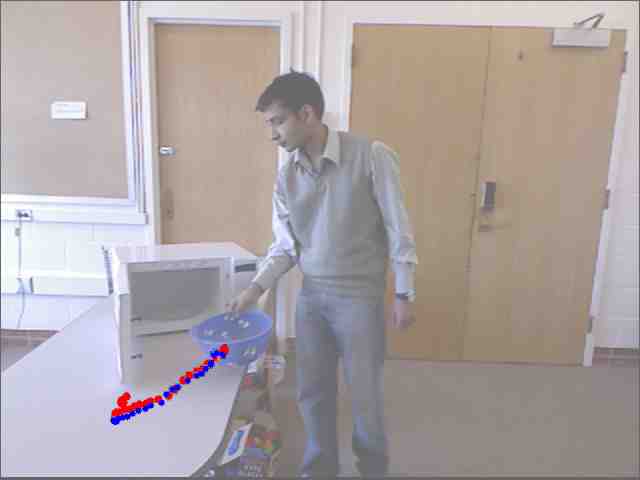}
\includegraphics[width=.16\linewidth,height=0.8in]{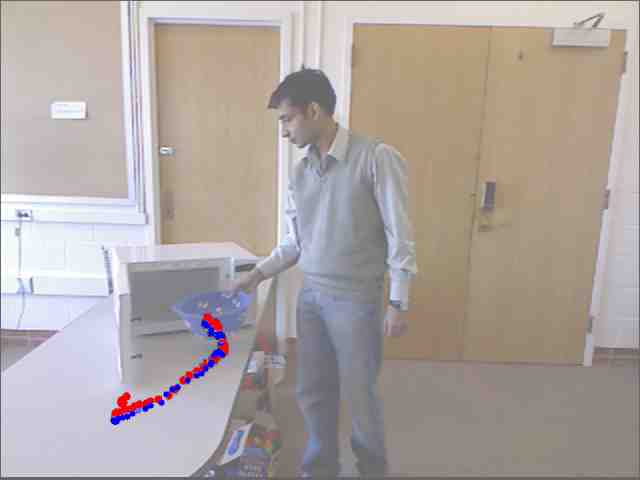}
\includegraphics[width=.16\linewidth,height=0.8in]{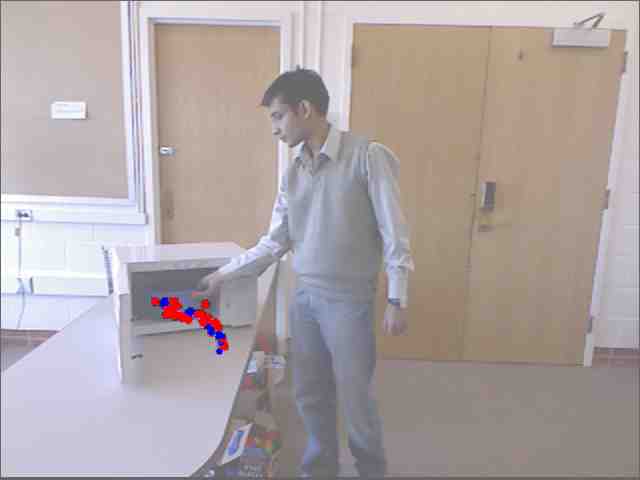}
\includegraphics[width=.16\linewidth,height=0.8in]{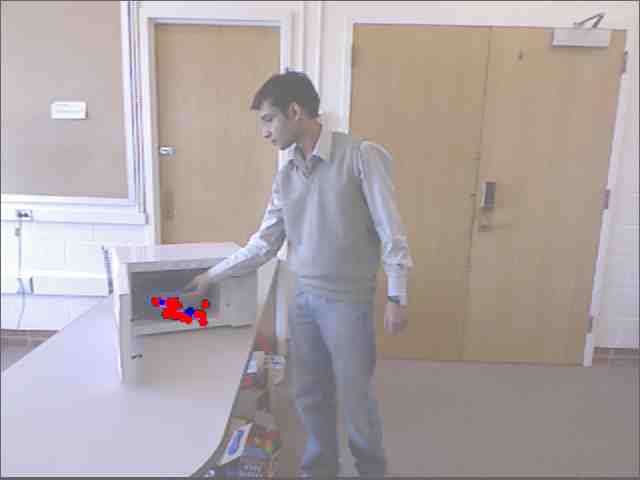}
\includegraphics[width=.16\linewidth,height=0.8in]{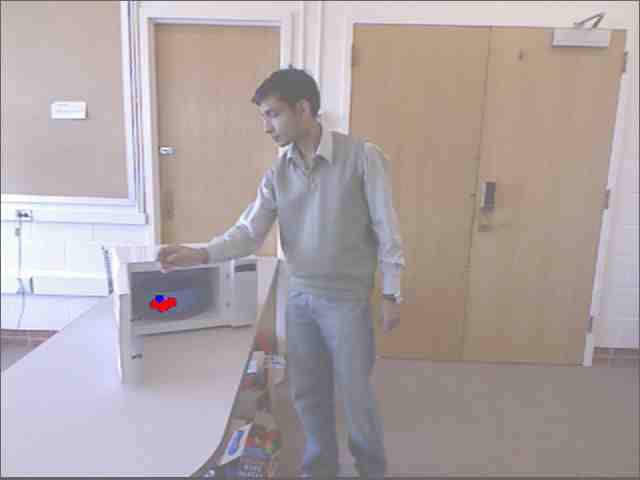}

\caption{Tracking Results: Blue dots represent the trajectory of the center of tracked bounding box and red dots represent the trajectory of the center of ground-truth bounding box. (Best viewed in color.)}

 \label{fig:tracking_result}
\end{figure*}

In order to evaluate our object detection and tracking method, we have generated the ground-truth
bounding boxes of the objects involved in the activities. We do this by manually labeling the object 
bounding boxes in the images corresponding to every $50^{th}$ frame. We compute the bounding
boxes in the rest of the frames by tracking using 
SIFT feature matching \citep{PeleECCV2008}, 
while enforcing depth consistency across the time frames for obtaining reliable object tracks. 

Fig.~\ref{fig:tracking_result} shows the 
visual output of our tracking algorithm. The center of the bounding box for each frame of the output
is marked with a blue dot and that of the ground-truth is marked with a red dot. 
We compute the overlap of the bounding boxes obtained from our tracking method with the generated ground-truth bounding boxes. Table \ref{tbl:tracking_result} shows the percentage 
overlap with the ground-truth when considering tracking from the given bounding box in the 
first frame both with and without object detections. As can be seen from 
Table \ref{tbl:tracking_result}, our tracking algorithm produces greater than 10\% overlap with the
ground truth bounding boxes for 77.8\% of the frames. Since, we only require
that an approximate bounding box of the objects are given, 10\% overlap is sufficient. We study the effect of
the errors in tracking on the performance of our algorithm in Section \ref{sec:cad120results}.
\begin{table}[h]
\caption{{\bf Object Tracking Results}, showing the \% of frames which have $\ge$40\%, $\ge$20\% and $\ge$10\% overlap with the ground-truth object bounding boxes.}
 \label{tbl:tracking_result}
\newcolumntype{P}[2]{>{\footnotesize#1\hspace{0pt}\arraybackslash}p{#2}}
\setlength{\tabcolsep}{2pt}
\centering
\resizebox{\hsize}{!}
{
\begin{tabular}
{@{}p{0.33\linewidth} |P{\centering}{12mm}|P{\centering}{12mm}|P{\centering}{12mm}@{}}
\whline{1.1pt}
 & $\ge$40\% & $\ge$20\%  & $\ge$10\%    \\
\hline
tracking w/o detection  & 49.2  & 65.7 & 75  \\
tracking + detection & 53.5 & 69.4& 77.8  \\
\whline{1.1pt}
\end{tabular}
}
\end{table}

\begin{table*}[tb!]
\caption{{\bf Results on Cornell Activity Dataset \citep{SungICRA2012}}, tested on \emph{``New Person"} data for 12 activity classes.}
 \label{tbl:cad_result}
\centering
{\scriptsize
\begin{tabular}{l|cc|cc|cc|cc|cc|cc}
\whline{1.1pt}
& \multicolumn{2}{c|}{bathroom} & \multicolumn{2}{c|}{bedroom} & \multicolumn{2}{c|}{kitchen} & \multicolumn{2}{c|}{living room} & \multicolumn{2}{c|}{office} & \multicolumn{2}{c}{Average}\\
 &   prec.~(\%) & rec.~(\%) & prec.~(\%) & rec.~(\%)  &   prec.~(\%) & rec.~(\%)  &   prec.~(\%) & rec.~(\%)   &   prec.~(\%) & rec.~(\%)  &   prec.~(\%) & rec.~(\%)   \\
\hline
~\cite{SungICRA2012} & 72.7 & \textbf{65.0} & \textbf{76.1} & 59.2 & 64.4 & 47.9 & 52.6 & 45.7 &73.8  & 59.8 & 67.9 & 55.5 \\
Our method & \textbf{88.9} &61.1 & 73.0 & \textbf{66.7} & \textbf{96.4} & \textbf{85.4} & \textbf{69.2} & \textbf{68.7} &  \textbf{76.7} & \textbf{75.0}  & \textbf{80.8} & \textbf{71.4} \\
\whline{1.1pt}
\end{tabular}
}
\end{table*}

\subsection{Labeling results on the Cornell Activity Dataset 60 (CAD-60)}
 Table \ref{tbl:cad_result} shows the precision and recall of the high-level activities on the CAD-60 dataset \citep{SungICRA2012}. Following 
Sung et al.'s \citeyearpar{SungICRA2012} experiments, we considered the same five groups of activities based on their location, and learnt a separate model 
for each location. To make it a fair comparison, we do not assume perfect segmentation of sub-activities and do not use any object information. Therefore, we train our model with only sub-activity nodes and consider segments of uniform size (20 frames per segments). We consider only a subset of our features described in Section \ref{sec:model} that are possible to compute from the tracked human skeleton and RGB-D data provided in this dataset. 
Table \ref{tbl:cad_result} shows that our model significantly outperforms Sung et al.'s MEMM model even when using only the sub-activity nodes and a simple segmentation algorithm. 


\begin{table*}[tb!]
\caption{{\bf Results on our CAD-120 dataset}, showing average micro precision/recall, and average macro precision and recall for affordance, sub-activities and high-level activities.
Standard error is also reported.}
 \label{tbl:labeling_result}
{\footnotesize
\newcolumntype{P}[2]{>{\footnotesize#1\hspace{0pt}\arraybackslash}p{#2}}
\setlength{\tabcolsep}{2pt}
\centering
\resizebox{\hsize}{!}
 {
\begin{tabular}
{@{}p{0.22\linewidth} |P{\centering}{15mm}P{\centering}{15mm}P{\centering}{15mm}|P{\centering}{15mm}P{\centering}{15mm}P{\centering}{15mm}| P{\centering}{15mm}P{\centering}{15mm}P{\centering}{15mm}@{}}
\multicolumn{10}{c}{Full model, assuming ground-truth temporal segmentation is given.}\\
\hline
\whline{1.1pt}
 & \multicolumn{3}{c|}{Object Affordance} & \multicolumn{3}{c|}{Sub-activity} & \multicolumn{3}{c}{High-level Activity} \\
\cline{2-10}
  &\multicolumn{1}{c}{micro} & \multicolumn{2}{c|}{macro} & \multicolumn{1}{c}{micro} &  \multicolumn{2}{c|}{macro}  & \multicolumn{1}{c}{micro} &  \multicolumn{2}{c}{macro}  \\
\whline{0.4pt}
 method & $P/R$ (\%) & Prec.~(\%)  & \multicolumn{1}{c|}{Recall (\%)} &  $P/R$ (\%) & Prec.~(\%) &  \multicolumn{1}{c|}{Recall (\%)} & P/R (\%) & Prec.~(\%) &  \multicolumn{1}{c}{Recall (\%)}  \\
\whline{0.8pt}
 \emph{max class}   & 65.7 $\pm$ 1.0 & 65.7 $\pm$ 1.0 & 8.3 $\pm$ 0.0 & 29.2 $\pm$ 0.2  & 29.2 $\pm$ 0.2  & 10.0 $\pm$ 0.0 & 10.0 $\pm$ 0.0 & 10.0 $\pm$ 0.0   & 10.0 $\pm$ 0.0   \\
\emph{image only}                  & 74.2 $\pm$ 0.7 & 15.9 $\pm$ 2.7 & 16.0 $\pm$ 2.5 & 56.2 $\pm$ 0.4 & 39.6 $\pm$ 0.5 & 41.0 $\pm$ 0.6 & 34.7 $\pm$ 2.9 & 24.2 $\pm$ 1.5 & 35.8 $\pm$ 2.2  \\
\emph{SVM multiclass}                    & 75.6 $\pm$ 1.8 & 40.6 $\pm$ 2.4 & 37.9 $\pm$ 2.0 & 58.0 $\pm$ 1.2 & 47.0 $\pm$ 0.6 & 41.6 $\pm$ 2.6 & 30.6 $\pm$ 3.5 & 27.4 $\pm$ 3.6 & 31.2 $\pm$ 3.7 \\
\emph{MEMM} \citep{SungICRA2012}                      &  - & - & - &  -  & - & - & 26.4 $\pm$ 2.0 & 23.7 $\pm$ 1.0 & 23.7 $\pm$ 1.0 \\
\whline{0.6pt}
\emph{object only}                           & 86.9 $\pm$ 1.0 & 72.7 $\pm$ 3.8 & 63.1 $\pm$ 4.3 & - & - & -                                        & 59.7 $\pm$ 1.8 & 56.3 $\pm$ 2.2 & 58.3 $\pm$ 1.9  \\
\emph{sub-activity only}                    & - & -  & - &                                      71.9 $\pm$ 0.8  & 60.9 $\pm$ 2.2 & 51.9 $\pm$ 0.9 & 27.4 $\pm$ 5.2 & 31.8 $\pm$ 6.3 & 27.7 $\pm$ 5.3  \\
\emph{no temporal interactions}             & 87.0 $\pm$ 0.8 & 79.8 $\pm$ 3.6 & 66.1 $\pm$ 1.5 & 76.0 $\pm$ 0.6 & 74.5 $\pm$ 3.5 & 66.7 $\pm$ 1.4 & 81.4 $\pm$ 1.3 & 83.2 $\pm$ 1.2 & 80.8 $\pm$ 1.4 \\ 
\emph{no object interactions}                & 88.4 $\pm$ 0.9 & 75.5 $\pm$ 3.7 & 63.3 $\pm$ 3.4 & 85.3 $\pm$ 1.0 & 79.6 $\pm$ 2.4 & 74.6 $\pm$ 2.8 & 80.6 $\pm$ 2.6 & 81.9 $\pm$ 2.2 & 80.0 $\pm$ 2.6 \\
\emph{full model}                           & 91.8 $\pm$ 0.4 & 90.4 $\pm$ 2.5 & 74.2 $\pm$ 3.1 & 86.0 $\pm$ 0.9 & 84.2 $\pm$ 1.3 & 76.9 $\pm$ 2.6 & 84.7 $\pm$ 2.4 & 85.3 $\pm$ 2.0 & 84.2 $\pm$ 2.5   \\
\emph{full model with tracking}                     & 88.2 $\pm$ 0.6 & 74.5 $\pm$ 4.3 & 64.9 $\pm$ 3.5 & 82.5 $\pm$ 1.4 & 72.9 $\pm$ 1.2 & 70.5 $\pm$ 3.0 & 79.0 $\pm$ 4.7  & 78.6 $\pm$ 4.1 & 78.3 $\pm$ 4.9  \\
\hline
\multicolumn{10}{c}{}\\
\multicolumn{10}{c}{Full model, \emph{without} assuming any ground-truth temporal segmentation is given.}\\
\hline
\emph{full, 1 segment.~(best)} 		& 83.1 $\pm$ 1.1 & 70.1 $\pm$ 2.3 & 63.9 $\pm$ 4.4 & 66.6 $\pm$ 0.7 & 62.0 $\pm$ 2.2 & 60.8 $\pm$ 4.5 & 77.5 $\pm$ 4.1 & 80.1 $\pm$ 3.9 & 76.7 $\pm$ 4.2  \\
\emph{full, 1 segment.~(averaged)} 		& 81.3 $\pm$ 0.4 & 67.8 $\pm$ 1.1 & 60.0 $\pm$ 0.8 & 64.3 $\pm$ 0.7 & 63.8 $\pm$ 1.1 & 59.1 $\pm$ 0.5 & 79.0 $\pm$ 0.9 & 81.1 $\pm$ 0.8 & 78.3 $\pm$ 0.9  \\
\emph{full, multi-seg learning }         & 83.9 $\pm$ 1.5 & 75.9 $\pm$ 4.6 & 64.2 $\pm$ 4.0 & 68.2 $\pm$ 0.3 & 71.1 $\pm$ 1.9 & 62.2 $\pm$ 4.1 & 80.6 $\pm$ 1.1  & 81.8 $\pm$ 2.2 & 80.0 $\pm$  1.2 \\
\emph{full, multi-seg learning + tracking }   & 79.4 $\pm$ 0.8 & 62.5 $\pm$ 5.4 & 50.2 $\pm$ 4.9 & 63.4 $\pm$ 1.6 & 65.3 $\pm$ 2.3 & 54.0 $\pm$ 4.6 & 75.0 $\pm$ 4.5 & 75.8 $\pm$ 4.4 & 74.2 $\pm$ 4.6  \\
\whline{1.1pt}
\end{tabular}
}
}
 \vskip .1in
\end{table*}

\begin{figure*}[tb!]
 \centering 
 \begin{minipage}[t]{0.34\linewidth}
\centering
  \includegraphics[width=\linewidth,height=1.55in]{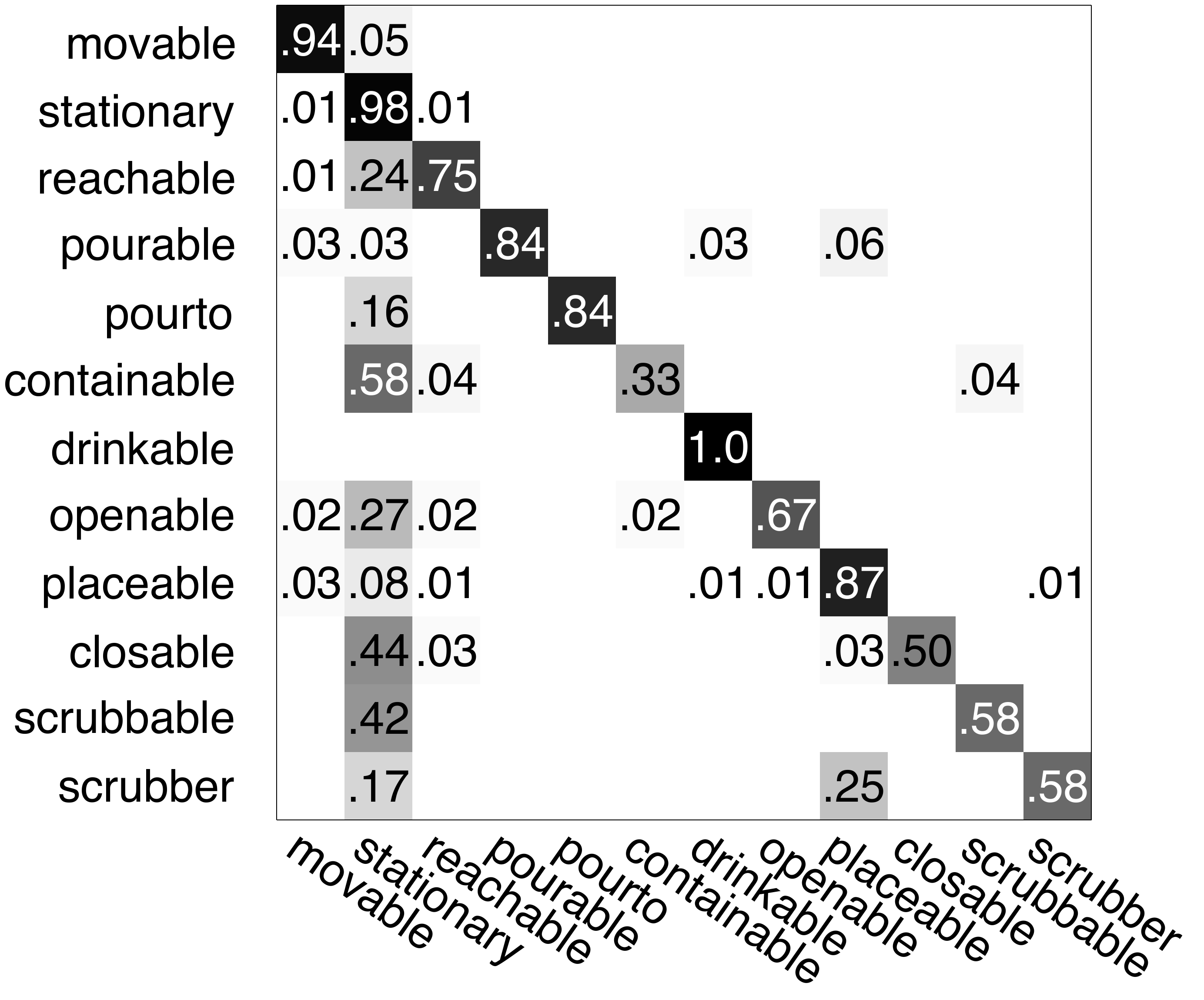}
\end{minipage}
 \begin{minipage}[t]{0.3\linewidth}
 \vskip -1.55in
\centering 
 \includegraphics[width=\linewidth,height=1.5in]{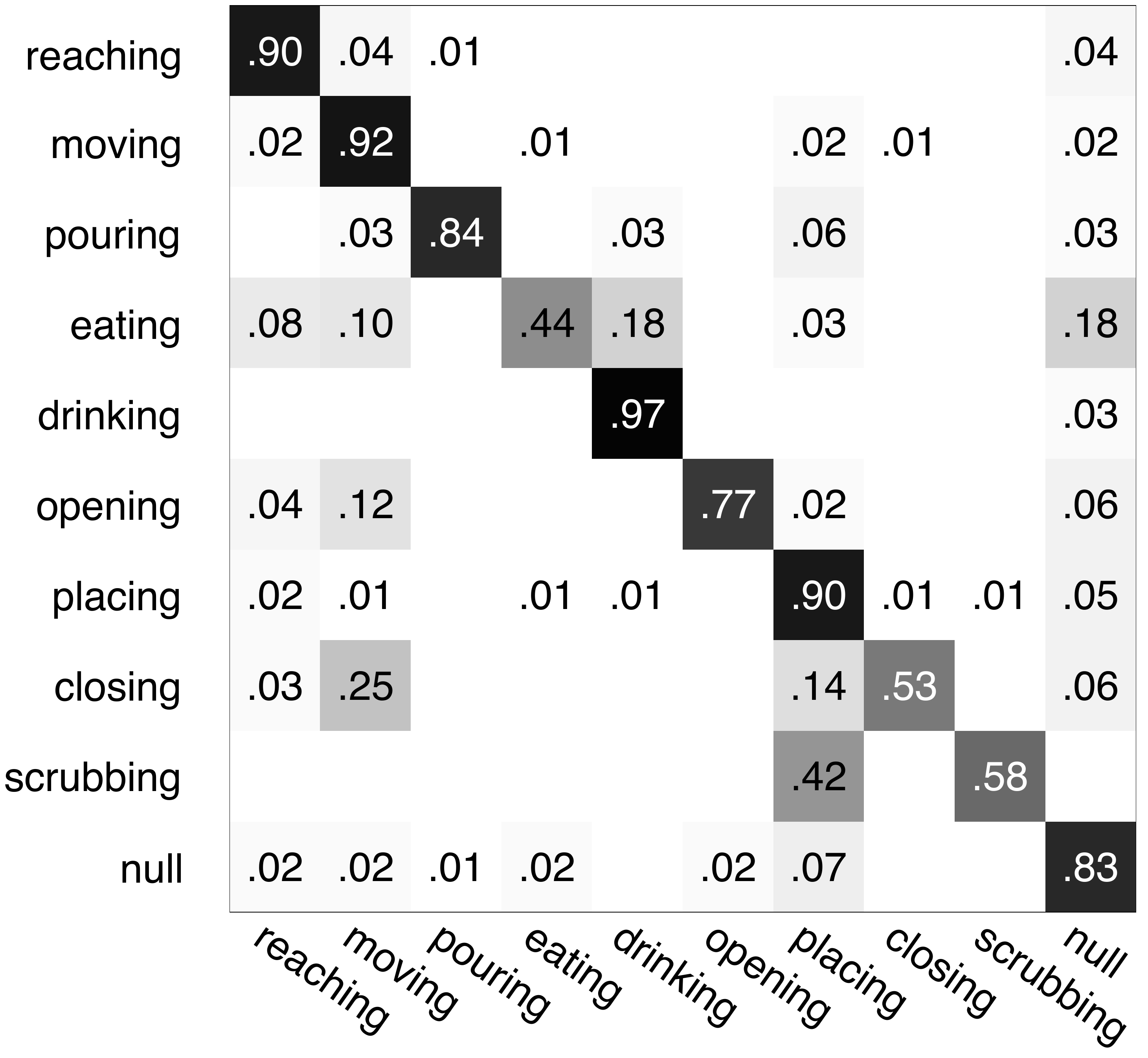}
 \end{minipage}
 \begin{minipage}[t]{0.34\linewidth}
\vskip -1.55in
\centering
 \includegraphics[width=\linewidth,height=1.58in]{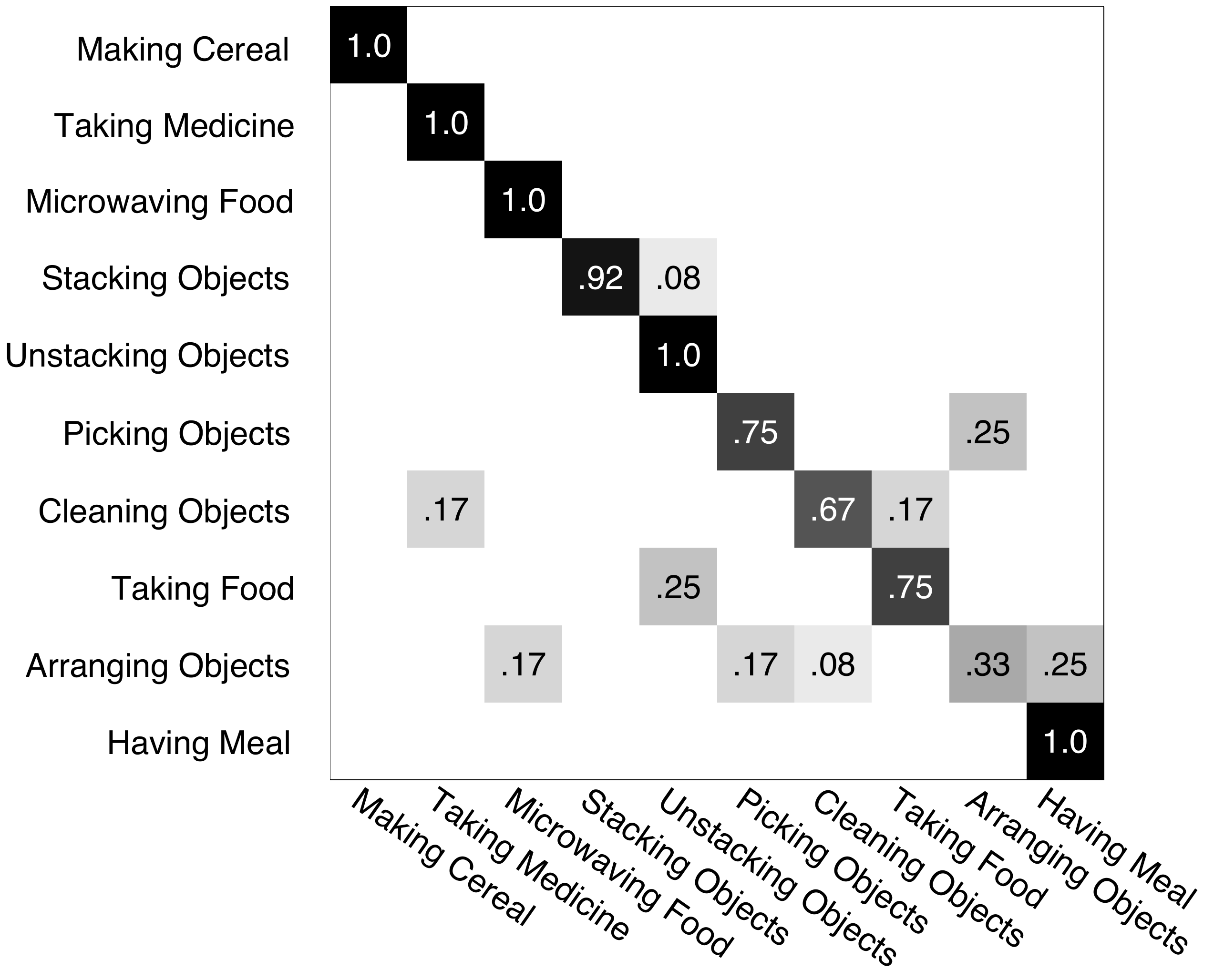}
 \end{minipage}
 \caption{Confusion matrix for affordance labeling (left), sub-activity labeling (middle) and high-level
  activity labeling (right) of the test RGB-D videos.}
 \label{fig:ActivityconfusionMatix}
 \end{figure*}

\subsection{Labeling results on the Cornell Activity Dataset 120 (CAD-120)}
\label{sec:cad120results}
Table \ref{tbl:labeling_result} shows the performance of various models on object affordance, sub-activity and high-level activity labeling. 
These results are obtained using 4-fold cross-validation and averaging performance across 
the folds. Each fold constitutes the activities performed by one subject, therefore the model is trained on activities of 
three subjects and \emph{tested on a new subject}. We report both the micro and macro averaged precision and recall over various classes along with standard error. Since our algorithm can only predict one label for each segment, micro precision and recall are same as the percentage of correctly classified segments. Macro precision and recall are  the averages of precision and recall respectively for all classes.

\begin{figure*}[t!]
\begin{minipage}[t]{\linewidth}
\begin{minipage}[t]{0.16\linewidth}
\centering
\includegraphics[width=\linewidth,height=0.8in]{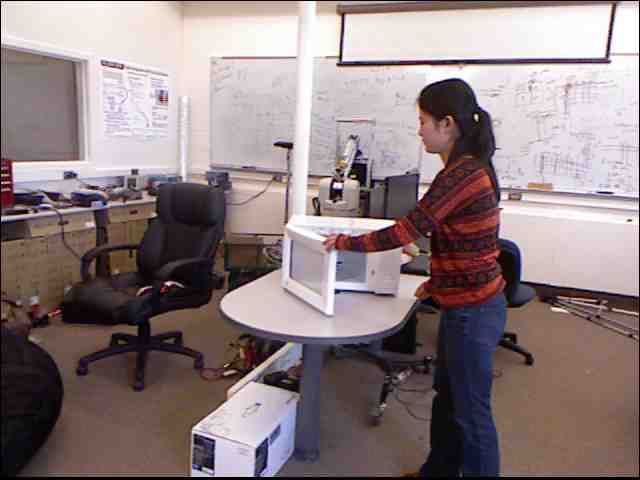}
{\scriptsize Subject \emph{opening}  \emph{openable} object1}
\end{minipage}
\begin{minipage}[t]{0.16\linewidth}
\centering
\includegraphics[width=\linewidth,height=0.8in]{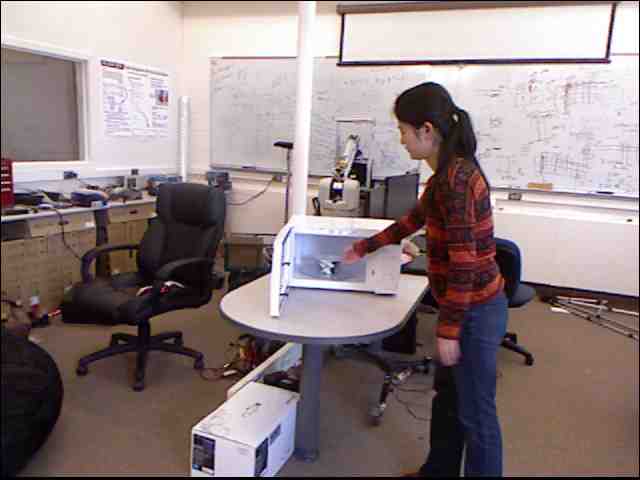}
{\scriptsize Subject \emph{reaching}  \emph{reachable} object2}

\end{minipage}
\begin{minipage}[t]{0.16\linewidth}
\centering
\includegraphics[width=\linewidth,height=0.8in]{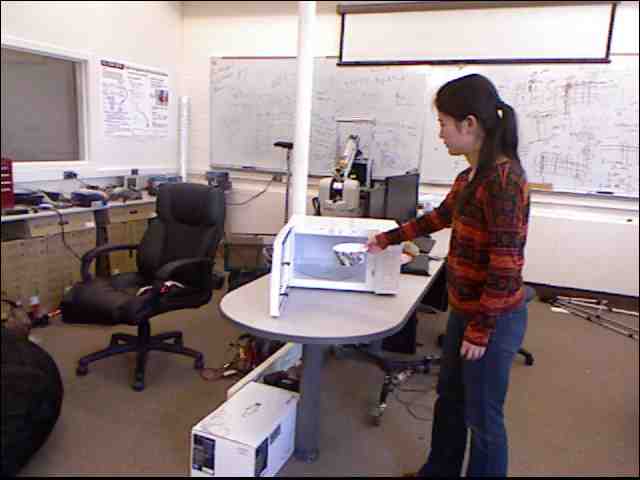}
{\scriptsize Subject \emph{moving} \emph{movable} object2}

\end{minipage}
\begin{minipage}[t]{0.16\linewidth}
\centering
\includegraphics[width=\linewidth,height=0.8in]{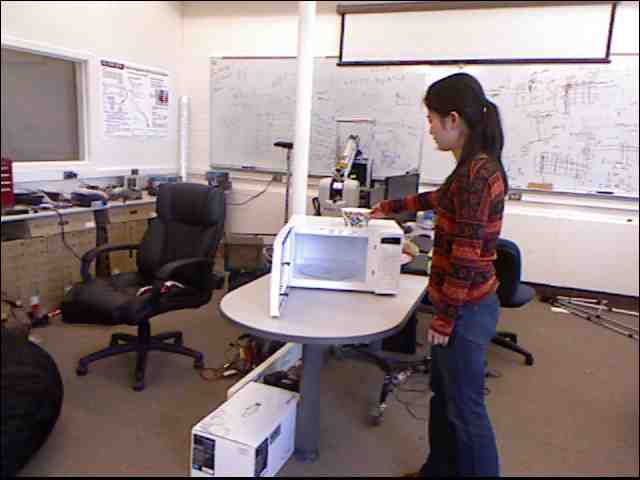}
{\scriptsize Subject \emph{placing}  \emph{placable} object2}
\end{minipage}
\begin{minipage}[t]{0.16\linewidth}
\centering
\includegraphics[width=\linewidth,height=0.8in]{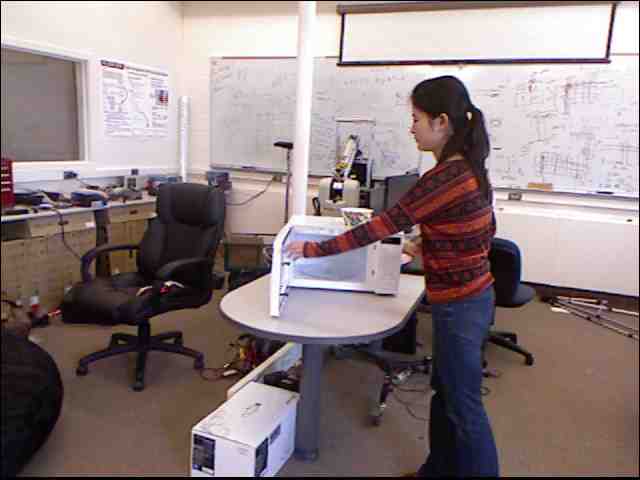}
{\scriptsize Subject \emph{reaching}  \emph{reachable} object1}
\end{minipage}
\begin{minipage}[t]{0.16\linewidth}
\centering
\includegraphics[width=\linewidth,height=0.8in]{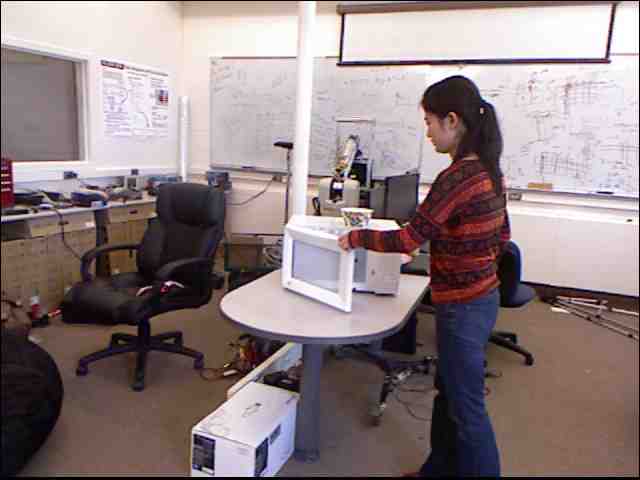}
{\scriptsize Subject \emph{closing}  \emph{closable} object1}
\end{minipage}
 \end{minipage}
 \vskip 0.048in
 
 \begin{minipage}[t]{\linewidth}
\begin{minipage}[t]{0.16\linewidth}
\centering
\includegraphics[width=\linewidth,height=0.8in]{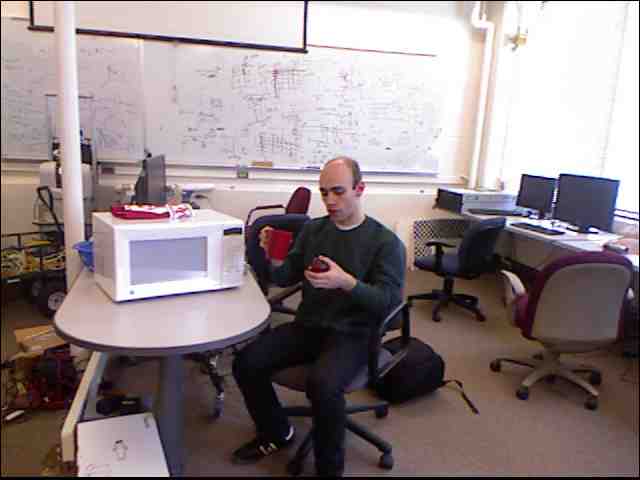}
{\footnotesize Subject \emph{moving} \emph{movable} object1}
\end{minipage}
\begin{minipage}[t]{0.16\linewidth}
\centering
\includegraphics[width=\linewidth,height=0.8in]{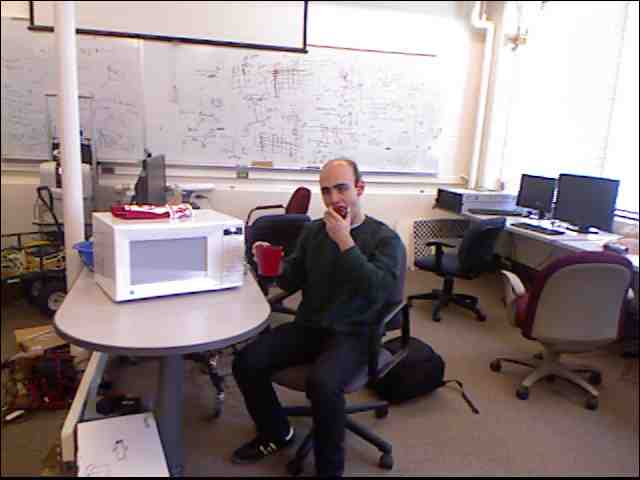}
{\footnotesize Subject \emph{eating} }
\end{minipage}
\begin{minipage}[t]{0.16\linewidth}
\centering
\includegraphics[width=\linewidth,height=0.8in]{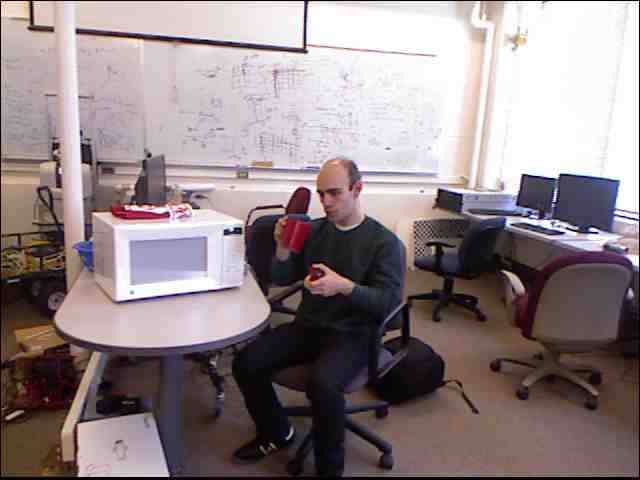}
{\footnotesize Subject \emph{moving} \emph{movable} object1}
\end{minipage}
\begin{minipage}[t]{0.16\linewidth}
\centering
\includegraphics[width=\linewidth,height=0.8in]{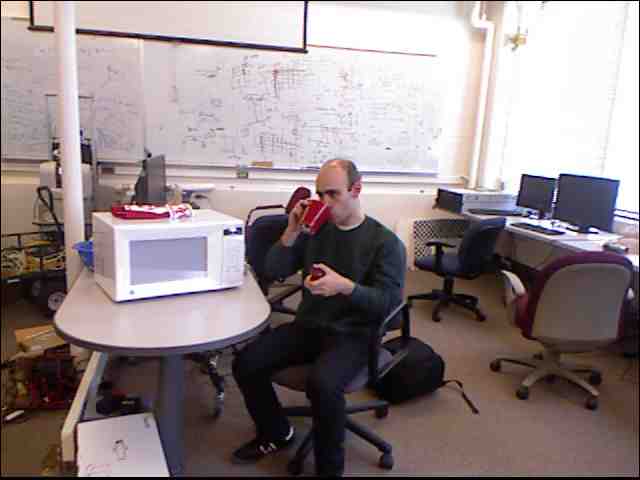}
{\footnotesize Subject \emph{drinking} from \emph{drinkable} object1 }
\end{minipage}
\begin{minipage}[t]{0.16\linewidth}
\centering
\includegraphics[width=\linewidth,height=0.8in]{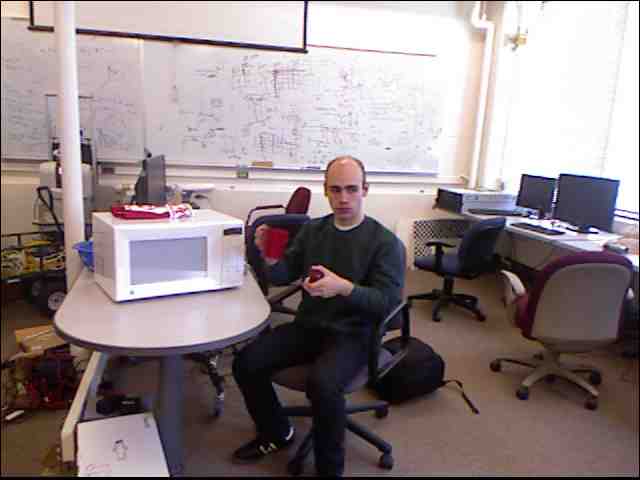}
{\footnotesize Subject \emph{moving} \emph{movable} object1}
\end{minipage}
\begin{minipage}[t]{0.16\linewidth}
\centering
\includegraphics[width=\linewidth,height=0.8in]{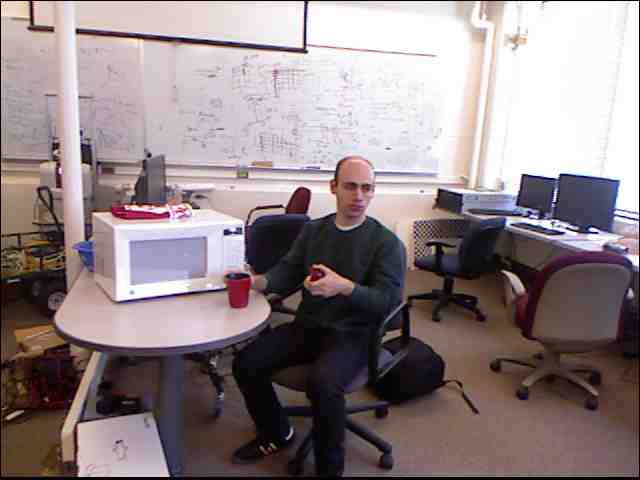}
{\footnotesize Subject \emph{placing} \emph{placeable} object1}
\end{minipage}
 \end{minipage}
 \vskip 0.048in
 
 \begin{minipage}[t]{\linewidth}
 \begin{minipage}[t]{0.16\linewidth}
\centering
\includegraphics[width=\linewidth,height=0.8in]{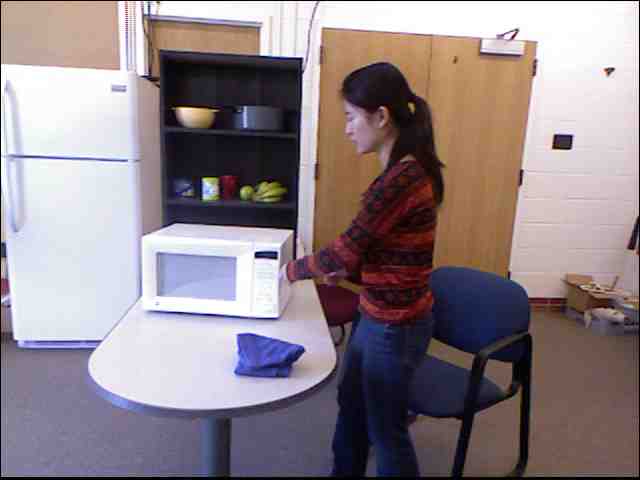}
{\footnotesize Subject \emph{reaching} \emph{reachable} object1 }
\end{minipage}
\begin{minipage}[t]{0.16\linewidth}
\centering
\includegraphics[width=\linewidth,height=0.8in]{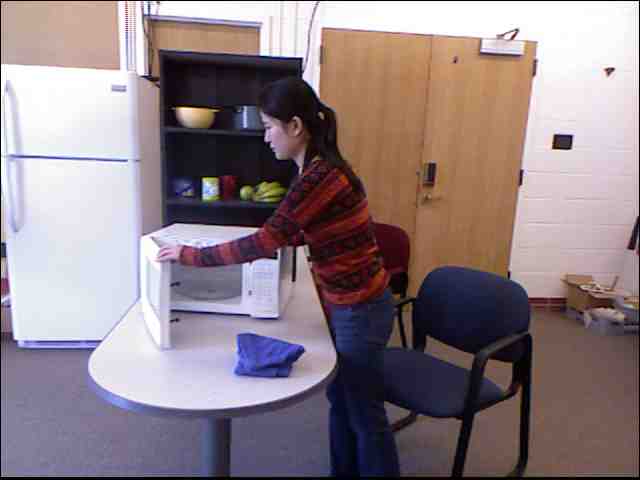}
{\footnotesize Subject \emph{opening}  \emph{openable} object1 }
\end{minipage}
\begin{minipage}[t]{0.16\linewidth}
\centering
\includegraphics[width=\linewidth,height=0.8in]{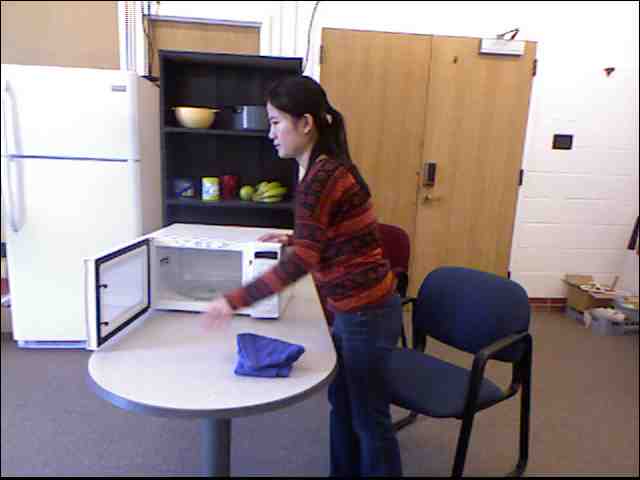}
{\footnotesize Subject \emph{reaching} }
\end{minipage}
\begin{minipage}[t]{0.16\linewidth}
\centering
\includegraphics[width=\linewidth,height=0.8in]{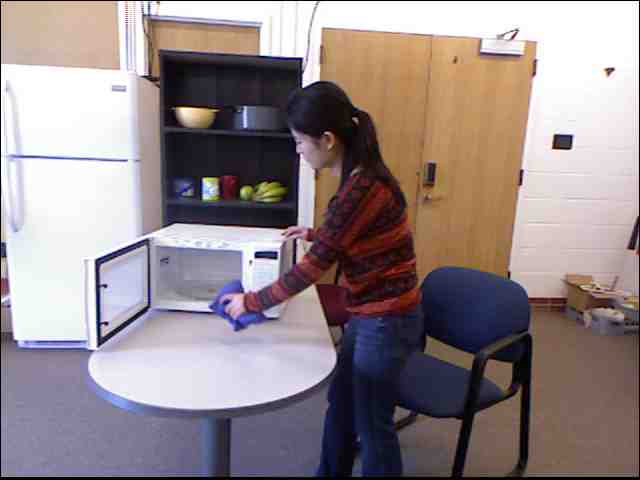}
{\footnotesize Subject \emph{moving} \emph{movable} object2}
\end{minipage}
\begin{minipage}[t]{0.16\linewidth}
\centering
\includegraphics[width=\linewidth,height=0.8in]{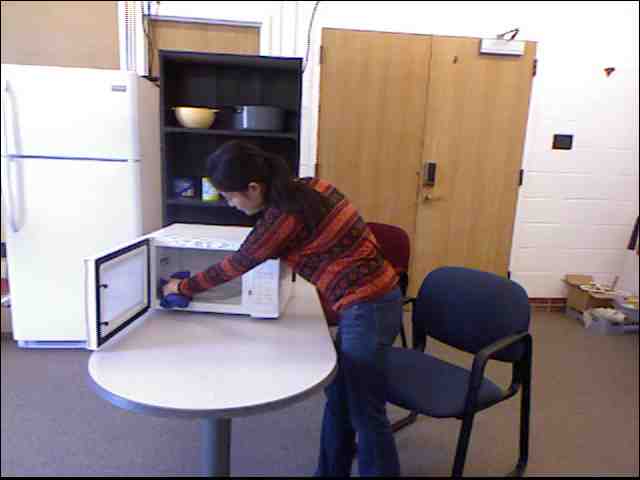}
{\footnotesize Subject \emph{scrubbing} \emph{scrubbable} object1 with \emph{scrubber} object2  }
\end{minipage}
\begin{minipage}[t]{0.16\linewidth}
\centering
\includegraphics[width=\linewidth,height=0.8in]{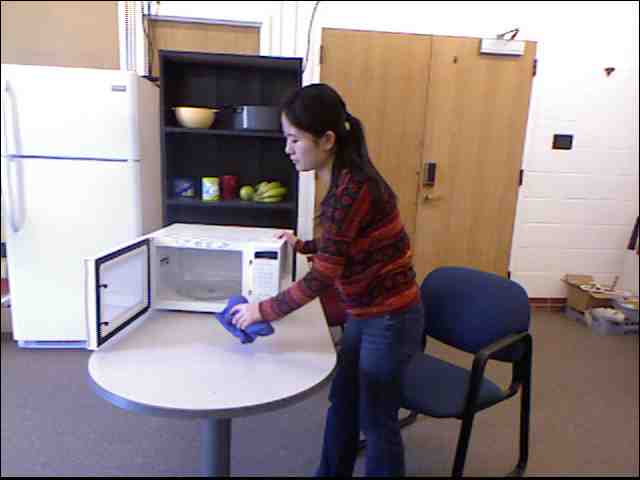}
{\footnotesize Subject \emph{moving} \emph{movable} object2}
\end{minipage}
 \end{minipage}
 
  \caption{Descriptive output of our algorithm: Sequence of images from the \emph{taking food} (Top Row), \emph{having meal} (Middle Row) and \emph{cleaning objects} (Bottom Row) activities labeled with sub-activity and object affordance labels. 
 A single frame is sampled from the temporal segment to represent it.}
\label{fig:labelingresult}
 \end{figure*}

Assuming ground-truth temporal segmentation is given, the results for our \emph{full model} are
shown in Table~\ref{tbl:labeling_result} on line 10, its variations on lines 5--9 and the baselines on lines 1--4.
 The results in lines 11--14 correspond to the case when temporal segmentation is not assumed. 
In comparison to a basic SVM multiclass model \citep{joachims2009cutting} (referred to as \emph{SVM multiclass} when using all features and \emph{image only} when using only image features), which is equivalent to only considering the nodes in our MRF without any edges, our model performs significantly better.
 We also compare with the high-level activity classification results obtained from the method presented in 
 \cite{SungICRA2012}. We ran their code on our dataset and obtain accuracy of 26.4\%, whereas our method gives  an 
 accuracy of 84.7\% when ground truth segmentation is available and 80.6\% otherwise.   
Figure \ref{fig:labelingresult} shows a sequence of images from \emph{taking food} activity along with the inferred labels.
Figure \ref{fig:ActivityconfusionMatix} shows the confusion matrix for labeling affordances, sub-activities and high-level 
activities with our proposed method. We can see that there is a strong diagonal with a few errors such as \emph{scrubbing} misclassified as \emph{placing}, and \emph{picking objects} misclassified as \emph{arranging objects}.

 We analyze our model to gain insight into which interactions provide useful information 
 by comparing our full model to variants of our model.

{\bf How important is object context for activity detection?} We show the importance of object context for sub-activity labeling by learning a variant of our model without the object nodes (referred to as \emph{sub-activity only}).   With object context, the micro precision increased by 14.1\% and both macro precision and recall increased by around 23.3\% over \emph{sub-activity only}.
Considering object information (affordance labels and occlusions) also improved the high-level 
activity accuracy by three-fold.    

{ \bf How important is activity context for affordance detection?} We also show the importance of context from sub-activity for affordance detection by learning our model without the sub-activity nodes (referred to as \emph{object only}).  With sub-activity context, the micro precision increased by 4.9\% and the macro precision and recall increased by 17.7\%
and 11.1\% respectively for affordance labeling over \emph{object only}.
The relative gain is less compared with that obtained in sub-activity detection as the \emph{object only} 
model still has object--object context which helps in affordance detection.    

{\bf How important is object--object context for affordance detection?}  In order to study the 
effect of the object--object  interactions for affordance detection, we learnt our model without the object-object edge potentials (referred to as \emph{no object interactions}). We see a considerable improvement in affordance detection when the object interactions are modeled, the macro recall increased by 14.9\% and the macro precision by about 10.9\%. This shows that sometimes just the context from the human activity alone is not sufficient to determine the affordance of an object.   

{ \bf How important is temporal context?} We also learn our model without the temporal edges (referred to as \emph{no temporal interactions}). Modeling temporal interactions increased the micro precision by 4.8\% and 10.0\% for affordances and sub-activities respectively and increased the micro precision for high-level activity by 3.3\%.

{\bf How important is reliable human pose detection?}   
In order understand the effect of the errors in human pose tracking,
we consider the affordances that require direct contact by human hands,
such as movable, openable, closable, drinkable, etc.
The distance of the predicted hand locations to the object should be
zero at the time of contact.  We found that for the correct
predictions, these distances had a mean of 3.8 cm
and variance of 48.1 cm. However, for the incorrect predictions, these
distances had a mean that was 43.3\% higher and a variance that was
53.8\% higher.  This indicates that
the prediction accuracies can potentially be improved with more robust
human pose tracking.


{\bf How important is reliable object tracking?} We show the effect of having reliable 
object tracking by comparing to the results obtained from using our object tracking 
algorithm mentioned in Section \ref{sec:objtracking}. We see that using the object tracks
generated by our algorithm gives slightly lower micro precision/recall values compared with 
using ground-truth object tracks, around 3.5\% drop in affordance and sub-activity detection,
and 5.7\% drop in high-level activity detection. The drop in macro precision and recall are higher,
which shows that the performance of few classes are effected more than the others.  
In future work, one can increase accuracy by improving object tracking.

\begin{figure*}[t]
 \centering
 \includegraphics[width=\linewidth,height=2in]{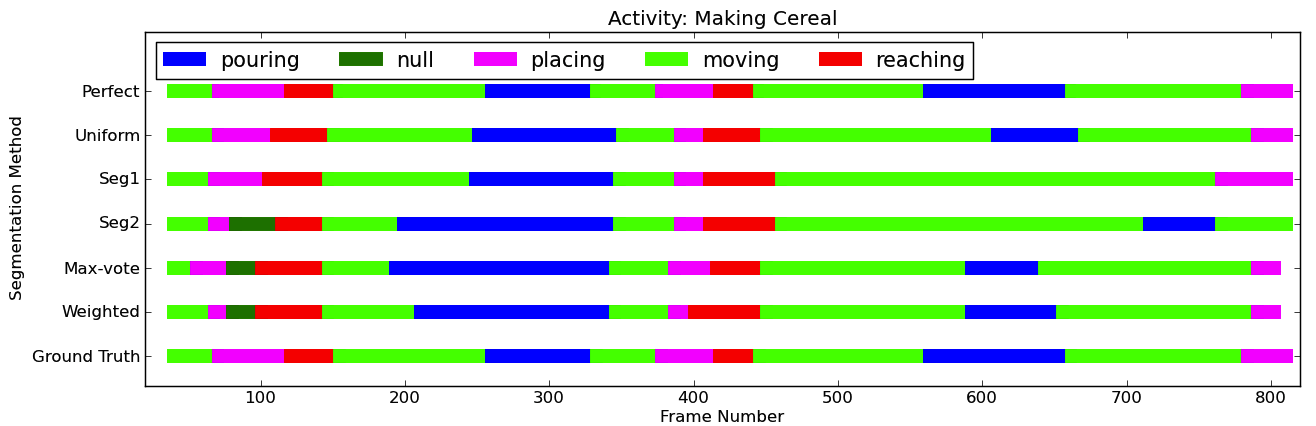}
 \caption{Comparison of the sub-activity labeling of various segmentations. This activity involves the sub-activities: \emph{reaching}, \emph{moving}, \emph{pouring} and \emph{placing} as colored in red, green, blue and magenta respectively. The x-axis denotes the time axis numbered with frame numbers. It can be seen that the various individual segmentation labelings are not perfect and make different mistakes, but our method for merging these segmentations selects the correct label for many frames. 
 }
 \label{fig:activityresult}
 \end{figure*}

\medskip
\noindent
{\bf Results with multiple segmentations.}
Given the RGB-D video and initial bounding boxes for objects in the first frame, we obtain the final labeling using our method described in 
Section \ref{sec:segmentation}. To generate the segmentation hypothesis set $\mathcal{H}$ we consider three different segmentation algorithms, and generate multiple segmentations by changing 
their parameters as described in Section \ref{sec:temporalseg}.
The lines 11--13 of Table \ref{tbl:labeling_result} show the results of the best performing segmentation,
average performance all the segmentations considered, and our proposed method for 
combining the segmentations respectively.  We see that our method improves the 
performance over considering a single best performing segmentation: macro precision 
increased by 5.8\% and 9.1\% for affordance and sub-activity labeling respectively. 
Fig.~\ref{fig:activityresult} shows the comparison of the sub-activity labeling of various segmentations, our end-to-end labeling and the ground-truth labeling for one 
\emph{making cereal} high-level activity video. 
It can be seen that the various individual segmentation labelings are not perfect 
and make different mistakes, but our method for merging these segmentations selects the 
correct label for many frames.
Line 14 of Table~\ref{tbl:labeling_result} show the results of our proposed method for 
combining the segmentations along with using our object tracking algorithm. The 
numbers show a drop compared with the case of using ground-truth tracks, therefore 
providing a scope for improvement by using more reliable tracking algorithms.

\subsection{Robotic Applications}
\label{sec:robotapp}
We demonstrate the use of our learning algorithm in two robotics applications.
First, we show that the knowledge of the activities currently being performed enables
a robot to assist the human by performing an appropriate response action.
Second, we show that the knowledge of the affordances of the objects enables
a robot to use them appropriately when manipulating them.



We use Cornell's Kodiak, a PR2 robot, in our experiments.
Kodiak is mounted with a Microsoft Kinect, which is used as the main input sensor 
to obtain the RGB-D video stream.
We used the OpenRAVE libraries \citep{diankov_thesis}
for programming the robot to perform the pre-programmed assistive tasks.

\begin{figure*}[t!]
\includegraphics[width=0.16\linewidth,height=0.8in]{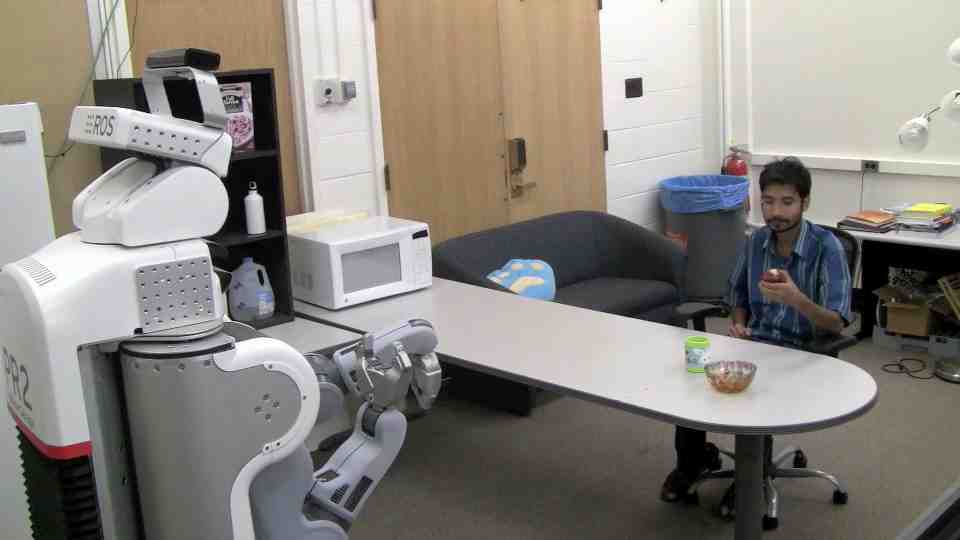}
\includegraphics[width=0.16\linewidth,height=0.8in]{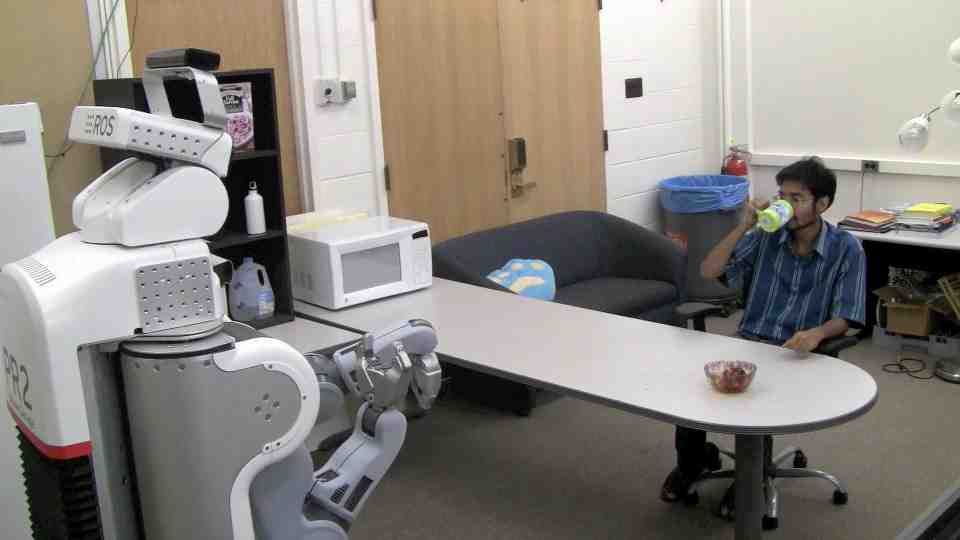}
\includegraphics[width=0.16\linewidth,height=0.8in]{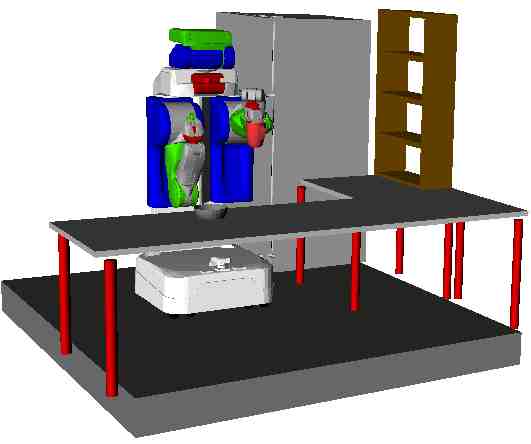}
\includegraphics[width=0.16\linewidth,height=0.8in]{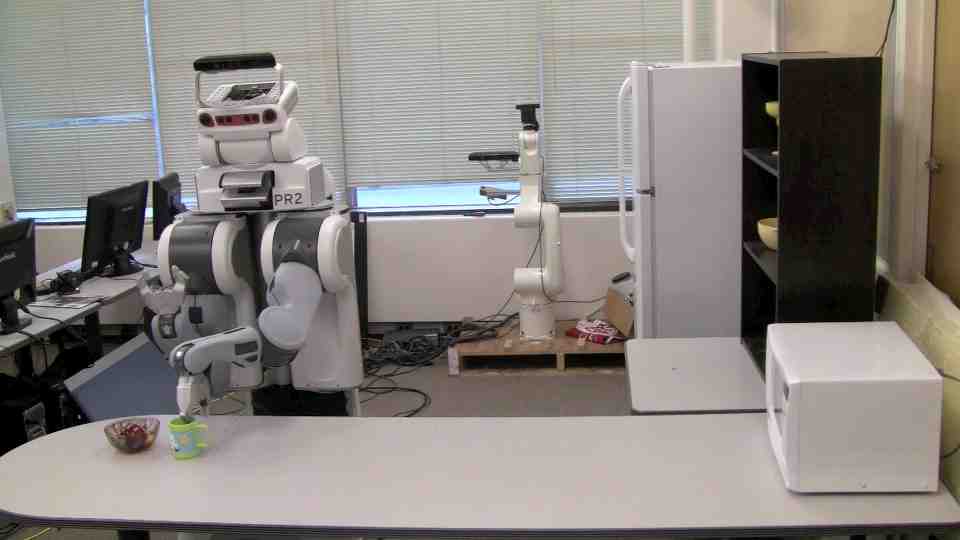}
\includegraphics[width=0.16\linewidth,height=0.8in]{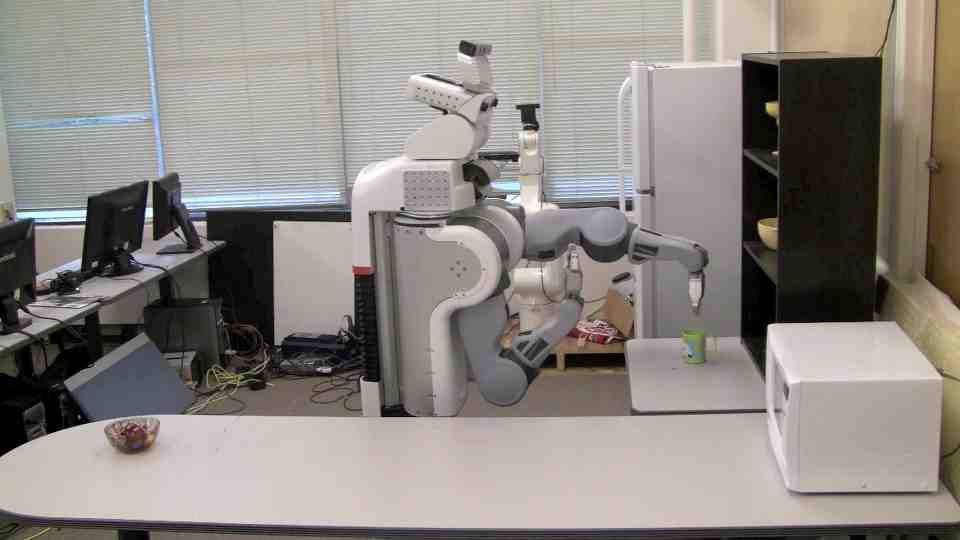}
\includegraphics[width=0.16\linewidth,height=0.8in]{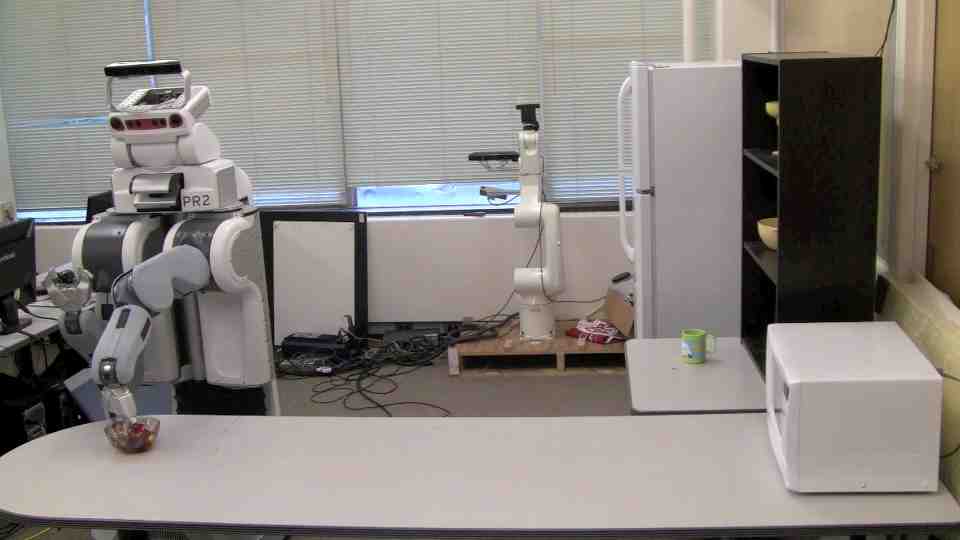}
\vskip 0.048in
\includegraphics[width=0.16\linewidth,height=0.8in]{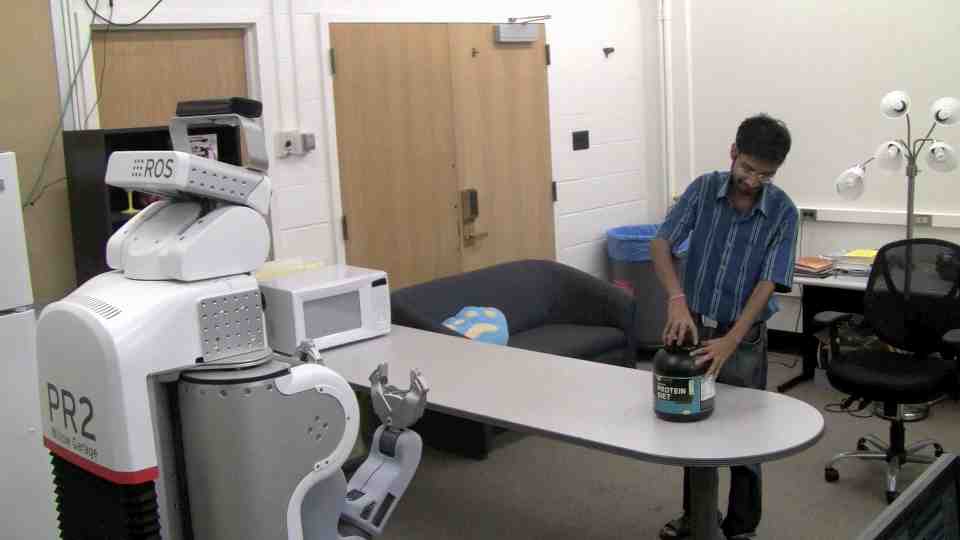}
\includegraphics[width=0.16\linewidth,height=0.8in]{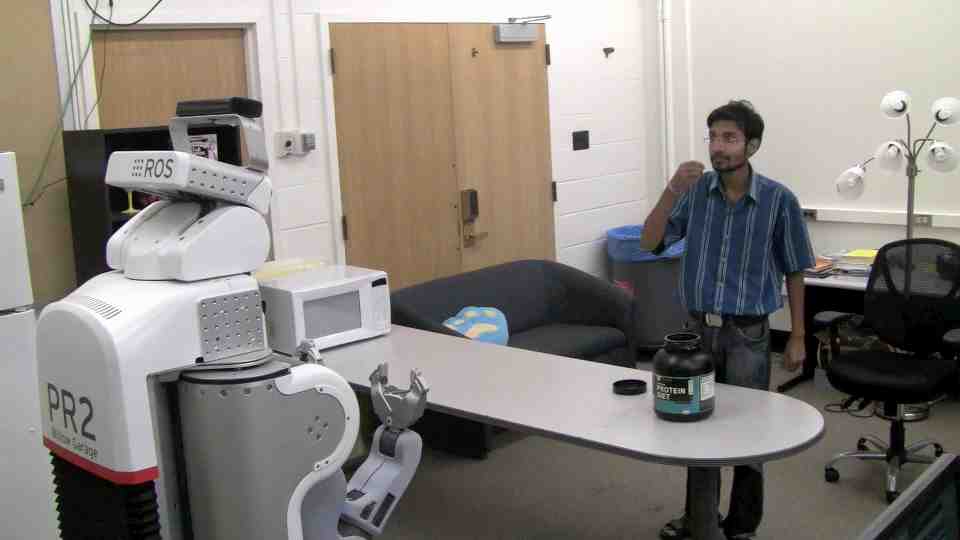}
\includegraphics[width=0.16\linewidth,height=0.8in]{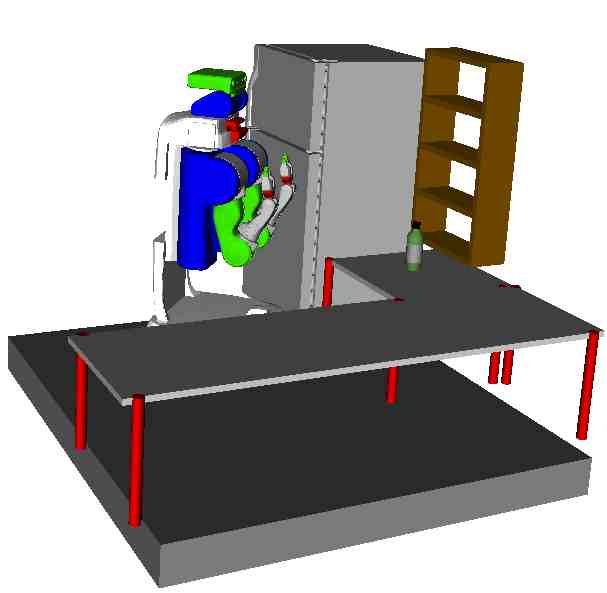}
\includegraphics[width=0.16\linewidth,height=0.8in]{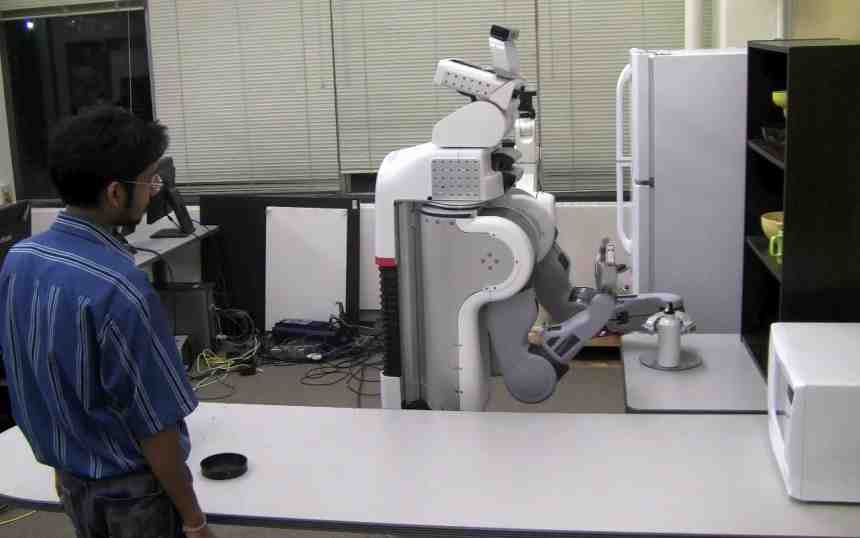}
\includegraphics[width=0.16\linewidth,height=0.8in]{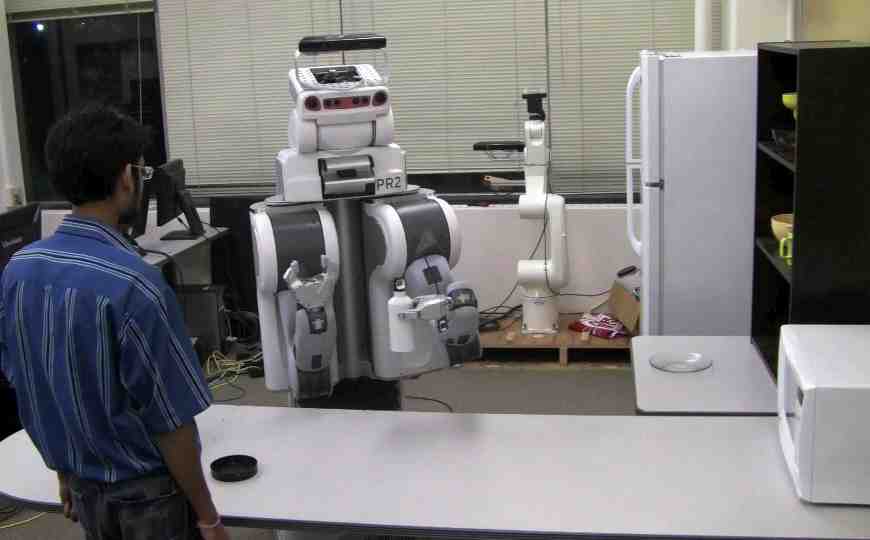}
\includegraphics[width=0.16\linewidth,height=0.8in]{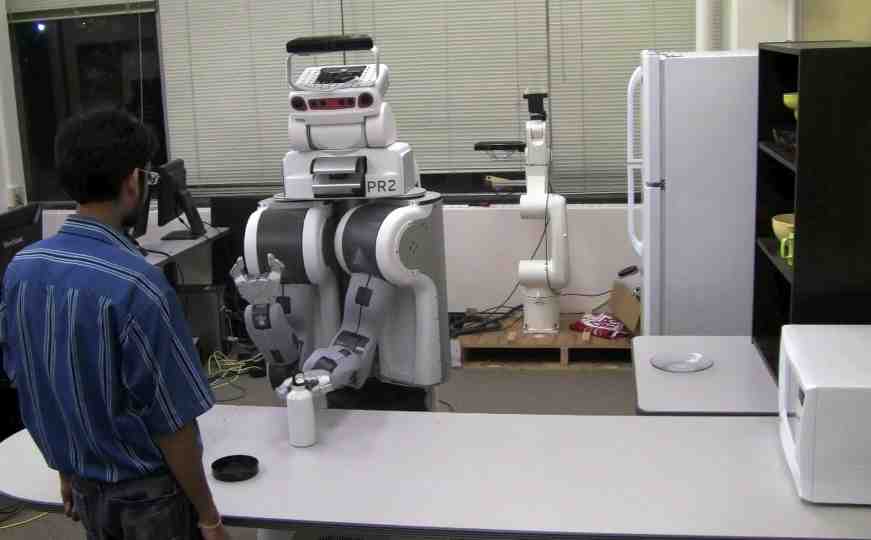}
\vskip 0.048in
\includegraphics[width=0.16\linewidth,height=0.8in]{images/responses/cereal/act1.jpg}
\includegraphics[width=0.16\linewidth,height=0.8in]{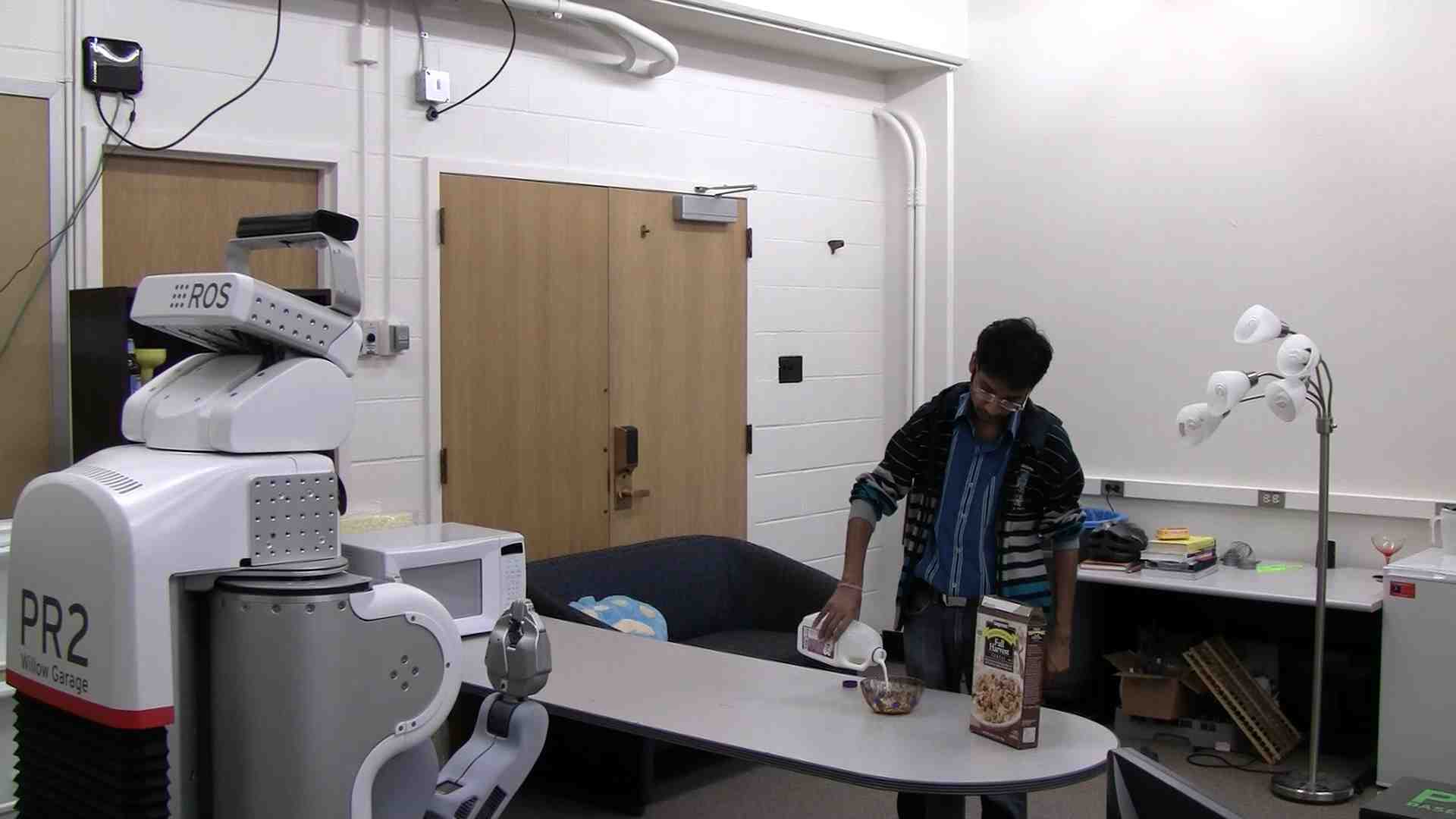}
\includegraphics[width=0.16\linewidth,height=0.8in]{images/responses/cereal/sim1.jpg}
\includegraphics[width=0.16\linewidth,height=0.8in]{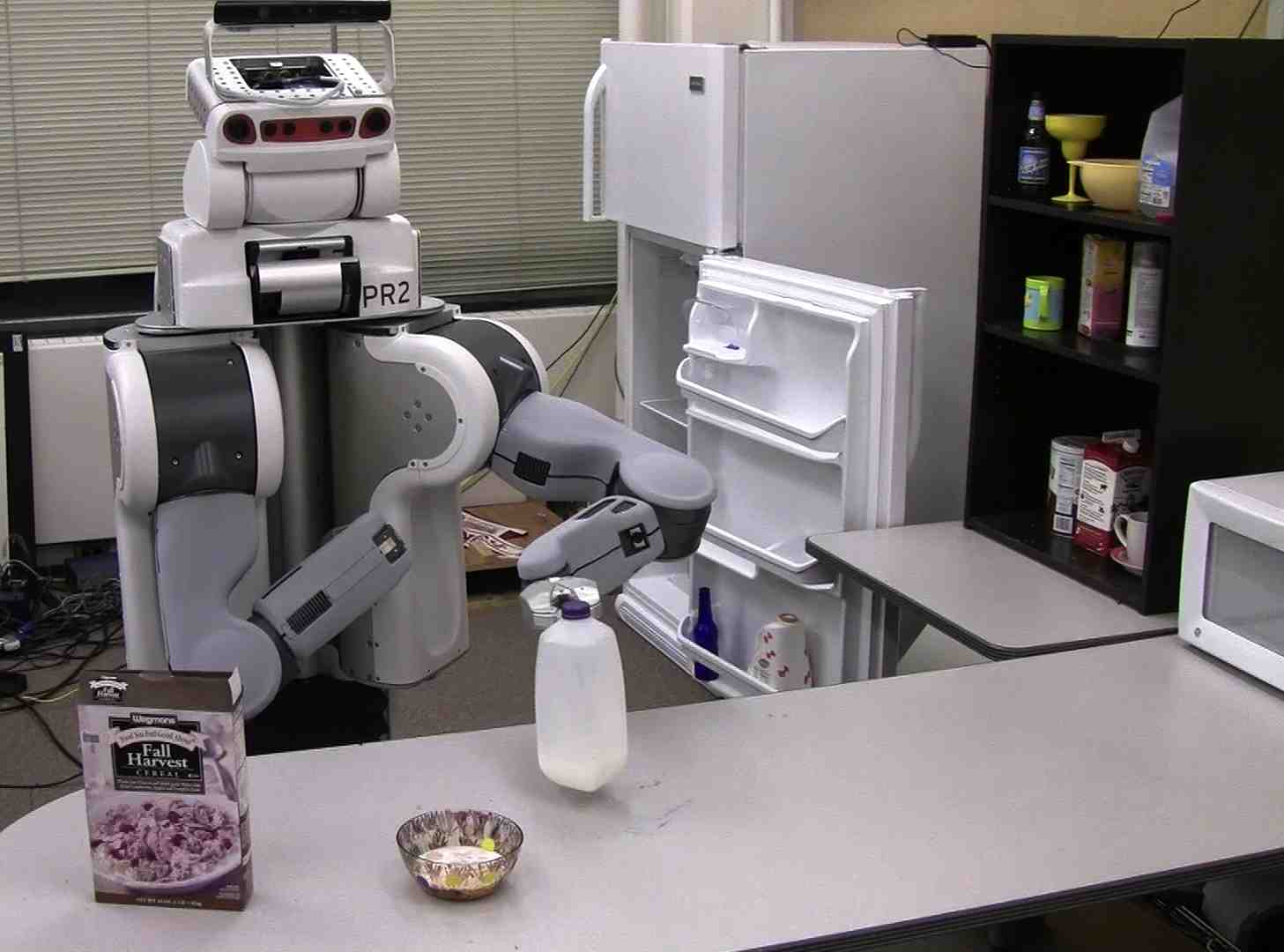}
\includegraphics[width=0.16\linewidth,height=0.8in]{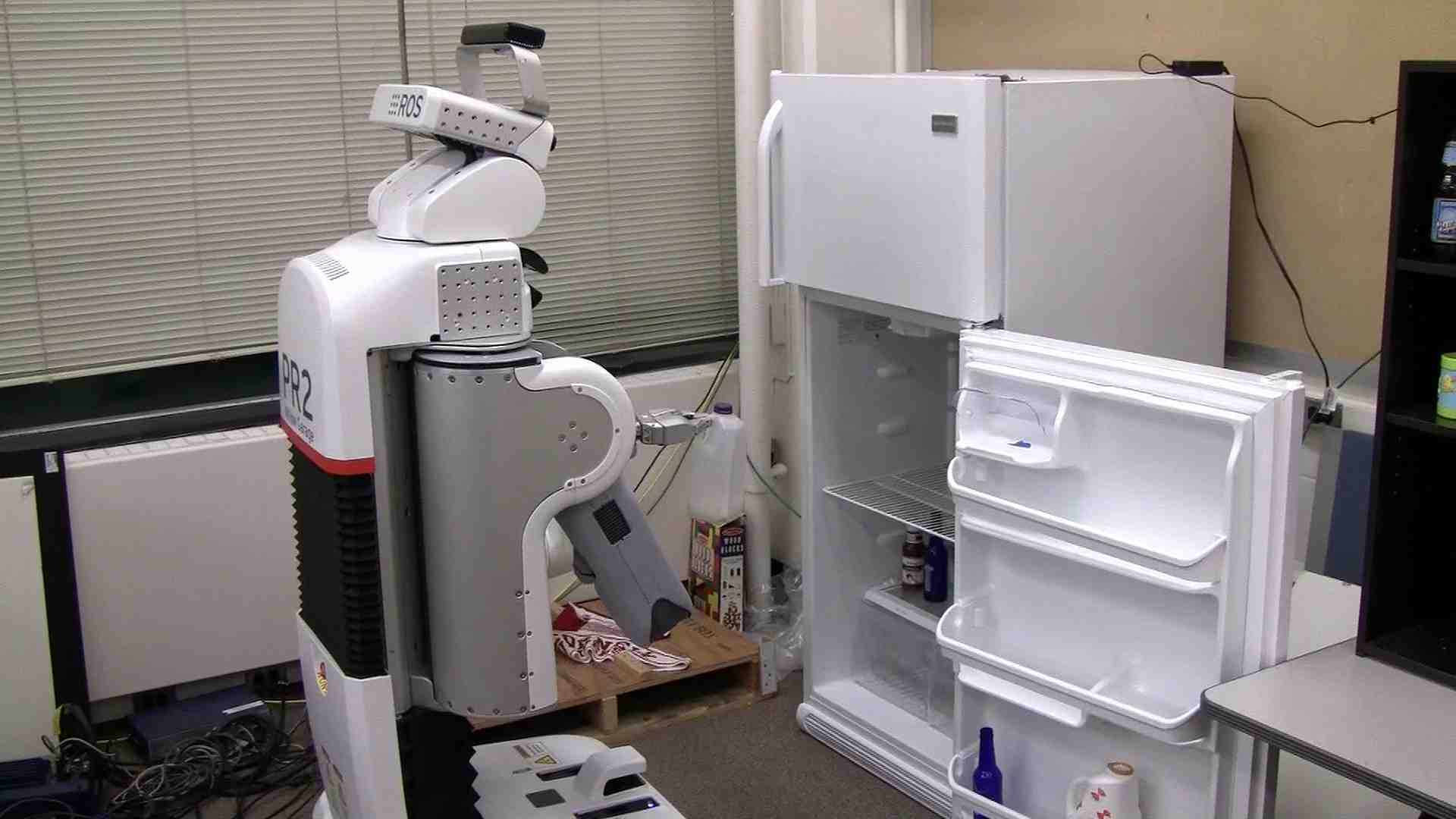}
\includegraphics[width=0.16\linewidth,height=0.8in]{images/responses/cereal/res3.jpg}\\
 \caption{Robot performing the task of assisting humans: (top row) robot clearing the table after detecting \emph{having a meal} activity, (middle row) robot fetching a bottle of water after detecting \emph{taking a medicine} activity and (third row) robot putting milk in the fridge after detecting \emph{making cereal} activity. First two columns show the robot observing the activity, third row shows the robot planning the response in simulation and the last three columns show the robot performing the response action.  }
\label{fig:robotresponse}
 \end{figure*}

\smallskip
\noindent
{\bf Assisting Humans.}
There are several modes of operation for a robot performing assistive tasks. For example, the
robot can perform some tasks completely autonomous, independent of the humans. 
For some other tasks, the robot needs to act more reactively. That is, depending on the task and
current human activity taking place, perform a complementary sub-task. For example, bring a glass 
of water when a person is attempting to take medicine (and there is no glass within person's reach).
Such a behavior is possible only when the activities are successfully detected.
In this experiment, we demonstrate that our algorithm for detecting the human activities enables
a robot to take such (pre-programmed) reactive actions.\footnote{Our goal in this paper is on activity
detection, therefore we pre-program the response actions using existing open-source tools in ROS. 
In future, one would need to make significant advances in several fields to make this
useful in practice, e.g., object detection \cite{koppula:Nips11,koppulaIJRR12}, grasping \cite{saxena2008roboticgrasping,jianggrasping}, human-robot interaction, and so on.}

We consider the following three scenarios: 
 \begin{itemize}
\item \emph{Having Meal}: The subject eats food from a bowl and drinks water from a cup in this activity.
 On detecting the \emph{having meal} activity, the robot assists by clearing the table (i.e. move the cup and the bowl to another place) after the subject finishes eating.
 \item \emph{Taking Medicine}: The subjects opens the medicine container, takes the medicine, and waits as there is no water nearby. The robot assists the subject by bringing a glass of water on detecting the \emph{taking medicine} activity.
 \item \emph{Making Cereal}: The subject prepares cereal by pouring cereal and milk in to a bowl. 
 On detecting the activity, the robot responds by taking the milk and putting it into the refrigerator. 
\end{itemize}
Our robot was placed in a kitchen environment so that it can observe the activity being performed 
by the subject. We found that our robot successfully detected the activities and performed 
the above described reactive actions. 
Fig.~\ref{fig:robotresponse} shows the sequence of images of the robot detecting the activity being performed, planning the response in simulation and then performing the appropriate response for all three activities described above.

\smallskip
\noindent
{\bf Using Affordances.}
An important component of our work is to learn affordances. In particular, by observing how the
humans interact with the objects, a robot can figure out the affordances of the objects. Therefore, it can use 
these inferred affordances to interact with objects in a meaningful way.
For example, given an instruction of 
`clear the table', the 
robot should be able to perform the response in a desirable way: move the 
bowl with cereal \emph{without} tilting it, and not move the microwave.
In this experiment, we demonstrate that our algorithm for labeling the affordances explicitly
helps in manipulation. 

In our setting, we directly infer 
the object affordances (movable, pourable, drinkable, etc.). 
Therefore, we only need to encode the low-level control 
actions of each affordance, e.g.
to move only \emph{movable} objects, and to 
execute constrained movement, i.e. no rotation in the xy plane, for objects with 
affordances such as \emph{pour-to}, \emph{pourable} or \emph{drinkable}, etc. 
The robot is allowed to observe various activities performed with the objects and it uses our learning algorithms to infer the affordances associated with the objects.  
When an instruction is given to the robot, such as
`clear the table' or `move object x', it uses the inferred affordances to perform the response.

We demonstrate this in two scenarios for the task of `clearing the table': 
detecting movable objects and detecting constrained movement. 
We consider a total of seven activities with nine unique objects. Some objects were used 
in multiple activities, with a total of 19 object instances.
Two of these activities were other high-level activities that were \emph{not seen} during training, 
but comprise sequences of the learned affordances and sub-activities. The results are summarized in 
Table \ref{tbl:robotaff}.
 
 \begin{table}[h]
\caption{{\bf Robot Object Manipulation Results}, showing the accuracy achieved by the Kodiak PR2 performing the two manipulation tasks, with and without multiple observations.}
 \label{tbl:robotaff}
 {\scriptsize
\newcolumntype{P}[2]{>{\footnotesize#1\hspace{0pt}\arraybackslash}p{#2}}
\setlength{\tabcolsep}{2pt}
\centering
\resizebox{\hsize}{!}
{

\begin{tabular}
{@{}p{0.26\linewidth} |P{\centering}{12mm}|P{\centering}{16mm}|P{\centering}{16mm}@{}}
\whline{1.1pt}
 task & \# instance & accuracy (\%) & accuracy (\%) (multi.~obsrv.) \\
\hline
object movement  & 19  & 100 & 100  \\
constrained  movement & 15 & 80 & 100  \\
\whline{1.1pt}
\end{tabular}
}
}
\end{table}

In the scenario of detecting movable objects, the robot was programmed to move only objects with 
inferred \emph{movable} affordance, to a specified location. There were a total of 
15 instances of movable objects. The robot was able to correctly identify all \emph{movable} objects using our 
model and could perform the moving task with 100\% accuracy. 

In the scenario of constrained movement, i.e. the robot should not tilt the objects 
which contain food items or liquids when moving them. In order to achieve this, we have programmed 
the robot to perform constrained movement without tilting the objects if it has inferred at least one of 
the following affordances: \{\emph{drinkable, pourable, pour-to}\}. The robot was able to correctly identify
constraint movement for 80\% of the movable instances. 
Also, if we let the robot observe the activities for a longer time, 
i.e. let the subject perform multiple activities with the objects and aggregate the affordances 
associated with the objects before performing the task, the robot is able to 
perform the task with 100\% accuracy. 

These experiments show that robot can use the affordances for manipulating the objects in a 
more meaningful way. Some affordances such as \emph{moving} are easy to detect, where as 
some complicated affordances such as \emph{pouring} might need more observations to be 
detected correctly. Also, by considering other high-level activities in addition to those used 
for learning, we have also demonstrated the generalizability of our algorithm for affordance detection. 

We have made the videos of our results, along with the CAD-120 dataset and code, available at 
\url{http://pr.cs.cornell.edu/humanactivities}

\section{Conclusion and Discussion}
\label{sec:conclusion}

In this paper, we have considered the task of jointly labeling human sub-activities and 
object affordances in order to obtain a descriptive labeling of the activities being performed in the
RGB-D videos. The activities we consider happen
over a long time period, and comprise several sub-activities performed in a sequence. 
We formulated
this problem as a MRF, and learned the parameters of the
model using a structural SVM formulation. Our model also incorporates the
temporal segmentation problem by computing multiple segmentations and
considering 
labeling 
over these segmentations as
latent variables. In extensive experiments over a challenging dataset, 
we show that our method 
achieves an accuracy of 79.4\% for affordance, 63.4\% for  sub-activity and 
75.0\% for high-level activity labeling on the activities performed by a different
 subject than those in the training set. We also showed that
 it is important to model the different properties (object affordances,
 object--object interaction, temporal interactions, etc.)  in order to 
 achieve good performance. We also demonstrate the use of our 
 activity and affordance labeling by a PR2 robot in the task of assisting 
 humans with their daily activities. We have shown that being able to
 infer affordance labels enables the robot to perform the tasks in a more 
 meaningful way.

In this growing area of RGB-D activity recognition, we have presented 
algorithms for activity and affordance detection and also demonstrated
their use in assistive robots, where our robot responds with pre-programmed actions. 
We have focused on 
the algorithms for temporal segmentation and labeling while using 
simple bounding-box detection and tracking algorithms. However, 
improvements to object perception and task-planning, while taking into consideration the 
human-robot interaction aspects, are needed for making assistive robots working efficiently alongside humans.  


\section{Acknowledgements}
We thank Li Wang for his significant contributions to the robotic experiments. We also thank 
Yun Jiang, Jerry Yeh, Vaibhav Aggarwal, and Thorsten Joachims for useful discussions. 
This research was funded in part by ARO award W911NF-12-1-0267, and by Microsoft Faculty Fellowship and Alfred P. Sloan Research Fellowship to one of us (Saxena).

{  
\bibliographystyle{apalike}
\bibliography{references}
}

\end{document}